\documentclass[iicol]{sn-jnl}

\usepackage{graphicx}%
\usepackage{multicol,multirow}%
\usepackage{amsmath,amssymb,amsfonts}%
\usepackage{amsthm}%
\usepackage{mathrsfs}%
\usepackage[title]{appendix}%
\usepackage{xcolor}%
\usepackage{textcomp}%
\usepackage{manyfoot}%
\usepackage{booktabs}%
\usepackage{algorithm}%
\usepackage{algorithmicx}%
\usepackage{algpseudocode}%
\usepackage{listings}%
\usepackage{lipsum}
\usepackage{geometry}

\usepackage{lmodern} 
\usepackage{natbib}
\usepackage{CJK}
\newcommand{\tabincell}[2]{\begin{tabular}{@{}#1@{}}#2\end{tabular}}
\usepackage{xtab,afterpage}
\usepackage{marvosym}
\usepackage{tabularx}

\usepackage{framed} 

\begin{document}

\title{Human Pose-based Estimation, Tracking and Action Recognition with Deep Learning: A Survey}

\author[1]{Lijuan Zhou}\email{ieljzhou@zzu.edu.cn}
\author[1]{Xiang Meng}\email{mengxiangzzu@163.com}
\equalcont{These authors contributed equally to this work.}
\author[1]{Zhihuan Liu}\email{liuzhihuanzzu@163.com}
\equalcont{These authors contributed equally to this work.}
\author[1]{Mengqi Wu}\email{mengqiwuzzu@163.com}
\equalcont{These authors contributed equally to this work.}
\author*[1]{Zhimin Gao}\email{iegaozhimin@zzu.edu.cn}
\author[2]{Pichao Wang}\email{pichaowang@gmail.com}

\affil[1]{School of Computer and Artificial Intelligence, Zhengzhou University, China}
\affil[2]{Amazon Prime Video, USA}

\abstract{Human pose analysis has garnered significant attention within both the research community and practical applications, owing to its expanding array of uses, including gaming, video surveillance, sports performance analysis, and human-computer interactions, among others. The advent of deep learning has significantly improved the accuracy of pose capture, making pose-based applications increasingly practical. This paper presents a comprehensive survey of pose-based applications utilizing deep learning, encompassing pose estimation, pose tracking, and action recognition.Pose estimation involves the determination of human joint positions from images or image sequences. Pose tracking is an emerging research direction aimed at generating consistent human pose trajectories over time. Action recognition, on the other hand, targets the identification of action types using pose estimation or tracking data. These three tasks are intricately interconnected, with the latter often reliant on the former. In this survey, we comprehensively review related works, spanning from single-person pose estimation to multi-person pose estimation, from 2D pose estimation to 3D pose estimation, from single image to video, from mining temporal context gradually to pose tracking, and lastly from tracking to pose-based action recognition. As a survey centered on the application of deep learning to pose analysis, we explicitly discuss both the strengths and limitations of existing techniques. Notably, we emphasize methodologies for integrating these three tasks into a unified framework within video sequences. Additionally, we explore the challenges involved and outline potential directions for future research.}


\maketitle
\section{Introduction}
Human pose estimation, tracking, and pose-based action recognition represent three fundamental research directions within the field of computer vision. These areas have a broad spectrum of applications, spanning from video surveillance, human-computer interactions, gaming, sports analysis, intelligent driving, and the emerging landscape of new retail stores. Articulated human pose estimation involves the task of estimating the configuration of the human body in a given image or video. Human pose tracking targets to generate consistent pose trajectories over time, which is usually used to analyze the motion proprieties of human. Human pose-based or skeleton-based action recognition is to recognize the types of actions based on the pose estimation or tracking data. Although these three tasks fall within the domain of human motion analysis, they are typically treated as distinct entities in the existing literature.

Human motion analysis is a long-standing research topic, and there are a vast of works and several surveys on this task~\citep{gavrila1999visual,aggarwal1999human,moeslund2001survey,wang2003recent,moeslund2006survey,poppe2007vision,sminchisescu20083d,ji2009advances,moeslund2011visual}. In these surveys, human detection, tracking, pose estimation and motion recognition are usually reviewed together. Several survey papers have summarized the research on human pose estimation~\citep{Liu2015ASO,sarafianos20163d}, tracking~\citep{yilmaz2006object,watada2010human,salti2012adaptive,smeulders2013visual,wu2015object}, and action recognition~\citep{cedras1995motion,turaga2008machine,poppe2010survey,guo2014survey}.
With the development of deep learning, the three tasks have achieved significant improvements compared to hand-crafted feature era~\citep{zhu2016handcrafted,wang2018rgb}. The previous surveys either reviewed the whole vision-based human motion domain~\citep{gavrila1999visual,aggarwal1999human,moeslund2001survey,wang2003recent,moeslund2006survey,poppe2007vision,sminchisescu20083d,ji2009advances}, or have focused on specific tasks~\citep{Liu2015ASO,sarafianos20163d,wang2018rgb,chen2020monocular,liu2022recent,sun2022human,zheng2023deep,xin2023transformer}. However, there is no such survey paper which simultaneously reviews pose estimation, pose tracking, and pose recognition. Inspired by Lagrangian viewpoint of motion analysis~\citep{rajasegaran2023benefits}, pose information and tracking are beneficial for action recognition. Therefore, these three tasks are closely related each other. It is significantly useful for reviewing the methods linking the three tasks together, and providing a deep understanding for the separate solution of each task and more exploration for a unified solution of joint tasks.

\begin{figure*}[t]
	\begin{center}
		{\includegraphics[scale=0.24]{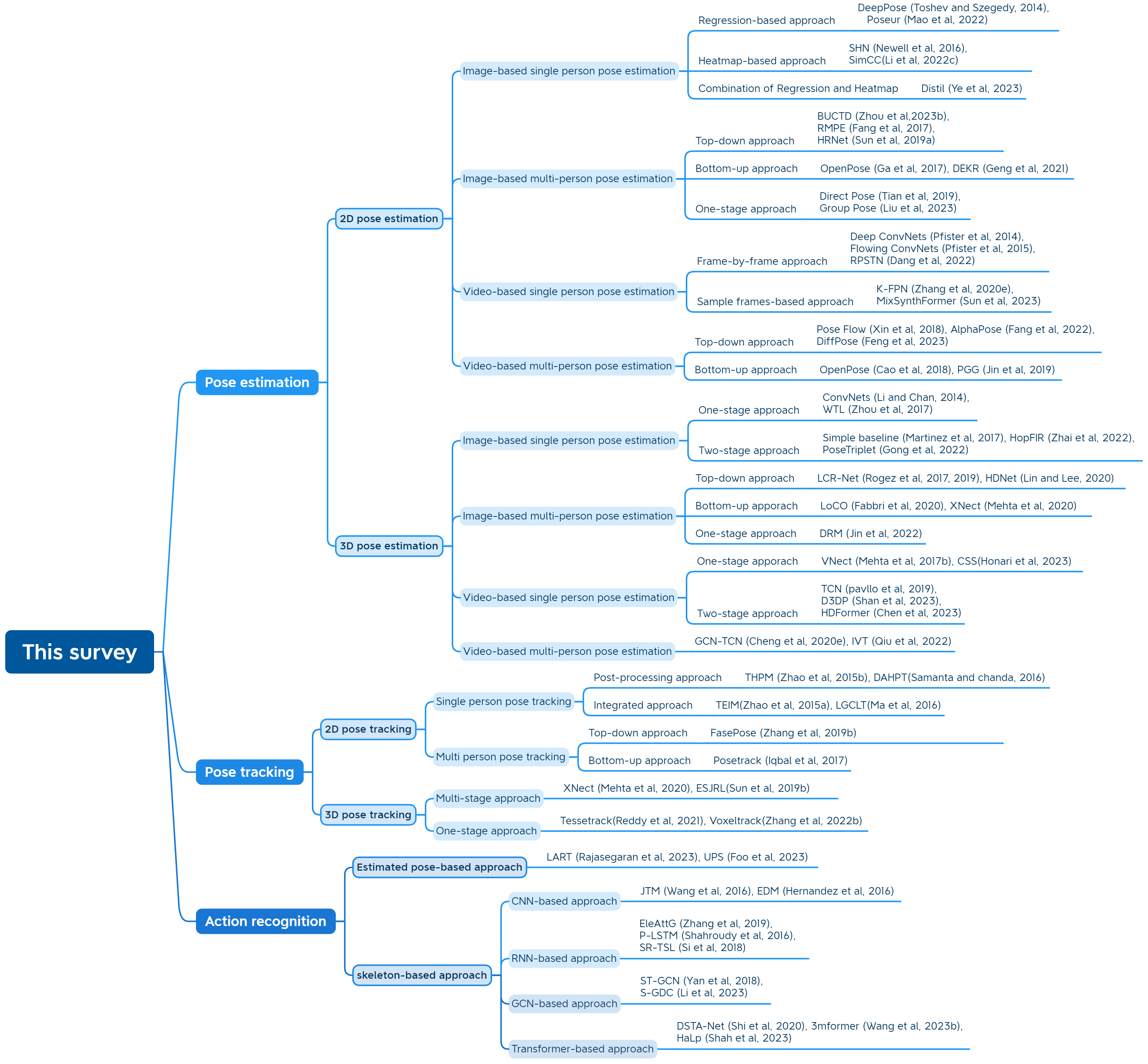}
			\caption{The taxonomy of this survey.}
			\label{fig:classification}
		}
	\end{center}
\end{figure*}

In this paper, we will conduct a comprehensive review of previous works using deep learning approach on these three tasks individually, and discuss the strengths and weaknesses of previous research paper. Furthermore, we elucidate the inherent connections that bind these three tasks together, while championing the adoption of a deep learning-based framework that seamlessly integrates them. Specifically, we will review previous works with deep learning from 2D pose estimation to 3D pose estimation from single images to videos, from mining temporal contexts gradually to pose tracking, and lastly from tracking to pose-based action recognition. According to the number of persons for pose estimation, 2D/3D pose estimation can be divided into single-person and multi-person pose estimation. Depending on the input to the networks, each category can be further divided into image and video-based single-person/multi-person pose estimation. To link the poses across the frames, pose tracking can be divided into post-processing and integrated methods for single-person pose tracking, top-down and bottom-up approaches for multi-person pose tracking. After getting the trajectory of poses in the videos, pose-based action recognition could be naturally conducted which can be divided into estimated pose and skeleton-based action recognition. The former takes RGB videos as the input and jointly conducts pose estimation, tracking, and action recognition. The latter extracts skeleton sequences captured by sensors such as motion capture, time-of-flight, and structured light cameras for action recognition. For skeleton-based action recognition, four categories are identified including Convolutional Neural Networks (CNN), Recurrent Neural Networks (RNN), Graph Neural Networks (GCN) and Transformer-based approaches. Fig.~\ref{fig:classification} illustrates the taxonomy of this survey.

The key novelty of this survey is the focus on three closely related tasks that use deep learning approach, which has never been done in previous surveys. In reviewing the various methods, consideration has been given to the connections between the three tasks, hence, this survey tends to discuss the advantages and limitations of the reviewed methods from the viewpoint of assembling them to get more practical applications. This is the first survey to put them together to analysis their inner connections in deep learning era. Besides, this survey distinguishes itself from other surveys through the following contributions:
\begin{itemize}
\item A thorough and all-encompassing coverage of the most advanced deep learning-based methodologies developed since 2014. This extensive coverage affords readers a comprehensive overview of the latest research methodologies and their outcomes.
\item An insightful categorization and analysis of methods on the three tasks, and highlights of the pros and cons, promoting potential exploration of better solutions.
\item An extensive review of the most commonly used benchmark datasets for these three tasks, and the state-of-the-art results on the benchmark datasets. 
\item An earnest discussion of the challenges of three tasks and potential research directions through limitation analysis of available methods.
\end{itemize}

Subsequent sections of this survey are organized as follows. Sections ~\ref{poseestimation} through ~\ref{sec:ar} delve into the methods of pose estimation, pose tracking, and action recognition, respectively. Commonly used benchmark datasets and the performance comparison for three tasks are described in Section~\ref{datasets}. Challenges of these three tasks and pointers to future directions are presented in Section~\ref{challenges}. The survey provides concluding remarks in Section~\ref{conclusion}.

\section{Pose estimation}\label{poseestimation}
Human representation can be approached through three distinct models: the kinematic model, the planar model, and the volumetric model. The kinematic model employs a combination of joint positions and limb orientations to faithfully depict the human body's structure. In contrast, the planar model utilizes rectangles to represent both body shape and appearance, while the volumetric model leverages mesh data to capture the intricacies of the human body's shape. It's essential to underscore that this paper exclusively focuses on the kinematic model-based human representation.

\begin{figure*}[t]
	\begin{center}
		{\includegraphics[scale=0.7]{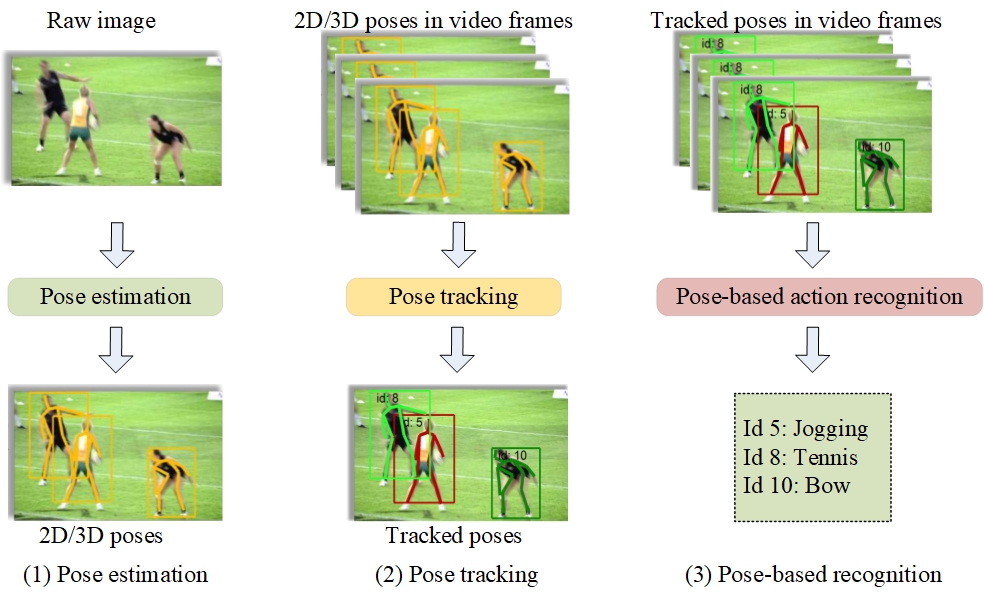}
			\caption{The relationship among the three tasks.}
			\label{fig:relationship}
		}
	\end{center}
\end{figure*}

Pose estimation, pose tracking and action recognition are three intimately interrelated tasks. Fig.~\ref{fig:relationship} shows the relationship among the three tasks.
Pose estimation aims to estimate joint coordinates from an image or a video. Pose tracking is an extension of pose estimation in the context of videos, which associates each estimated pose with its corresponding identity over time. It is interesting noting that a recent work~\citep{choudhury2023tempo} tends to estimate poses after tracking volumes of persons, which implies that the two-way relationship of pose estimation and tracking. Pose-based action recognition aims to give the tracked pose with an identity the corresponding action label. 
	
For pose estimation, we generally classify the reviewed methods into two categories, 2D pose estimation and 3D pose estimation. The 2D pose estimation is to estimate a 2D pose $(x,y)$ coordinates for each joint from a RGB image or video while 3D pose estimation is to estimate a 3D pose $(x,y,z)$ coordinates. 

\vspace{-0.3cm}
\subsection{2D pose estimation}

For 2D pose estimation, two sub-divisions are identified, single-person pose estimation and multi-person pose estimation. Depending on the input to the networks, single (multi) person pose estimation could be further divided into image-based single (multi) person pose estimation and video-based single (multi) person pose estimation.

\vspace{-0.3cm}
\subsubsection{Image-based single-person pose estimation}

\begin{figure*}[t]
	\begin{center}
		{\includegraphics[scale=0.5]{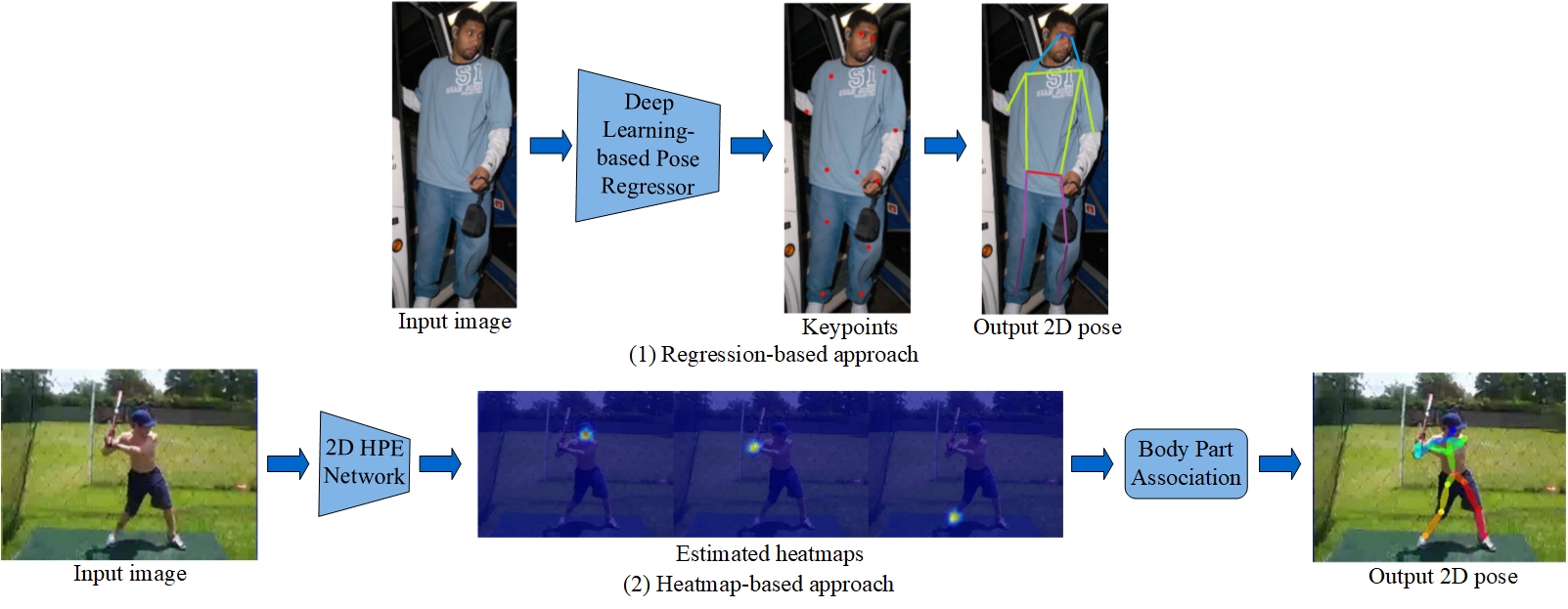}
			\caption{The framework of two approaches for image-based 2D SPPE.}
			\label{fig:img-SPPE}
		}
	\end{center}
\end{figure*}

For image-based Single-Person Pose Estimation (SPPE), the task involves providing the position and a rough scale of a person or their bounding box as a precursor to the estimation process. Early works adopt the pictorial structures framework that represents an object by a collection of parts arranged in a deformable configuration, and a part in the collection is an appearance template matched in an image. Different from early works, the deep learning-based methods target to locate keypoints of human parts. Two typical frameworks, namely, direct regression and heatmap-based approaches, are available for image-based single-person pose estimation. In the direct regression-based approach, keypoints are directly predicted from
the image features, whereas the heatmap-based approach initially generates heatmaps and subsequently
infers keypoint locations based on these heatmaps. Fig.~\ref{fig:img-SPPE} provides an illustrative overview of the general framework for image-based 2D SPPE, showcasing the two predominant approaches.

(1) Regression-based approach
   
\begin{figure*}[t]
 	\begin{center}
 		{\includegraphics[scale=0.5]{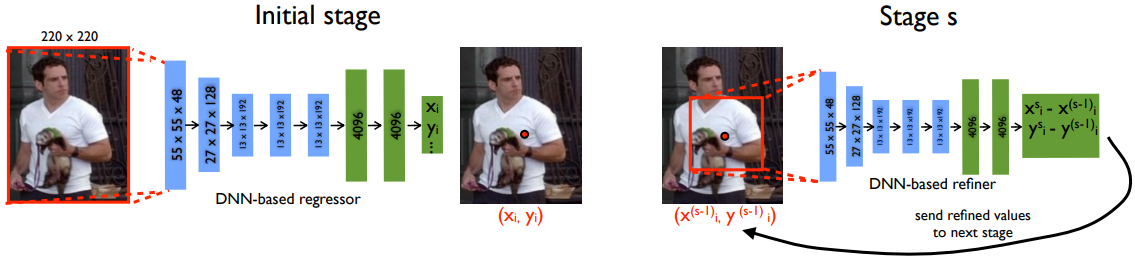}
 			\caption{The DeepPose architecture~\citep{toshev2014deeppose}.}
 			\label{fig:DeepPose}
 		}
 	\end{center}
\end{figure*}

 The pioneer work~\citep{toshev2014deeppose}, DeepPose, formulates pose estimation as a convolutional neural network(CNN)-based regression task towards body joints. A cascade of regressors are adopted to refine the pose estimates, as shown in Fig.~\ref{fig:DeepPose}. This work could reason about pose in a holistic fashion in occlusion situations. Carreira et al.~\citep{carreira2016human} introduced the Iterative Error Feedback approach, wherein prediction errors were recursively fed back into the input space, resulting in progressively improved estimations. Sun et al.~\citep{sun2017compositional} presented a reparameterized pose representation using bones instead of joints. This method defines a compositional loss function that captures the long range interactions within the pose by exploiting the joint connection structure. In more recent developments, ~\citep{luvizon2019human} introduced a novel approach that employed softmax functions to convert heatmaps into coordinates in a fully differentiable manner. This innovative technique was coupled with a keypoint error distance-based loss function and context-based structures.
 
Subsequently, researchers~\citep{mao2021tfpose,li2021pose,mao2022poseur,panteleris2022pe} began exploring pose estimation methods based on transformer architectures. The attention modules in transformers offered the ability to capture long-range dependencies and global evidence crucial for accurate pose estimation. For example, TFPose~\citep{mao2021tfpose} first introduced Transformer to the pose estimation framework in a regression-based manner. PRTR~\citep{li2021pose} introduced a two-stage, end-to-end regression-based framework that employed cascading Transformers, achieving state-of-the-art performance among regression-based methods. Mao et al.~\citep{mao2022poseur} framed pose estimation as a sequence prediction task, which they addressed with the Poseur model.
    
However, it's worth noting that these direct regression methods sometimes struggle in high-precision scenarios. This limitation may stem from the intricate mapping of RGB images to $(x,y)$ locations, adding unnecessary complexity to the learning process and hampering generalization. For instance, direct regression may encounter challenges when handling multi-modal outputs, where a valid joint appears in two distinct spatial locations. The constraint of producing a single output for a given regression input can limit the network's ability to represent small errors, potentially leading to over-training.
    
   (2) Heatmap-based approach
 
  Heatmaps have gained substantial attention due to its ability to provide comprehensive spatial information, making itself invaluable for training Convolutional Neural Networks (CNNs). This has spurred a surge of interest in the development of CNN architectures for pose estimation. Jain et al.~\citep{jain2014learning} pioneered an approach where multiple CNNs were trained for independent binary body-part classification, with each network dedicated to a specific feature. This strategy effectively constrained the network's outputs to a much smaller class of valid configurations, enhancing overall performance. Recognizing the importance of structural domain constraints, such as the geometric relationships between body joint locations, Tompson et al.~\citep{tompson2014joint} pursued a joint training approach, simultaneously training CNNs and graphical models for human pose estimation. Similarly, Chen and Yuille~\citep{chen2014articulated} adopt Convnets to learn conditional probabilities for the presence of parts and their spatial relationships within image patches. To address the limitations of pooling techniques in ~\citep{tompson2014joint} for improving spatial locality precision, Tompson et al.~\citep{tompson2015efficient} proposed a position refinement model (namely, a multi-resolution Convents) that is trained to predict the joint offset location within a localized region of the image. The works of ~\citep{tompson2014joint},~\citep{chen2014articulated} and~\citep{tompson2015efficient} sought to merge the representational flexibility inherent in graphical models with the efficiency and statistical power offered by CNNs. To avoid using graphical models, Wei et al.~\citep{wei2016convolutional} introduced the Convolutional Pose Machines to learn long-range spatial relationships without explicitly adopting graphical models. Hu and Ramanan~\citep{hu2016bottom} proposed an architecture that could be used for multiple stages of predictions, and ties weights in the bottom-up and top-down portions of computation as well as across iteration. Similarly, Newell et al.~\citep{newell2016stacked} proposed the Stacked Hourglass Network (SHN) for single-person pose estimation. The SHN leverages a series of successive pooling and upsampling steps to generate a final set of predictions, showcasing its efficacy. In addressing challenging scenarios characterized by severe part occlusions, Bulat and Tzimiropoulos~\citep{bulat2016human} presented a detection-followed-by-regression CNN cascade. This robust approach adeptly infers poses, even in the presence of significant occlusions. Lifshitz et al.~\citep{lifshitz2016human} introduced a novel voting scheme that harnesses information from the entire image, allowing for the aggregation of numerous votes to yield highly accurate keypoint detections. Chu et al.~\citep{chu2017multi} incorporated CNNs into their approach, enhancing it with a multi-context attention mechanism for pose estimation. This dynamic mechanism autonomously learns and infers contextual representations, directing the model's focus toward regions of interest. Furthermore, Yang et al.~\citep{yang2017learning} devised a Pyramid Residual Module (PRMs) to bolster the scale invariance of CNNs. PRMs effectively learn feature pyramids, which prove instrumental in precise pose estimation. 
    
 With the development of Generative Adversarial Networks (GAN)~\citep{goodfellow2014generative}, Chen et al.~\citep{chen2017adversarial} designed discriminators to distinguish the real poses from the fake ones to incorporate priors about the structure of human bodies. Ning et al.~\citep{ning2017knowledge} proposed to explore external knowledge to guide the network training process using learned projections that impose proper prior. Sun et al.~\citep{sun2017human} presented a two-stage normalization scheme, human body normalization and limb normalization, to make the distribution of the relative joint locations compact, resulting in easier learning of convolutional spatial models and more accurate pose estimation. Marras et al.~\citep{marras2017deep} introduced a Markov Random Field (MRF)-based spatial model network between the coarse and the refinement model that introduces geometric constraints on the relative locations of the body joints. To deal with annotating pose problem, Liu and Ferrari~\citep{liu2017active} presented an active learning framework for pose estimation. Ke et al.~\citep{ke2018multi} proposed a multi-scale structure-aware network for human pose estimation. 
 Peng et al.~\citep{peng2018jointly} proposed adversarial data augmentation for jointly optimize data augmentation and network training. The main idea is to design an augmentation network (generator) that competes against a target network (discriminator) by generating "hard" augmentation operations online. Tang et al.~\citep{tang2018deeply} introduced a Deeply Learned Compositional Model for pose estimation by exploiting deep neural networks to learn compositions of human body. Nie et al.~\citep{nie2018human} proposed the parsing induced learner including a parsing encoder and a pose model parameter adapter,  which estimates dynamic parameters in the pose model through joint learning to extract complementary useful features for more accurate pose estimation. Nie et al.~\citep{nie2018mutual} proposed to jointly conduct human parsing and pose estimation in one framework by incorporating information from their counterparts, giving more robust and accurate results. Tang and Wu~\citep{Tang_2019_CVPR} proposed a data-driven approach to group-related parts based on how much information they share, and then a part-based branching network (PBN) is introduced to learn representations specific to each part group. To speed up the pose estimation, Zhang et al.~\citep{Zhang_2019_CVPR} presented a Fast Pose Distillation (FPD) model that trains a lightweight pose neural network architecture capable of executing rapidly with low computational cost, by effectively transferring pose structure knowledge of a robust teacher network.
 
\begin{figure*}[t]
	\begin{center}
		{\includegraphics[scale=0.5]{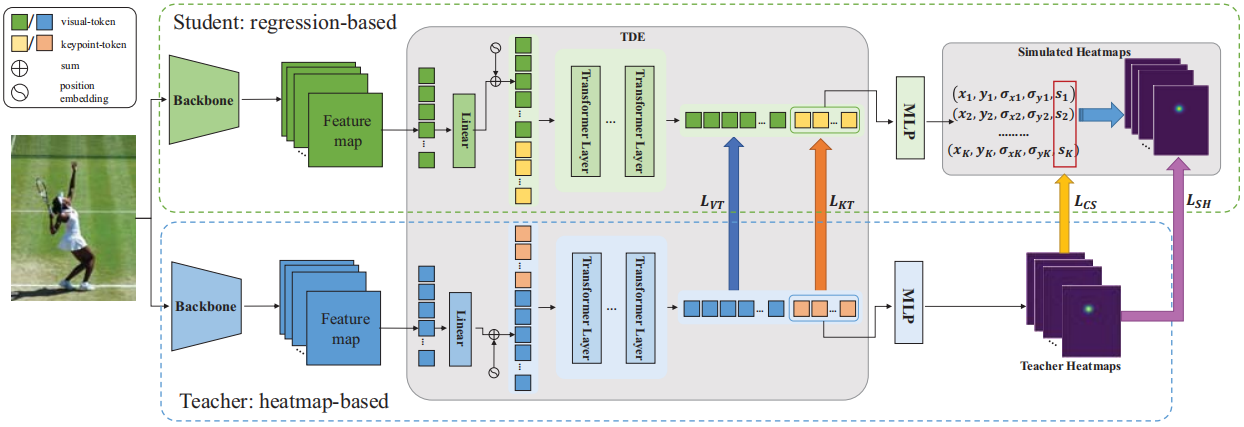}
			\caption{The DistilPose framework~\citep{ye2023distilpose}.}
			\label{fig:distilPose}
		}
	\end{center}
\end{figure*}

In summary, regression-based methods have advantages in speed but disadvantages in accuracy on pose estimation task. Heatmap-based methods can explicitly learn spatial information by estimating heatmap likelihood, resulting in high accuracy. However, heatmap-based methods suffer seriously a long-standing challenge known as the quantization error problem, which is caused by mapping the continuous coordinate values into discretized downscaled heatmaps. To address this problem, Li et al~\citep{li2022simcc} proposed a Simple Coordinate Classification (SimCC) method which formulates pose estimation as two classification tasks for horizontal and vertical coordinates. Despite the improvement in quantization error, the estimation of heatmaps requires exceptionally high computational cost, resulting in slow preprocessing operations. Therefore, how to take advantage of both heatmap-based and regression-based methods remains a challenging problem. Some works~\citep{li2021human,ye2023distilpose} tend to solve the above problem by transferring the knowledge from heatmap-based to regression-based models. However, due to the different output spaces of regression models and heatmap models, directly transferring knowledge between heatmaps and vectors may result in information loss. To the end, DistilPose~\citep{ye2023distilpose} (as shown in Fig.~\ref{fig:distilPose}) is proposed to transfer heatmap-based knowledge from a teacher model to a regression-based student model through token-distilling encoder and simulated heatmaps.

\vspace{-0.3cm}
\subsubsection{Image-based multi-person pose estimation}

\begin{figure*}[t]
	\begin{center}
		{\includegraphics[scale=0.35]{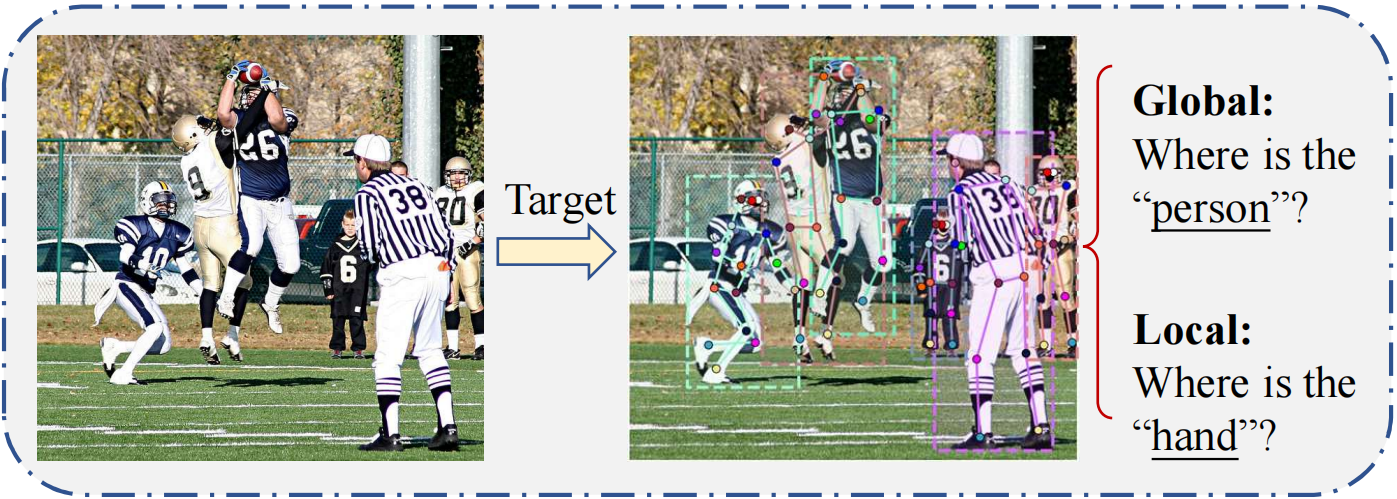}
			\caption{Perception of multi-person pose estimation task~\citep{yang2023explicit}.}
			\label{fig:per-MPPE}
		}
	\end{center}
\end{figure*}

Compared with single-person pose estimation (SPPE), multi-person pose estimation (MPPE) is more difficult. First, the number or the position of the person is not given, and the pose can occur at any position or scale; second, interactions between people induce complex spatial interference, due to contact, occlusion, and limb articulations, making association of parts difficult; third, runtime complexity tends to grow with the number of people in the image, making realtime performance a challenge. MPPE must address both global (human-level) and local (keypoint-level) dependencies (as depicted in Fig.~\ref{fig:per-MPPE}), which involve different levels of semantic granularity. Mainstream solutions are normally two-stage approaches, which divide the problem into two separate subproblems including global human detection and local keypoint regression. Typically, two primary frameworks have been proposed to tackle these subproblems, known as the top-down and bottom-up approaches. Inspired by the success of end-to-end object detection, another viable solution is the one-stage approach. This approach aims to develop a fully end-to-end trainable method capable of unifying the two disassembled subproblems.

\begin{figure*}[t]
	\begin{center}
		{\includegraphics[scale=0.65]{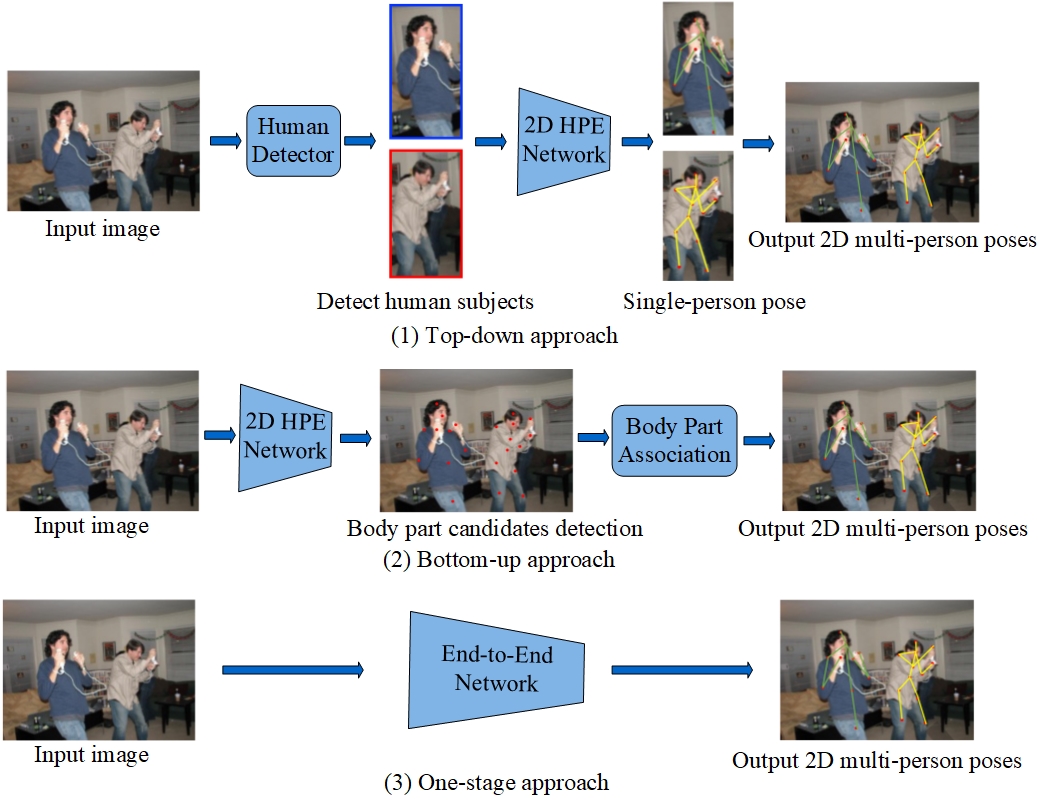}
			\caption{The framework of two approaches for image-based 2D MPPE. Part of the figure is from~\citep{zheng2020deep}.}
			\label{fig:img-MPPE}
		}
	\end{center}
\end{figure*}

(1) Top-down approach

Top-down approaches in multi-person pose estimation begin by detecting all individuals within a given image, as shown in Fig.~\ref{fig:img-MPPE}, and subsequently employ single-person pose estimation techniques within each detected bounding box. 

A group of methods~\citep{papandreou2017towards,he2017mask,xiao2018simple,moon2019posefix,sun2019deep,cai2020learning,huang2020devil,zhang2020distribution,wang2020graph,xu2022adaptive,JIANG2023109429,10093110} aim to designing and improving modules within pose estimation networks. Papandreou et al.~\citep{papandreou2017towards} adopt Faster RCNN~\citep{ren2015faster} for person detection and keypoints estimation within the bounding box. They introduce an aggregation procedure to obtain highly localized keypoint predictions, along with a keypoint-based Non-Maximum-Suppression (NMS) to prevent duplicate pose detection. Sun et al.~\citep{sun2019deep} proposed a novel High-Resolution net(HRNet) to learn such representation. To address systematic errors in standard data transformation and encoding-decoding structures that degrade top-down pipeline performance, Huang et al.~\citep{huang2020devil} proposed solutions to correct common biased data processing in human pose estimation.

Human detectors may fail in the first step of top-down pipeline due to occlusion affected by the overlapping of limbs. Another group of works~\citep{iqbal2016multi,fang2017rmpe,chen2018cascaded,Su_2019_CVPR,qiu2020peeking} aim to address this issue.
Fang et al.~\citep{fang2017rmpe} proposed a novel Regional Multi-person Pose Estimation (RMPE) to facilitate pose estimation even when inaccurate human bounding boxes exist. Chen et al.~\citep{chen2018cascaded} designed a Cascaded Pyramid Network (CPN) that contains GlobalNet and RefineNet for localizing simple and hard keypoints with occlusion respectively. Su et al.~\citep{Su_2019_CVPR} proposed two novel modules to perform the enhancement of the information for the multi-person pose estimation under occluded scenes, namely, Channel Shuffle Module (CSM) and Spatial, Channel-wise Attention Residual Bottleneck (SCARB), where CSM promoting cross-channel information communication among the pyramid feature maps and SCARB highlighting the information of feature maps both in the spatial and channel-wise context. An occluded pose estimation and correction module~\citep{qiu2020peeking} is proposed to solve the occlusion problem in crowd pose estimation.

Much like single-person pose estimation, multi-person pose estimation has also undergone rapid advancements, transitioning from CNNs to vision transformer networks. Some recent works tend to treat transformer as a better decoder. TransPose~\citep{yang2021transpose} processes the features extracted by CNNs to model the global relationship. Zhou et al.~\citep{zhou2023rethinking} proposed a Bottom-Up Conditioned Top-Down pose estimation (BUCTD) method which modifies TransPose to accept conditions as side-information generated by CTD. Different from other top-down methods, BUCTD applies a bottom-up model as a person detector. TokenPose~\citep{li2021tokenpose} proposes a token-based representation to estimate the locations of occluded keypoints and model the relationship among different keypoints. HRFormer~\citep{YuanFHLZCW21} proposes to fuse multi-resolution features by a transformer module. The above works either require CNNs for feature extraction or careful designs of transformer structures. In contrast, a simple yet effective baseline model, ViTPose~\citep{xu2022vitpose}, is proposed based on the plain vision transformers. 
    
(2) Bottom-up approach

In contrast to the top-down approach, the bottom-up approach initially detects all individual body parts or keypoints and subsequently associates them with the corresponding subjects using part association strategies.
The seminal work of Pishchulin et al.~\citep{pishchulin2016deepcut} proposed a bottom-up approach that jointly labels part detection candidates and associates them to individual people. However, solving the integer linear programming problem over a fully connected graph is an NP-hard problem and the average processing time is on the order of hours. In the work by Insafutdinov et al. ~\citep{insafutdinov2016deepercut}, a more robust part detector and innovative image-conditioned pairwise terms were proposed to enhance runtime efficiency. Nevertheless, this work encountered challenges in precisely regressing the pairwise representations and a separate logistic regression is required. Iqbal and Gall~\citep{iqbal2016multi} considered multi-person pose estimation as a joint-to-person association problem. They construct a fully connected graph from a set of detected joint candidates in an image and resolve the joint-to-person association and outlier detection using integer linear programming. OpenPose~\citep{cao2017realtime,cao2018realtime} proposes the first bottom-up representation of association scores via Part Affinity Fields (PAFs) which are a set of 2D vector fields that encode the location and orientation of limbs over the image domain. Kreiss et al.~\citep{Kreiss_2019_CVPR} proposed to use a Part Intensity Field (PIF) for body parts localization and a PAF for body part association with each other to form full human poses. To handle missed small-scale persons, Cheng et al.~\citep{CHENG2023109403} proposed multi-scale training and dual anatomical canters to enhance the network. The above methods mainly apply heatmap prediction based on overall \textit{L}2 loss to locate keypoints. However, minimizing \textit{L}2 loss cannot always locate all keypoints since each heatmap often includes multiple body joints. To solve this problem, Qu et al.~\citep{qu2023characteristic} proposed to optimize heatmap prediction based on minimizing the distance between the characteristic functions of the predicted and ground-truth heatmaps.

Different from the above two-stage bottom-up approach, some works focus on joint detection and grouping, which belong to single-stage bottom-up approach. Newell et al.~\citep{newell2017associative} simultaneously produced score maps and pixel-wise embedding to group the candidate keypoints among different people to get final multi-person pose estimation. Kocabas et al.~\citep{kocabas2018multiposenet} designed a MultiPoseNet that jointly handle person detection, person segmentation and pose estimation problems, by the implementation of Pose Residual Network (PRN) which receives keypoint and person detections, and produces accurate poses by assigning keypoints to person instances.
To deal with the crowded scene, Li et al.~\citep{Li_2019_CVPR} built a new benchmark called CrowdPose and proposed two components, namely, joint-candidate single-person pose estimation and global maximum joints association, for crowded pose estimation. Jin et al.~\citep{jin2020differentiable} proposed a new differentiable hierarchical graph grouping method to learn human part grouping. Cheng et al.~\citep{cheng2020higherhrnet} extended the HRNet and proposed a higher resolution network (HigherHRNet) by deconvolving the high-resolution hetamaps generated by HRNet to solve the variation challenge. Besides the above bottom-up methods, some methods directly regress a set of pose candidates from image pixels and the keypoints in each candidate might be from the same person. A post-processing step is required to generate the final poses which are more spatially accurate. For instance, single-stage multi-person Pose Machine (SPM) method~\citep{nie2019single} applies a hierarchical structured 2D/3D pose representation to assist the long-range regression. The keypoints are predicted based on person-agnostic heatmaps so that grouping post-processing is required to assemble keypoints to the full-body pose. Disentangled Keypoint Regression (DEKR)~\citep{geng2021bottom} regresses pose candidates by learning representations that focus on keypoint regions. The pose candidates were scored and ranked to generate the final poses based on keypoints and center heatmap estimation loss. PolarPose~\citep{10034548} aims to simplify 2D regression to a classification task by performing it in polar coordinate.

    
(3) One-stage approach 

The one-stage approach aims to learn an end-to-end network for MPPE without person detection and grouping post-processing. Tian et al.~\citep{tian2019directpose} first proposed a one-stage method based on DirectPose to directly predict instance aware keypoints for all persons from an image. To boost both accuracy and speed, Mao et al.~\citep{mao2021fcpose} later presented a Fully Convolutional Pose (FCPose) estimation framework to build dynamic filters in compact keypoint heads. Meanwhile, Shi et al.~\citep{shi2021inspose} designed InsPose, which adaptively adjusts the network parameters for each instance. To reduce the effect of false positive poses in regression loss, the Single-stage Multi-person Pose Regression (SMPR) network~\citep{MIAO2023109743} was presented by adapting three positive pose identification strategies for initial and final pose regression, and the Non-Maximum Suppression (NMS) step. These methods could avoid the need for heuristic grouping in bottom-up methods or bounding-box detection and region of interest (RoI) cropping in top-down ones. However, they still require hand-crafted operations, like NMS, to remove duplicates in the postprocessing stage. To further remove NMS, a multi-person Pose Estimation framework with TRansformers (PETR)~\citep{shi2022end} regards pose estimation as a set prediction, which is the first fully end-to-end framework without any postprocessing. The above one-stage methods adopts a pose decoder with randomly initialized pose queries, making keypoint matching across persons ambiguous and training convergence slow. To this end, Yang et al.~\citep{yang2023explicit} proposed an Explicit box Detection process for pose estimation (ED-pose) by realizing each box detection using a decoder and cascading them to form an end-to-end framework, making the model fast in convergence, precise and scalable. 

Although the above end-to-end methods have achieved promising performance, they rely on complex decoders. For instance, ED-pose includes a human detection decoder and a human-to-keypoint detection decoder to detect human and keypoint boxes explicitly.
PETR includes a pose decoder and a joint decoder. In contrast, Group Pose~\citep{liu2023group} only uses a simple transformer decoder for pursing efficiency.  


In summary, top-down approaches directly leverage existing techniques for single-person pose estimation, but suffer from early commitment: if the person detector fails as it is prone to do when people are in close proximity, there is no recourse to recovery. Furthermore, the runtime of these top-down approaches is proportional to the number of people. For each detection, a single-person pose estimator is run, thus, the more people there are, the greater the computational cost. In contrast, bottom-up approaches are attractive due to their robustness to early commitment and the potential to decouple runtime complexity from the number of people in the image. Yet, bottom-up approaches do not directly leverage global contextual cues from other body parts and individuals. One-stage methods eliminate the intermediate operations like grouping, ROI, bounding-box detection, NMS and bypass the major shortcomings of both top-down and bottom-up methods.

\vspace{-0.3cm}
 \subsubsection{Video-based single-person pose estimation}
\begin{figure*}[t]
	\begin{center}
		{\includegraphics[scale=0.65]{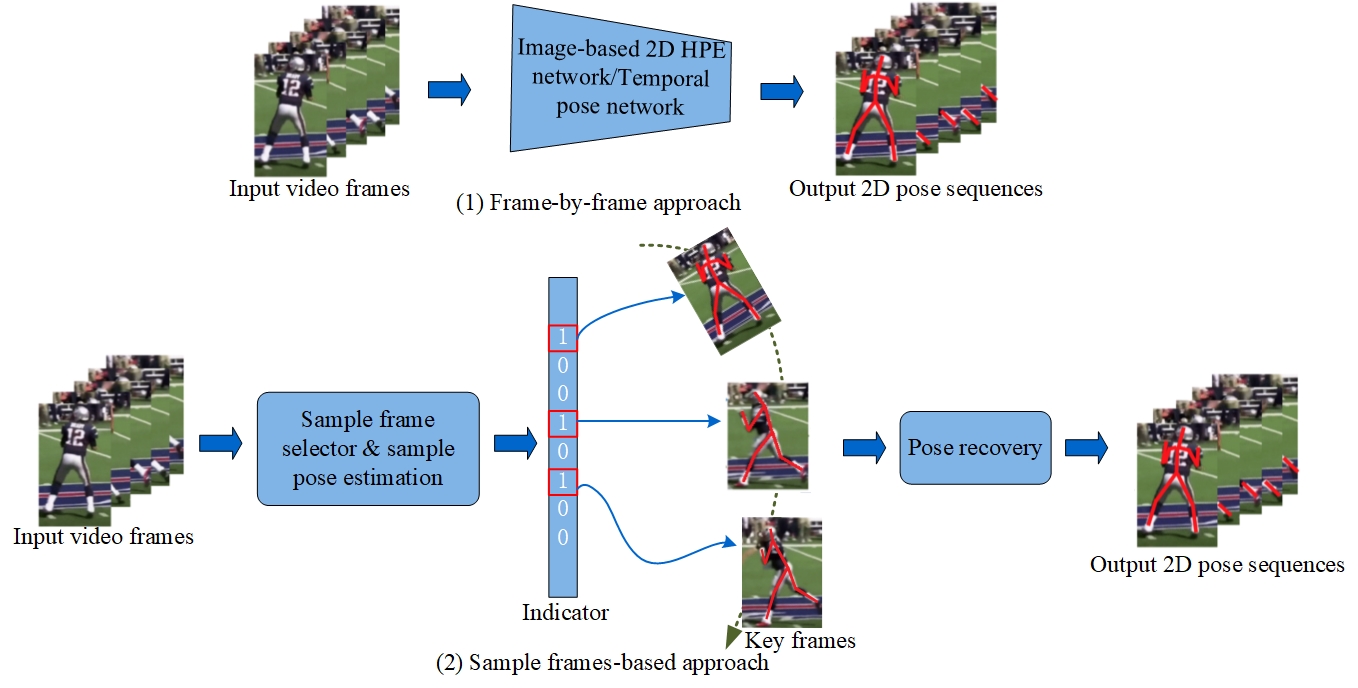}
			\caption{The framework of two approaches for video-based 2D SPPE.}
			\label{fig:vid-SPPE-2D}
		}
	\end{center}
\end{figure*}

Video-based pose estimation aims to estimate single or multiple poses in each video frame. Compared with image-based pose estimation, it is more challenging due to high variation in human pose and foreground appearance such as clothing and self-occlusion. For video-based pose estimation, human tracking is not considered in the video. Similar to image-based SPPE, direct regression and heatmap-based approaches are also available for video-based SPPE. However, differently, video-based pose estimation has the advantage of temporal information, which can enhance the accuracy of pose estimation but can also introduce additional computational overhead due to temporal redundancy. Therefore, achieving a balance between accuracy and efficiency is paramount for video-based pose estimation. Based on handling the efficiency, video-based SPPE approaches are categorized into the frame-by-frame approach and sample frames-based ones. Fig.~\ref{fig:vid-SPPE-2D} illustrates the general framework of two approaches for video-based SPPE.

(1) Frame-by-frame approach

The frame-by-frame approach, illustrated in Fig.~\ref{fig:vid-SPPE-2D}, focuses on estimating poses individually for each frame in the video sequence. With the success of image-based pose estimation, this category of methods mainly apply image-based pose estimation methods on each video frame by incorporating temporal information to keep geometric consistency across frames. The temporal information is normally captured by fusion from concatenated consecutive frames, applying 3D temporal convolution, using dense optical flow and pose propagation.

In the early stages of this approach, Pfister et al.~\citep{pfister2014deep} proposed to use deep ConvNets for estimating human pose in videos. They designed a regression layer to predict the location of upper-body joints while considering temporal information through the direct processing of concatenated consecutive frames along the channel axis. Grinciunaite et al.~\citep{grinciunaite2016human} extended 2D convolution into 3D convolution and temporal information can be efficiently represented in the third dimension of 3D convolutional for video-based human pose estimation. 

Some works tend to use optical flow to produce smooth movement. Pfister et al.~\citep{pfister2015flowing} 
used dense optical flow to predict joint positions for all neighboring frames and design spatial fusion layers to learn dependencies between the human parts locations. Song et al.~\citep{song2017thin} also utilized optical flow warping to capture high temporal consistency and propose spatio-temporal message passing layer to incorporate domain-specific knowledge into deep networks. Jain et al.~\citep{jain2014modeep} use Local Contrast Normalization and Local Motion Normalization to process the RGB image and optical-flow features respectively and then combine them to feed into Part-Detector network. These methods have high complexity due to dense flowing computation, making them not applicable in real-time applications. 

Subsequently, some works~\citep{gkioxari2016chained,charles2016personalizing,luo2018lstm,nie2019dynamic,li2019temporal,liwen2019temporal,xu2021vipnas,dang2022relation,jin2023kinematic} apply pose propagation which transfer features from previous frames to the current frame in an online fashion. For example, Charles et al.~\citep{charles2016personalizing} proposed a personalized ConvNet to estimate human pose including four stages: initial annotation, spatial matching, temporal propagation, 
and self evaluation. In the initial annotation stage, high-precision pose estimation is obtained by using flowing Convnets. Then Image patches from the new
frames without annotations are matched to image patches of body joints in frames with annotations by spatial matching process.  Dense optical flow is used for temporal propagation. Finally, the quality of the spatial-temporal propagated annotations is automatically evaluated to optimize the model. Luo et al.~\citep{luo2018lstm} proposed Long Short-Term Memory (LSTM) pose machines by combining  Convolutional Pose Machine (CPM)~\citep{wei2016convolutional} and LSTM network learning the temporal dependency among video frames to effectively capture the geometric relationships of joints in space and time. Nie et al.~\citep{nie2019dynamic} designed a Dynamic Kernel Distillation (DKD) model. The DKD model introduces a pose kernel distillator and transmits pose knowledge in time. Xu et al.~\citep{xu2021vipnas} proposed a novel neural architecture search to select the most effective temporal feature fusion for optimizing the accuracy and speed across video frames. Dang et al.~\citep{dang2022relation} proposed a Relation-based Pose Semantics Transfer Network (RPSTN) by designing a joint relation-guided pose semantic propagator to learn the temporal semantic continuity of poses. Despite various strategies are applied to reduce computation cost, this category of methods still leads to sub-optimal efficiency improvement due to the estimation frame by frame.

(2) Sample frames-based approach

This category of approach aims to recover all poses based on the estimated poses from selected frames. As shown in Fig.~\ref{fig:vid-SPPE-2D}, the general workflow includes sample pose estimation and all poses recovering. One line of works generates sample poses by selecting keyframes and estimating the poses of keyframes. For example, Zhang et al~\citep{zhang2020key} introduced a Key-Frame Proposal Network (K-FPN) to select informative frames and a human pose interpolation module to generate all poses from the poses in keyframes based on human pose dynamics. Pose dynamic-based dictionary formulation may become challenging when the pose sequence to be interpolated becomes complex. Therefore, to effectively exploit the dynamic information, REinforced MOtion Transformation nEtwork (REMOTE)~\citep{ma2022remote} includes a motion transformer to conduct cross frame reconstruction. Although the computational efficiency of the above works is improved due to keyframes, they still require to take cost on keyframe selection, making it hard to further reduce the complexity. To solve this problem, Zeng et al.~\citep{zeng2022deciwatch} proposed a novel Sample-Denoise-Recover pipeline (namely DeciWatch) to uniformly sample less than 10\% of video frames for estimation. The estimatied poses based on sample frames are denoised with a Transformer architecture and the rest poses are also recovered by another Transformer network. DeciWatch can be used in both 2D/3D pose estimation from videos and it can maintain or even improve the pose estimation accuracy as the previous methods with small cost on computation. Although uniform sampling reduces the cost of selecting keyframes, a refinement module is added to clean noisy poses. In contrast, MixSynthFormer~\citep{sunmixsynthformer} deletes the refinement module by combining a transformer encoder with an MLP-based mixed synthetic attention, thus pursing highly efficient 2D/3D video-based pose estimation.

Overall, frame-by-frame approaches could benefit from image-based pose estimation but suffer from the computation complexity. Sample frame-based approaches offer a solution to improve efficiency but raise questions about how to obtain sample frames and recover poses. The paper employs uniform sampling; however, considering the significant variations in joint movements under different actions, an adaptive sampling strategy might be more suitable for further enhancing efficiency. Additionally, the design of dynamic recovery methods should be explored to handle non-uniform sampling effectively.


\vspace{-0.3cm}
 \subsubsection{Video-based multi-person pose estimation}
 
Given the video-based SPPE just introduced, it is natural to extend them to handle multiple individuals. Following the taxonomy of video-based SPPE, most video-based MPPE approaches fall into frame-by-frame category. They can be achieved by employing image-based MPPE frame by frame. Therefore, the approaches of video-based MPPE can be categorized into Top-down and Bottom-up approaches.
 
 (1) Top-down approach
 
\begin{figure*}
 	\begin{center}
 		{\includegraphics[scale=0.65]{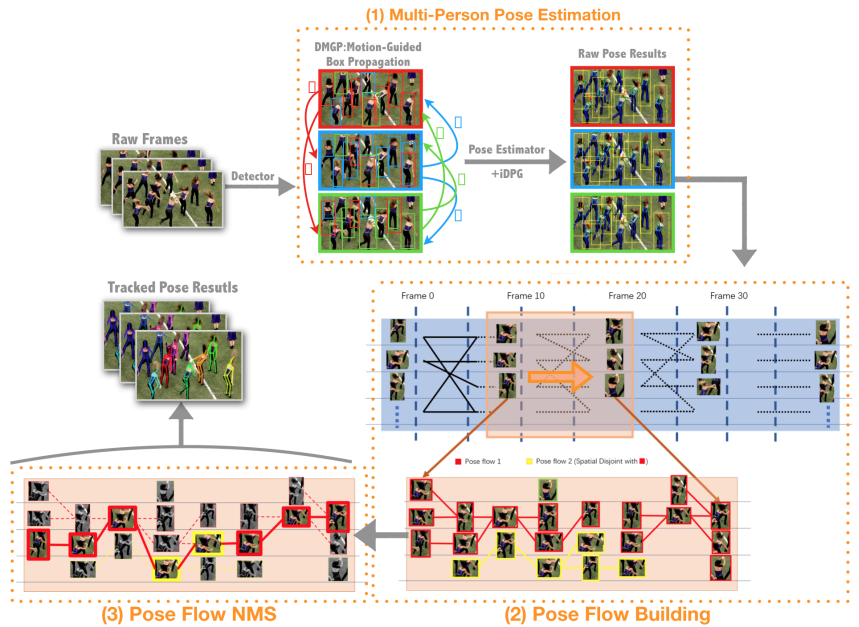}
 			\caption{The Pose Flow framework~\citep{xiu2018pose}.}
 			\label{fig:PoseFlow}
 		}
 	\end{center}
 \end{figure*}

 Top-down approaches mainly estimate poses by first detecting all persons for all frames and then conducting image-based single-person pose estimation frame by frame. Xiao et al.~\citep{xiao2018simple} proposed a simple baseline based on ResNet to estimate poses in each frame and the estimated poses were then tracked based on optical flow. Xiu et al.~\citep{xiu2018pose} estimated multiple poses for each frame based on RMPE method which can be replaced by other top-down methods for image-based MPPE. With the estimated poses in each frame, a Pose Flow Builder (PF-Builder) is proposed for building the association of cross-frame poses by maximizing overall confidence along the temporal sequence (as shown in Fig.~\ref{fig:PoseFlow}), and a Pose Flow Non-Maximum Suppression (PF-NMS) is designed to robustly reduce redundant pose flows and re-link temporal disjoint ones. Girdhar et al.~\citep{girdhar2018detect} estimated poses for each frame based on Mask R-CNN and then generated keypoint predictions linked over the video by lightweight tracking. Wang et al.~\citep{wang2020combining} proposed a clip tracking network to perform pose estimation and tracking simultaneously. To construct the clip tracking network, the 3D HRNet is proposed for estimating poses which incorporating temporal dimension into the original HRNet. AlphaPose~\citep{fang2022alphapose} is also proposed for joint pose estimation and tracking. In particular, all persons for each frame are firstly detected using off-the-shelf object detectors like YoloV3 or EfficientDet. To solve the quantization error, the symmetric integral keypoints regression method is then proposed to localize keypoints in different scales accurately. Pose-guided alignment module is applied on the predicted human re-id feature to obtain pose-aligned human re-id features after removing redundant poses based on NMS. At last, a pose-aware identity embedding is presented to produce tracking identity. Estimating poses frame by frame ignores motion dynamics which is fundamentally important for accurate pose estimation from videos. A recent method~\citep{feng2023mutual} presents Temporal Difference Learning based on Mutual Information (TDMI) for pose estimation. A multi-stage temporal difference encoder was designed for learning informative motion representations and a representation disentanglement module was introduced to distill task-relevant motion features to enhance frame representation for pose estimation. The temporal difference features can be applied in pose tracking by measuring the similarity of motions for data association. Gai et al.~\citep{gai2023spatiotemporal} proposed a Sptiotemporal Learning Transformer for video-based Pose estimation (SLT-Pose) to capture the shallow feature information. With the introduction of diffusion models in computer vision tasks (eg. image segmentation~\citep{amit2021segdiff}, object detection~\citep{chen2022diffusiondet}), DiffPose~\citep{feng2023diffpose} is the first diffusion model and formulates video-based pose estimation as a conditional heatmap generation problem.
 
 (2) Bottom-up approach
 
  Bottom-up approaches estimate poses by applying body part detection and grouping frame by frame. For example, one of the commonly used image-based MPPE methods, OpenPose\citep{cao2018realtime}, can be also applied for MPPE from video by directly estimating poses frame by frame. Jin et al.~\citep{jin2019multi} proposed a Pose-Guided Grouping (PGG) network for joint pose estimation and tracking. PGG consists of two components including SpatialNet and TemporalNet. SpatialNet tackles multi-person pose estimation by body part detection and part-level spatial grouping for each frame. TemporalNet extends SpatialNet to deal with online human-level temporal grouping.
 
 Overall, 2D HPE has been significantly improved with the development of deep learning techniques. For the image-based SPPE, heatmap-based approaches generally outperform regression-based ones in accuracy but may be of challenge in the quantization error problem. When extending SPPE to MPPE, both top-down and bottom-up approaches have their advantages and disadvantages. Moreover, both approaches have a challenge of reliable detection of individual persons under significant occlusion. Person detector in top-down approaches may fail in identifying the boundaries of overlapped human bodies. Body part association for occluded scenes may fail in bottom-up approaches. One-stage approaches bypass both the shortcomings of top-down and bottom-up ones,  yet they are still less frequently used. With the advancement of image-based pose estimation, it is natural to extend it to videos by directly applying off-the-shelf image-based pose estimation methods frame by frame or incorporating a temporal network. Sample frames-based methods are preferred for the pose estimation from videos since they can largely improve efficiency without looking at all frames, while they have been used less in the video-based MPPE. Considering the benefits of one-stage approaches for image-based MPPE, more effort is required to explore one-stage approaches for video-based ones.
 
 \vspace{-0.3cm}
\subsection{3D pose estimation}
Generally speaking, recovering 3D pose is considered more difficult than 2D pose estimation, due to the larger 3D pose space and more ambiguities. An algorithm has to be invariant to some factors, including background scenes, lighting, clothing shape and texture, skin color, and image imperfections, among others.  

\vspace{-0.3cm}
\subsubsection{ Image-based single-person pose estimation}
\begin{figure*}[t]
	\begin{center}
		{\includegraphics[scale=0.65]{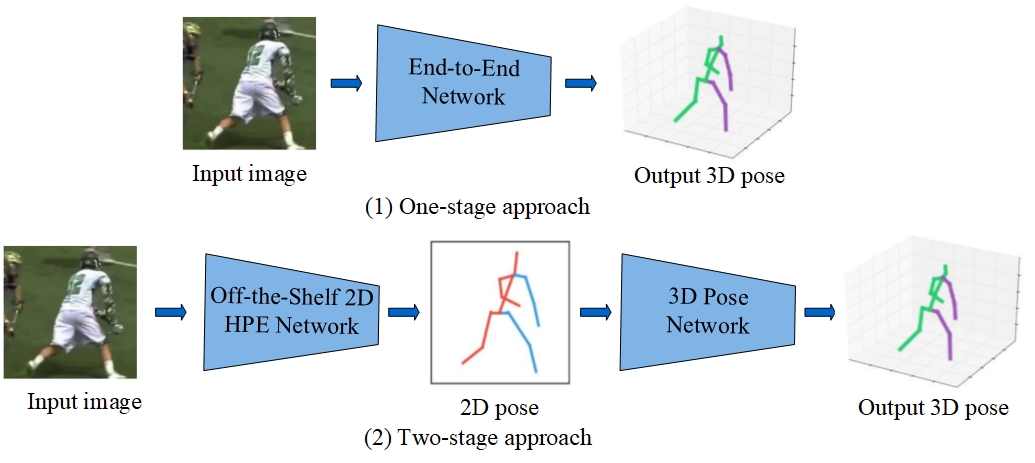}
			\caption{The framework of two approaches for image-based 3D SPPE.}
			\label{fig:img-SPPE-3D}
		}
	\end{center}
\end{figure*}

Imaged-based single-person 3D human pose estimation (HPE) can be classified into skeleton-based and mesh-based approaches. The former one estimates 3D human joints as the final output and the latter one is required to reconstruct 3D human mesh representation. Since this paper focuses only on the kinematic model-based human representation, we only review skeleton-based approaches which can be further categorized into one-step pose estimation and two-steps pose estimation (recover 3D pose from 2D pose). Fig.~\ref{fig:img-SPPE-3D} shows the general framework of the two approaches for image-based 3D SPPE.

(1) One-stage approach

This category of approaches directly infer 3D pose from images without estimating 2D pose representation.
Li and Chan~\citep{li20143d} first proposed to estimate 3D poses from monocular images using ConvNets. The framework consists of two types of tasks: joint point regression and joint point detection. Both tasks take bounding box images containing human subjects as input. The regression task aims to estimate the positions of joint points relative to the root joint position, while each detection task classifies whether one specific joint is present in the local window or not.

The multi-task learning framework is the first to show that deep neural networks can be applied to 3D human pose estimation from single images. However, one drawback of these regression-based methods is their limitation in predicting only one pose for a given image This may cause difficulties in images where the pose is ambiguous due to partial self-occlusion, and hence several poses might be valid. In contrast, Li et al.~\citep{li2015maximum} proposed a unified framework for maximum-margin structured learning with a deep neural
network for 3D human pose estimation, where the unified framework can jointly learn the image and pose feature representations and the score function. Tekin et al.~\citep{tekin2016structured} introduced an architecture relying on an overcomplete auto-encoder to learn a high-dimensional latent pose representation for joint dependencies. Zhou et al.~\citep{zhou2016deep} proposed a novel method which directly embeds a kinematic object model into the deep neutral network learning, where the kinematic function is defined on the appropriately parameterized object motion variables. Mehta et al.~\citep{mehta2017monocular} explored transfer learning to leverage the highly relevant middle and high-level features from 2D pose datasets in conjunction with the existing annotated 3D pose datasets. Similarly, Zhou et al.~\citep{zhou2017towards} introduced a Weakly-supervised Transfer Learning (WTL) method that employs mixed 2D and 3D labels in a unified deep neural network, which is end-to-end and fully exploits the correlation between the 2D pose and depth estimation sub-tasks. Since regressing directly from image space, one-step-based methods often require a high computation cost.

(2) Two-stage approach

This category of approaches infer 3D pose from the intermediately estimated 2D pose. They are often conducted in two steps: 1) estimating 2D pose based on image-based single-person 2D pose estimation methods. 2) Lifting the 2D pose to 3D pose through a simple regressor. For instance, Martinez et al.~\citep{martinez2017simple} proposed a simple baseline based on a fully connected residual network to regress 3D poses from 2D poses. This baseline method achieves good results at that time, however, it could fail due to reconstruction ambiguity of over-reliance on 2D pose detector. To overcome this problem, several techniques are applied such as replacing 2D poses with heatmaps for estimating 3D poses~\citep{tekin2017learning,zhou2019hemlets}, regressing 3D poses from 2D poses and depth information~\citep{wang2018drpose3d,CARBONERALUVIZON2023109714}, selecting best 3D poses from 3D pose hypotheses using ranking networks~\citep{jahangiri2017generating,sharma2019monocular,li2019generating}. 

With the introduction of Graph convolutional networks(GCN)-based representation for human joints, some methods~\citep{ci2019optimizing,zhao2019semantic,choi2020pose2mesh,zeng2020srnet,liu2020comprehensive,zou2021modulated,xu2021graph,2023Learning,10179252} apply GCN for lifting 2D to 3D poses. To overcome the limitations of shared weights in GCN, a locally connected network (LCN)~\citep{ci2019optimizing} was proposed which leverages a fully connected network and GCN to encode the relationship among joints. Similarly, Zhao et al.~\citep{zhao2019semantic} proposed a semantic-GCN to learn channel-wise weights for edges. A Pose2Mesh~\cite{choi2020pose2mesh} based on GCN was proposed to refine the intermediate 3D pose from its PoseNet. Xu and Takano~\citep{xu2021graph} proposed a Graph Stacked Hourglass (GraphSH) networks which consists of repeated encoder-decoder for representing three different scales of human skeletons. To overcome the loss of joint interactions in current GCN methods, Zhai et al.~\citep{zhai2023hopfir} proposed Hop-wise GraphFormer with Intragroup Joint Refinement (HopFIR) for lifting 3D poses. 

Inspired by the recent success in the nature language field, there is a growing interest in exploring the use of Transformer architecture for vision tasks. Lin et al.~\citep{lin2021end} first applied Transformer for 3D pose estimation. A multi-layer Transformer with progressive dimensionality reduction was proposed to regress the 3D coordinates of joints. Here, the standard transformer ignores the interaction of adjacency nodes. To overcome this problem, Zhao et al.~\citep{zhao2022graformer} proposed a graph-oriented Transformer which enlarges the receptive field through self-attention and models graph structure by GCN to improve the performance on 3D pose estimation. 

For in-the-wild data, it is difficult to obtain accurate 3D pose annotations. To deal with the lack of 3D pose annotation problem, some weakly supervised, self-supervised, or unsupervised methods~\citep{zhou2017towards,yang20183d,habibie2019wild,chen2019unsupervised,wandt2019repnet,iqbal2020weakly,kundu2020self,schmidtke2021unsupervised,yu2021towards,gong2022posetriplet,chai2023global} were proposed for estimating 3D poses from in-the-wild images without 3D pose annotations. A weakly supervised transfer learning method~\citep{zhou2017towards} was proposed to transfer the knowledge from 3D annotations of indoor images to in-the-wild images. 3D bone length constraint-induced loss was applied in the weakly supervised learning. Habibie et al.~\citep{habibie2019wild} applied a projection loss to refine 3D pose without annotation. A lifting network~\citep{chen2019unsupervised} was proposed to recover 3D poses in a self-supervised mode by introducing a geometrical consistency loss based on the closure and invariance lifting property. The previous self-supervised methods have largely relied on weak supervisions like consistency loss to guide the learning, which inevitably leads to inferior results in real-world scenarios with unseen poses. Comparatively, Gong et al.~\citep{gong2022posetriplet} propose a PoseTriplet method that allows explicit generating 2D-3D pose pairs for augmenting supervision, through a self-enhancing dual-loop learning framework. Benefiting from the reliable 2D pose detection, two-step-based approaches generally outperform one-step-based ones. 
 
\vspace{-0.3cm}
\subsubsection{ Image-based multi-person pose estimation}
\begin{figure*}[t]
	\begin{center}
		{\includegraphics[scale=0.65]{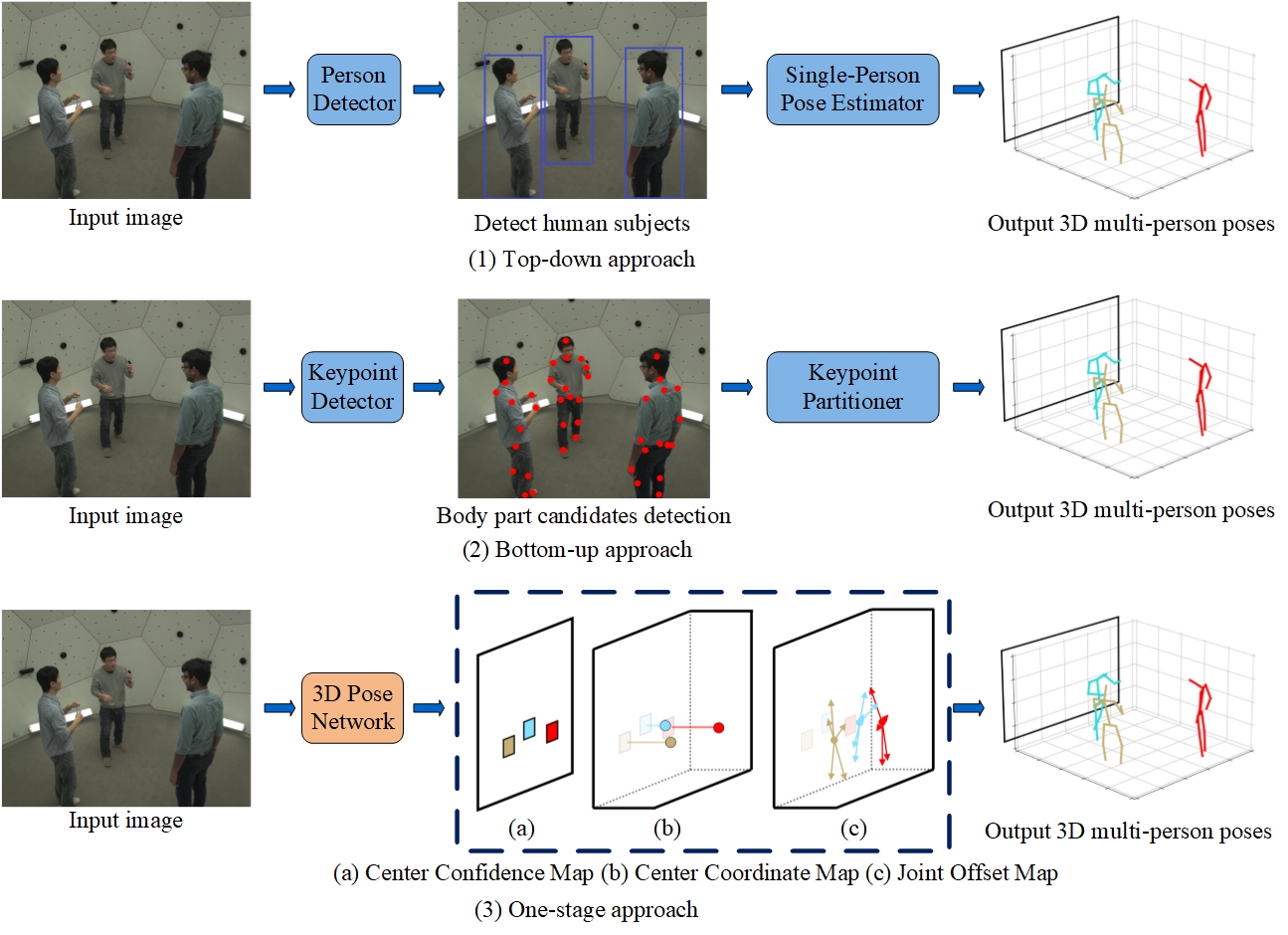}
			\caption{The framework of two approaches for image-based 3D MPPE. Part of the figure is from~\citep{wang2022distribution}.}
			\label{fig:img-MPPE-3D}
		}
	\end{center}
\end{figure*}

Similar to 2D multi-person pose estimation, 3D multi-person pose estimation for images can be also divided into: top-down approaches, bottom-up approaches and one-stage approaches. Top-down and bottom-up approaches involve two stages for pose estimation. Fig.~\ref{fig:img-MPPE-3D} illustrates the general framework of the two approaches for image-based 3D MPPE.

(1) Top-down approach

Top-down approaches first detect each person based on human detection networks and then generate 3D poses based on single-person estimation approaches. Localization Classification-Regression Network (LCR-Net)~\citep{rogez2017lcr,rogez2019lcr} proposes a pose proposal network to generate human bounding boxes and a series of human pose hypotheses. The pose hypotheses were refined based on the cropped ROI features for generating 3D poses. Moon et al.~\citep{moon2019camera} proposed a camera distance-aware method for estimating the camera-centric human poses which consists of human detection, absolute 3D human root localization, and root-relative 3D single-person pose estimation modules. Here, the root-relative poses ignore the absolute locations of each pose. Comparatively, Lin and Lee~\citep{lin2020hdnet} proposed the Human Depth Estimation Network (HDNet) for absolute root joint localization in the camera coordinate space. HDNet could estimate the human depth with considerably high performance based on the prior knowledge of the typical size of the human pose and body joints. The top-down methods mostly estimate poses based on each bounding box, which results in the doubt that the top-down models are not able to understand multi-person relationships and handle complex scenes. To address this limitation, Wang et al.~\citep{wang2020hmor} proposed a hierarchical multi-person ordinal relations (HMOR) to leverage the relationship among multiple persons for pose estimation. HMOR could encode the interaction information as ordinal relations, supervising the networks to output 3D poses in the correct order. Cha et al.~\citep{cha2022multi} designed a transformer-based relation-aware refinement to capture the intra- and inter-person relationships. Although the top-down approaches achieve high accuracy, they suffer high computation costs as person number increases. Meanwhile, these methods may neglect global information (inter-person relationship) in the scene since poses are individually estimated.

(2) Bottom-up approach

Bottom-up approaches first produce all body joint locations and then associate joints to each person according to root depth and part relative depth. Zanfir et al.~\citep{zanfir2018deep} proposed MubyNet to group
human joints according to body part scores based on integrated 2D and 3D information. One group of bottom-up approaches aim to group body joints belonging to each person. Learning on Compressed Output (LoCO) method~\citep{fabbri2020compressed} first applied volumetric heatmaps to produce
joint locations with an encoder-decoder network for feature compression, and a distance-based heuristic was then applied
to retrieve 3D pose for each person. A distance-based heuristic was applied for linking joints. The previous methods are trained in a fully-supervised fashion which requires 3D pose annotations, while Kundu et al. \citep{kundu2020unsupervised} proposed a unsupervised method for 3D pose estimation. Without paired 2D images and 3D pose annotations, a frozen network was applied to exploit the shared latent space between two different modalities based on cross-modal alignment. 

Another group of bottom-up approaches focus on occlusion. Mehta et al.\citep{mehta2018single} combined the joint location maps and the
occlusion-robust pose-maps to infer the 3D poses. The joint location redundancy is applied to infer occluded joints. XNect~\citep{mehta2020xnect} encodes the immediate local context of joints in the kinematic tree to address occlusion. Zhen et al.~\citep{zhen2020smap} developed 3D part affinity field for depth-aware part association by reasoning about inter-person occlusion, and utilized a refined network to refine the 3D pose given predicted 2D and 3D joint coordinates. All of these methods handle occlusion from the perspective of single-person and require initial grouping joints into individuals, which results in error-prone estimates in multi-person scenarios. Liu et al.~\citep{liu2022explicit} proposed an occluded keypoints reasoning module based on a deeply supervised encoder distillation network to reason about the invisible information from the visible ones. Chen et al.~\citep{chen2023multi} presented Articulation-aware Knowledge Exploration (AKE) for keypoints associated with a progressive scheme in the occlusion situation. In comparison to top-down approaches, bottom-up approaches offer the advantage of not requiring repeated single-person pose estimation and they enjoy linear computation. However, the bottom-up approaches require a second association stage for joint grouping. Furthermore, since all persons are processed at the same scale, these methods are inevitably sensitive to human scale variations, which limits their applicability in wild videos.

(3) One-stage approach

One-stage approaches treat pose estimation as parallel human center localizing and center-to-joint regression problem. Instead of separating joints localizing and grouping in the two-stage approaches, these approaches predict each of the joint offsets from the detected center points, which is usually set as the root joint of human. Since the joint offsets are directly correlated to estimated center points, this strategy avoids the manually designed grouping post-processing and is end-to-end trainable. 
Zhou et al.\citep{zhou2019objects} modeled an object as a single point and regressed joints from image features at the human center. Wei et al.~\citep{wei2020point} proposed to regress joints from point-set anchors which serve as prior of basic human poses. Wang et al.~\citep{wang2022distribution} reconstructed joints from 2.5D human centers and 3D center-relative joint offsets. Jin et al.~\citep{jin2022single} proposed a Decoupled Regression Model (DRM) by solving 2D pose regression and depth regression. Recently, Qiu et al.~\citep{qiu2023weakly} estimated 3D poses directly by fine-tuning a Weakly-Supervised Pre-training (WSP) network on 3D pose datasets.

\vspace{-0.3cm}
\subsubsection{Video-based single-person pose estimation}

Instead of estimating 3D poses from images, videos can provide temporal information to improve the accuracy and robustness of pose estimation. Similar to image-based 3D HPE, video-based 3D HPE can also be categorized into one-stage and two-stage approaches. 

(1) One-stage approach

There are few research belong to this category of approaches. Tekin et al.~\citep{tekin2016direct} proposed a regression function to directly predict the 3D pose in a given frame of a sequence from a spatio-temporal volume centered around it. This volume comprises bounding boxes surrounding the person in consecutive frames coming before and after the central one. Mehta et al.~\citep{mehta2017vnect} proposed the VNect, which is capable of obtaining a temporally consistent, full 3D skeletal pose of a human from a monocular RGB camera by Convents regression and kinematic skeleton fitting. The VNect could regress 2D and 3D joint locations simultaneously. Dabral et al.~\citep{dabral2018learning} proposed two structure-aware loss functions: illegal angle loss and left-right symmetry loss to directly predict 3D body pose from the video sequence. The illegal angle loss is to distinguish the internal and external angle of a 3D joint and the symmetry loss is defined as the difference in lengths of left/right bone pairs. Qiu~\citep{qiu2022ivt} proposed an end-to-end framework based on Instance-guided Video Transformer (IVT) to predict 3D single and multiple poses directly from videos. An unsupervised feature extraction method~\citep{9921314} based on Constrastive Self-Supervised (CSS) learning was presented to capture rich temporal features for pose estimation. Time-variant and time-invariant latent features are learned using CSS by reconstructing the input video frames and time-variant features are then applied to predicting 3D poses.

(2) Two-stage approach

Similar to two-step 3D poses estimated from images, two-step 3D HPE involves two stages: estimating 2D poses and lifting 3D poses from 2D poses. However, the difference is that a sequence of 2D poses is applied for lifting a sequence of 3D poses in video-based 3D HPE. Based on different lifting methods, this category of approaches can be summarized into Seq2frame and Seq2seq-based methods.

Seq2frame-based methods pay attention to predicting the central frame of the input video to produce a robust prediction and less sensitivity to noise. Pavllo et al.~\citep{pavllo20193d} presented a Temporal Convolutional Network (TCN) on 2D keypoint trajectories with semi-supervised training method. In the network, 1D convolutions are used to capture temporal information with fewer parameters. In semi-supervised training, the 3D pose estimator is used as the encoder and the decoder maps the predicted pose back to the 2D space. Some following works improved the performance of TCN by solving the occlusion problem~\citep{cheng2019occlusion}, utilizing the attention~\citep{liu2020gast}, or decomposing the pose estimation task into bone length and bone direction prediction~\citep{chen2021anatomy}. Except TCN, Cai et al.~\citep{cai2019exploiting} employs GCN for modeling temporal information in which learning multi-scale features for 3D human body estimation from a short sequence of 2D joint detection. Without convolution architecture involved, Zheng et al.~\citep{zheng20213d} proposed a PoseFormer based on a spatial-temporal transformer for estimating the 3D pose of the center frame. To overcome the huge computational cost of PoseFormer when increasing the frame number for better performance, PoseFormerV2~\citep{zhao2023poseformerv2} applies a frequency-domain representation of 2D pose sequences for lifting 3D poses. Similarly, Li et al.~\citep{li2022exploiting} proposed a stridden transformer encoder to reconstruct 3D pose of the center frame by reducing the sequence redundancy and computation cost. Li et al.~\citep{li2022mhformer} further designed a Multi-Hypothesis transFormer (MHFormer) to exploit spatial-temporal representations of multiple pose hypotheses. Based on MHFormer, MHFormer++~\citep{li2023multi} is proposed to further model local information of joints by incorporating graph Transformer encoder and effectively aggregate multi-hypothesis features by adding a fusion block. With the similar idea of pose hypothesis~\citep{li2022mhformer,li2023multi}, DiffPose~\citep{holmquist2022diffpose} and Diffusion-based 3D Pose (D3DP)~\citep{shan2023diffusionbased} aim to apply a diffusion model to predict multiple adjustable hypotheses for a given 2D pose due to its ability of high-field samples. The aforementioned Transformer-based methods~\citep{zheng20213d,zhao2023poseformerv2,li2022exploiting,li2023multi} mainly model spatial and temporal information sequentially by different stages of networks, thus resulting in insufficient learning of motion patterns. Therefore, Tang et al.~\citep{tang20233d} proposed Spatio-Temporal Criss-cross Transformer (STCFormer) by stacking multiple STC attention blocks to model spatial and temporal information in parallel with a two-pathway network.

Seq2seq-based methods reconstruct all frames of input sequence at once for improving coherence and efficiency of 3D pose estimation. The earlier methods apply recurrent neural network (RNN) or long short-term memory (LSTM) as the Seq2Seq network. Lin et al.~\citep{lin2017recurrent} designed a Recurrent 3D Pose Sequence Machine(RPSM) for estimating 3D human poses from a sequence of images. The RPSM consists of three modules: a 2D pose module; a 3D pose recurrent module and a feature adaption module for transforming the pose representations from 2D to 3D domain. Hossain et al.~\citep{rayat2018exploiting} presented a sequence-to-sequence network by using LSTM units and residual connections on the decoder side. The sequence of 2D joint locations is as input to the sequence-to-sequence network to predict a temporally coherent sequence of 3D poses. Lee et al.~\citep{lee2018propagating} proposed propagating long short-term memory networks (p-LSTMs) to estimates depth information from 2D joint location through learning the intrinsic joint interdependency. Katircioglu et al.~\citep{katircioglu2018learning} proposed a deep learning regression architecture to learn a high-dimensional latent pose representation by using an autoencoder and a Long Short-Term Memory network is proposed to enforce temporal consistency on 3D pose predictions. Raymond et al.~\citep{yeh2019chirality} proposed Chirality Nets. In Chirality Nets, fully connected layers, convolutional layers, batch-normalization, and LSTM/GRU cells can be chiral. According to this kind of symmetry, it naturally estimates 3D pose by exploiting the left/right mirroring of the human body. Later, there are some methods~\citep{wang2020motion,yu2023gla,zhang2022mixste,ijcai2023p65,9815549,zhu2022motionbert} apply GCN or transformer for Seq2seq learning. Wang et al.~\citep{wang2020motion} exploited a GCN-based method combining a corresponding loss to model motion in both short temporal intervals and long temporal ranges. Zhang et al.~\citep{zhang2022mixste} proposed a mixed spatio-temporal encoder(MixSTE) which includes a temporal transformer to model the temporal motion of each joint and a spatial transformer to learn inter-joint spatial correlations. The MixSTE directly reconstructs the entire frames to improve the coherence between input and output sequences. Chen et al.~\citep{ijcai2023p65} proposed High-order Directed Transformer (HDFormer) to reconstruct 3D pose sequences from 2D pose sequences by incorporating self-attention and high-order attention to model joint-joint, bone-joint, and hyperbone-joint interactions.

\vspace{-0.3cm}
\subsubsection{Video-based multi-person pose estimation}
Different from the image-based multi-person pose estimation, video-based multi-person pose estimation often suffers from fast motion, large variability in appearance and clothing, and person-to-person occlusion. A successful approach in this context must be capable of accurately identifying the number of individuals present in each video frame, as well as determining the precise joint locations for each person and effectively associating these joints over time. 

With the improvement of video-based single-person 3D HPE, one method of video-based multi-based 3D HPE is two-step-based method that first detects each person based on human detection networks and then generates 3D poses based on video-based single-person 3D HPE methods. Cheng et al.~\citep{cheng2021graph} proposed a novel framework for integrating graph convolutional network (GCN) and time convolutional network (TCN) to estimate multi-person 3D pose. In particular, bounding boxes are firstly detected for representing humans and 2D poses are then estimated based on the bounding box. The 3D poses for each frame are estimated by feeding 2D poses into joint- and bone-GCNs. The 3D pose sequence is finally fed into temporal TCN to enforce the temporal and human-dynamic constraints. This category of methods applies top-down technique to estimate 3D poses, which rely on detecting each person independently. Therefore, it is likely to suffer from inter-person occlusion and close interactions. To overcome this problem, the same author\citep{cheng2021monocular} later proposed an Multi-person Pose Estimation Integration (MPEI) network by adding a bottom-up branch for capturing global-awareness poses on the same top-down branch as the paper~\citep{cheng2021graph}. The final 3D poses are estimated based on matching the estimated 3D poses from both bottom-up and top-down branches. An interaction-aware discriminator was applied to enforce the natural interaction of two persons. To overcome the occlusion problem, Park et al.~\citep{park2023robust} presented POTR-3D to lift 3D pose sequences by directly processing 2D pose sequences rather than a single frame at a time, and devise a data augmentation strategy to generate occlusion-aware data with devise views. Capturing long-range temporal information normally requires computing on more frames, which results in high computational cost. To cope with this problem, a recent work, TEMporal POse estimation method (TEMPO)~\citep{choudhury2023tempo}, learns a spatio-temporal representation by a recurrent architecture to speed up the inference time while preserving estimation accuracy. To be specific, persons are firstly detected and represented by feature volumes. A spatio-temporal pose representation is then learned by recurrently combining features from current and previous timesteps. It is finally decoded into an estimation of the current pose and poses at future timestaps. Note that the poses are estimated based on the tracking results of feature volumes, which hints that pose estimation performance can be improved by pose tracking. Moreover, TEMPO also provides a solution for action prediction. 

In the above two-step-based methods, the result of the latter step depends on the ones of the former step. Therefore, one-step pose estimation is proposed recently based on end-to-end network. IVT~\citep{qiu2022ivt} can be also used to predict multiple poses directly from videos. The instance-guided tokens include deep features and instance 2D offsets (from body center to keypoints) which are sent into a video transformer to capture the contextual depth information between multi-person joints in spatial and temporal dimensions. A cross-scale instance-guided attention mechanism is introduced to handle the variational scales among multiple persons. 

In summary, 3D HPE has made significant advancements recent years. Due to the progress in 2D HPE, a large number of 3D image/video-based single-person HPE methods apply 2D to 3D lifting strategy. When extending single-person to multi-person in 3D image/video-based HPE, two step (top-down and bottom-up) and one-step methods are always applied. Although top-down methods could achieve promising results by the state-of-the-art person detection and single-person methods, they suffer from high computation cost as person number increases and the missing of inter-person relationship measurement. The bottom-up methods could enjoy linear computation, however, they are sensitive to human scale variations. Therefore, one-step based methods are preferable for 3D image/video-based multi-person HPE. When extending image-based 3D single/multi-person HPE to video-based ones, temporal information is measured for learning joint association across frames. Similar to images-methods, two-step-based methods are commonly used due to the success of 2D to 3D lifting strategy. Among them, Seq2seq-based methods are preferable, as they contribute to enhancing the coherence and efficiency of 3D pose estimation. To capture the temporal information,  TCN
(Temporal Convolutional Networks), RNN (Recurrent Neural Network)-related architectures, and Transformers are commonly used networks.

\section{Pose tracking} 
\label{posetracking}

Pose tracking aims to estimate human poses from videos and link the poses across frames to obtain a number of trackers. It is related to video-based pose estimation, but it requires to capturing the association of estimated poses across frames which is different from video-based pose estimation. With the pose estimation methods reviewed in Section 2, the main task of pose tracking becomes pose linking. The fundamental problem of pose linking is to measure the similarity between pairs of poses in adjacent frames. The pose similarity is normally measured based on temporal information (eg. optical flow, temporal smoothness priors), and appearance information from images. Following the taxonomy of two kinds of estimated poses, we divide the pose tracking methods into two categories: 2D pose tracking and 3D pose tracking. 
\vspace{-0.3cm}
\subsection{2D pose tracking}
According to the number of persons for tracking, 2D pose tracking can be divided into single-person and multi-person pose tracking. Fewer methods solve the problem of single-person pose tracking since they actually aim to update the estimated poses for obtaining more accurate poses with temporal consistency. Therefore, pose tracking mainly solves the tracking problem of multiple persons. Nevertheless, we will give a review of two categories of methods including single-person and multi-person pose tracking.

\vspace{-0.3cm}
\subsubsection{Single-person pose tracking}
\begin{figure*}[t]
	\begin{center}
		{\includegraphics[scale=0.65]{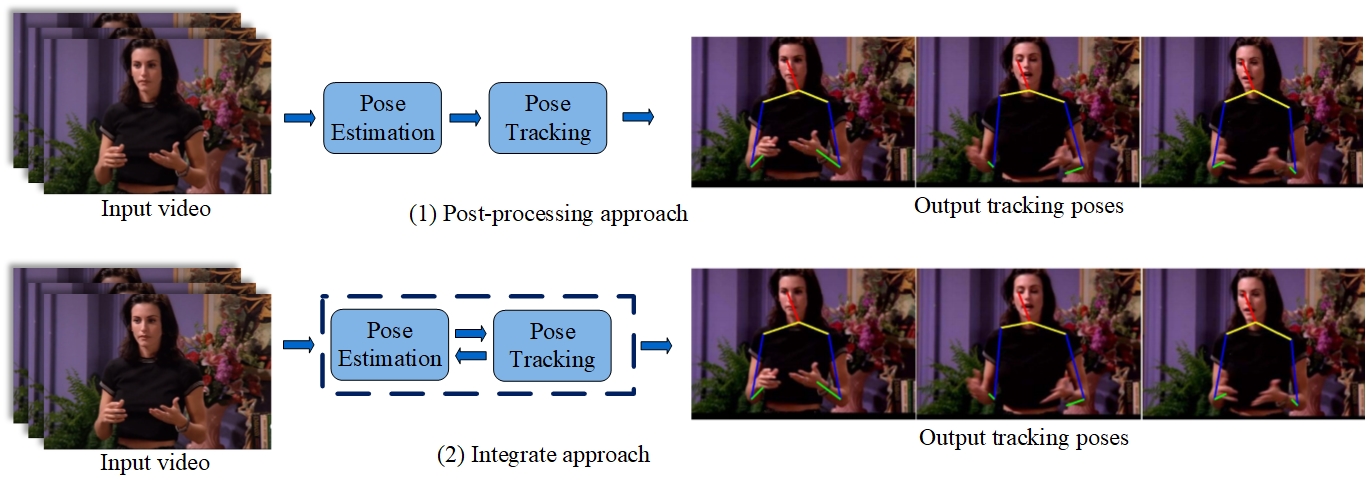}
			\caption{The framework of two approaches for 2D Single person pose tracking.}
			\label{fig:img-SPPT-2D}
		}
	\end{center}
\end{figure*}

Based on the core idea of updating the estimated poses by tracking, this category of approaches can be usually divided into two types, post-processing and integration approaches. The post-processing approaches estimate the pose of each frame individually, and then correlation analysis is conducted on the estimated poses across different frames to reduce inconsistencies and generate a smooth result. The integrated approaches unite pose estimation and visual tracking within a single framework. Visual tracking ensures the temporal consistency of the poses, while pose estimation enhances the accuracy of the tracked body parts. By combining the strengths of both visual tracking and pose estimation, the integrated approaches achieve improved results in pose tracking. Fig.~\ref{fig:img-SPPT-2D} illustrates the general framework of the two approaches for single person pose tracking.

(1) \textbf{Post-processing approach}

Zhao et al.~\citep{zhao2015tracking} proposed to track human body pose by adopting the max-margin Markov model. They proposed a spatio-temporal model composed of two sub-models for spatial parsing and temporal parsing respectively. Spatial parsing is used to estimate candidate human poses in a frame, while temporal parsing determines the most probable pose part locations over time. An inference iteration of sub-models is conducted to obtain the final result. Samanta et al. ~\citep{samanta2016data} proposed a data-driven method for human body pose tracking in video data. They initially estimated the pose in the first frame of the video, and employed local object tracking to maintain spatial relationships between body parts across different frames.

(2) \textbf{Integrated approach}

Zhao et al.~\citep{zhao2015learning} proposed a two-step iterative method that combines pose estimation and visual tracking into a unified framework to compensate for each other, the pose estimation improves the accuracy of visual tracking, and the result of visual tracking facilitates the pose estimation. The two steps are performed iteratively to get the final pose. In addition, they designed a reinitialization mechanism to prevent pose tracking failures. Previous methods required future frames or entire sequences to refine the current pose and were difficult to track online. Ma et al.~\citep{ma2016local} solved the problem of online tracking human pose of joint motion in dynamic environments. They proposed a coupled-layer framework composed of a global layer for pose tracking and a local layer for pose estimation. The core idea is to decompose the global pose candidate in any particular frame into several local part candidates and then recombine selected local parts to obtain an accurate pose for the frame.
 
Post-processing approaches first obtain a set of plausible pose assumptions from the video and then stitch together compatible detections over time to form pose tracking. However, due to the multiplicative cost of using global information, models in this category can usually only include local spatio-temporal trajectories (evidence). These local spatio-temporal trajectories may be ambiguous, thus leading to the disadvantage of objective models. Furthermore, post-processing methods are difficult to track online, but integrated approaches allow for a more robust and accurate representation of the poses over time, ensuring that the tracked body retrains its appropriate configuration throughout the tracking process.

\vspace{-0.3cm}
\subsubsection{Multi-person pose tracking}

Unlike single-person pose tracking, multi-person pose tracking involves measuring human interactions, which can introduce challenges to the tracking process. The number of the tracking people is unknown, and the human interaction may cause the occlusion and overlap. Similar to multi-person pose estimation, existing methods can be divided into two categories, top-down and bottom-up approaches.  

(1) \textbf{Top-down approach}

Top-down approaches~\citep{wang2020combining,fang2022alphapose} start by detecting the overall location and bounding box of the human body in frames and then estimates the keypoints of each person. Finally, the estimated human poses are associated according to similarity between poses in different frames. Girdhar et al.~\citep{girdhar2018detect} proposed a two-stage method for estimating and tracking human keypoints in complex multi-person videos. The method utilizes Mask R-CNN to perform frame-level pose estimation which detects person tubes and estimates keypoints in predicted tubes, then performs a person-level tracking module by using lightweight optimization to connect estimated keypoints over time. However, this method does not consider motion and pose information, which causes difficulty in tracking the occasional truncated human. To address the issue, Xiu et al.~\citep{xiu2018pose} employed pose flow as a unit and proposed a new pose flow generator which consists of Pose Flow Builder and Pose Flow NMS. They initially estimated multi-person poses by employing an improved RMPE, and then maximizing overall confidence to construct pose flows. Finally, pose flows were purified by applying Plow Flow NMS to obtain reasonable multi-pose trajectories. To ease the complexity of method, Xiao et al.~\citep{xiao2018simple} proposed a simple but effective method for pose estimation and tracking. They adopted the pose propagation and similarity measurement based on optical flow to improve the greedy matching method for pose tracking. Zhang et al.~\citep{zhang2019fastpose} solved the articulated multi-person pose estimation and real-time velocity tracking. An end-to-end multi-task network (MTN) was designed for simultaneously performing human detection, pose estimation, and person re-identification (Re-ID) tasks. Given the detection box, keypoints and Re-ID feature provided by MTN, an occlusion-aware strategy is applied for pose tracking. Ning et al.~\citep{ning2020lighttrack} proposed a top-down approach that combines single-person pose tracking (SPT) and visual object tracking (VOT) into a unified online functional entity that can be easily implemented with a replaceable single person pose estimator. They processed each human candidate separately and associated the lost tracked candidate to the targets from the previous frames through pose matching. The human pose matching can be achieved by applying the Siamese Graph Convolution Network as the Re-ID module. Umer et al.~\citep{umer2020self} proposed a method that relies on the correspondence relationship of keypoints to associate the figures in the video. It is trained on large image data sets to use self-monitoring for body pose estimation. In combination with the top-down human pose estimation framework, keypoint correspondence is used to recover lost pose detection based on the temporal context and associate detected and recovered poses for pose tracking. 

The methods discussed in this section typically begin by detecting the human body boundary, which can make them susceptible to challenges like occlusion and truncation. Moreover, most methods first estimate poses in each frame and then implement data association and refinement. This strategy essentially relies heavily on non-existent visual evidence in the case of occlusion, so detection is inevitably easy to miss. To this end, Yang et al.~\citep{yang2021learning} derived dynamic predictions through GNN that explicitly takes into account spatio-temporal and visual information. It leverages historical pose tracklets as input and predicts corresponding poses in
the following frames for each tracklet. The predicted poses will then be aggregated with the detected poses, so as to recover occluded joints that may have been missed by the estimator, significantly improving the robustness of the method.

The methods mentioned above primarily emphasize pose-based similarities for matching, which usually struggle to re-identify tracks that have been occluded for extended periods or significant pose deformations. In light of this, Doering et al.~\citep{doering2023gated} proposed a novel gated attention approach which utilizes a duplicate-aware association, and automatically adapts the impact of pose-based similarities and appearance-based similarities according to the attention probabilities associated with each similarity metric.

(2) \textbf{Bottom-up approach}

In contrast, bottom-up approaches first detect keypoints of the human body and then group the keypoints into individuals. The grouped keypoints are then connected and associated across frames to generate the complete pose. Iqbal et al.~\citep{iqbal2017posetrack} proposed a novel method which jointly models multi-person pose estimation and tracking in a single formula. They represented the detected body joints in the video by a spatio-temporal graph which can be divided into sub-graphs corresponding to the possible trajectories of each human body pose by solving an integer linear program. Raaj et al.~\citep{raaj2019efficient} proposed Spatio-Temporal Affinity Fields (STAF) across a video sequence for online pose tracking. The connections across keypoints in each frame are represented by Part Affinity Fields (PAFs) and connections between keypoints across frames are represented by Temporal Affinity Fields. Jin et al.~\citep{jin2019multi} viewed pose tracking as a hierarchical detection and grouping problem. They proposed a unified framework consisting of SpatialNet and TemporalNet. SpatialNet implements single-frame body part detection and part-level data association, and TemporalNet groups human instances in continuous frames into trajectories. The grouping process is modeled by a differentiable Pose-Guided Grouping (PGG) module to make the entire part detection and grouping pipeline fully end-to-end trainable.

The bottom-up approach relates joints spatially and temporally without detecting bounding boxes. Therefore, the computational cost of the methods is almost unaffected by the change in the number of human candidates. However, they require significant computational resources and often suffers from the ambiguous keypoints assignment without the global pose view. The top-down approach enhances single-frame pose estimation by incorporating temporal context information to correlate estimated poses across different frames. It simplifies the complex task and improves the keypoints assignment accuracy, although it may increase calculation cost in case of a large number of human candidates. In summary, the top-down approach outperforms the bottom-up approach both in accuracy and tracking speed, so most of the state-of-the-art methods follow the top-down approach.
\vspace{-0.3cm}
\subsection{3D pose tracking}
With the advancement of 3D pose estimation, pose tracking can be naturally extended into 3D space. Given that current methods primarily focus on multi-person scenarios, we categorize them into two groups without specifying single or multi-person tracking: multi-stage and one-stage approaches.

(1) Multi-stage approach

The multi-stage approaches generally track poses involving several steps such as 2D/3D pose estimation, lifting 2D to 3D poses and 3D pose linking. These tasks are served as independent sub-tasks. For example, Bridgeman et al.~\citep{bridgeman2019multi} performed independent 2D pose detection per frame and associated 2D pose detection between different camera views through a fast greedy algorithm. Then the associated poses are used to generate and track 3D pose.
Zanfir et al.~\citep{zanfir2018monocular} first conducted a single person feedforward-feedback model to compute 2D and 3D pose, and then performed joint multiple person optimization under constraints to reconstruct and track multiple person 3D pose. Metha et al.~\citep{mehta2020xnect} estimated 2D and 3D pose features and employed a fully-connected neural network to decode features into complete 3D poses, followed by a space-time skeletal model fitting. 

The above works firstly estimate poses and then link poses across frames in which the concept of tracking is to associate joints of the same person together over time, using joints localized independently in each frame. By contrast, Sun et al.~\citep{sun2019explicit} improved joint localization based on the information from other frames. They proposed to first learn the spatio-temporal joint relationships and then formulated pose tracking as a simple linear optimization problem. 

(2) One-stage approach

One-stage approach~\citep{reddy2021tessetrack,zhang2022voxeltrack,choudhury2023tempo,zou2023snipper} aims to train a single end-to-end framework for jointly estimating and linking 3D poses, which can propagate the errors of the sub-tasks in the multi-stage approaches back to the input image pixels of videos. For instance, Reddy et al.~\citep{reddy2021tessetrack} introduced Tessetrack to jointly infer about 3D pose reconstructions and associations in space and time in a single end-to-end learnable framework. Tessetrack consists of three key components: person detection, pose tracking and pose estimation. With the detected persons, a spatial-temporal person-specific representation is learned for measuring similarity to link poses by solving an assignment problem based on bipartite graph matching. All matched representations are then merged into a single representation which is deconvolved into a 3D pose and taken as the estimated pose. To handle the occlusions, VoxelTrack~\citep{zhang2022voxeltrack} introduces an occlusion-aware multi-view feature fusion strategy for linking poses. Specifically, it jointly estimates and tracks 3D poses from a 3D voxel-based representation constructed from multi-view images. Poses are linked over time by bipartite graph matching based on fused representation from different views without occlusion. PHALP~\citep{rajasegaran2022tracking} accumulates 3D representations over time for better tracking. It relies on a backbone for estimating 3D representations for each human detection, aggregating representations over time and forecasting future states, and eventually associating tracklets with detections using predicted representations in a probabilistic framework. Snipper~\citep{zou2023snipper} conducts a deformable attention mechanism to aggregate spatiotemporal information for multi-person 3D pose estimation, tracking, and motion forecasting simultaneously in a single shot. Similar to Snipper, TEMPO~\citep{choudhury2023tempo} performs a recurrent architecture to fuse both spatial and temporal information into a single representation, which enabling pose estimation, tracking, and forecasting from multi-view information without sacrificing efficiency.

Although both approaches have achieved good performance on 3D multi-person pose tracking, for the first approach, solving each sub-problem independently leads to performance degradation. 1) 2D pose estimation easily suffers from noise, especially in the presence of occlusion. 2) The accuracy of 3D estimation depends on the 2D estimates and associations across all views. 3) Occlusion-induced unreliable appearance features impact the accuracy of 3D pose tracking. As a result, the second approach has gained prominence in recent years in 3D multi-person pose tracking.

\section{Action Recognition}
\label{sec:ar}

\begin{figure}[t]
	\begin{center}
		{\includegraphics[scale=0.48]{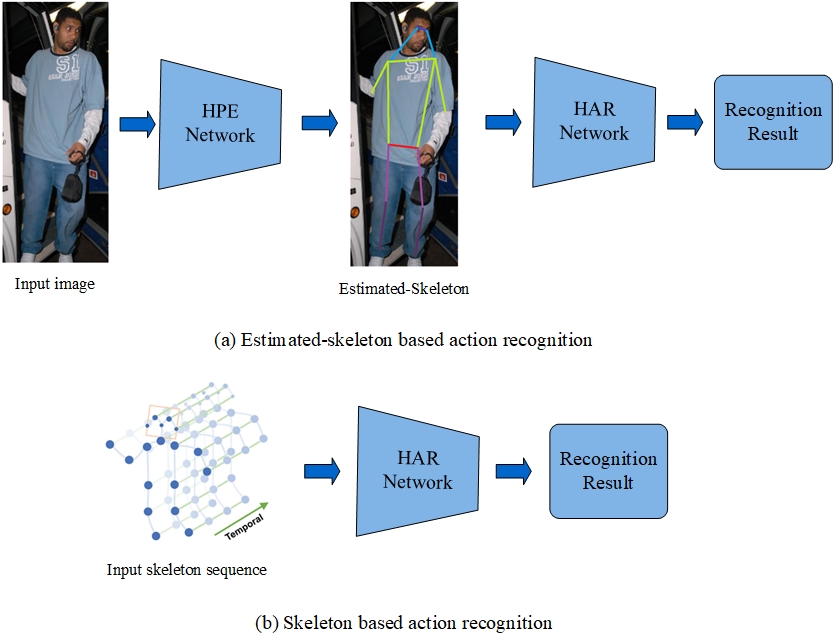}
			\caption{Two categories of approaches for action recognition. }
			\label{fig:img-AR}
		}
	\end{center}
\end{figure}

\begin{figure*}[t]
	\begin{center}
		{\includegraphics[scale=0.6]{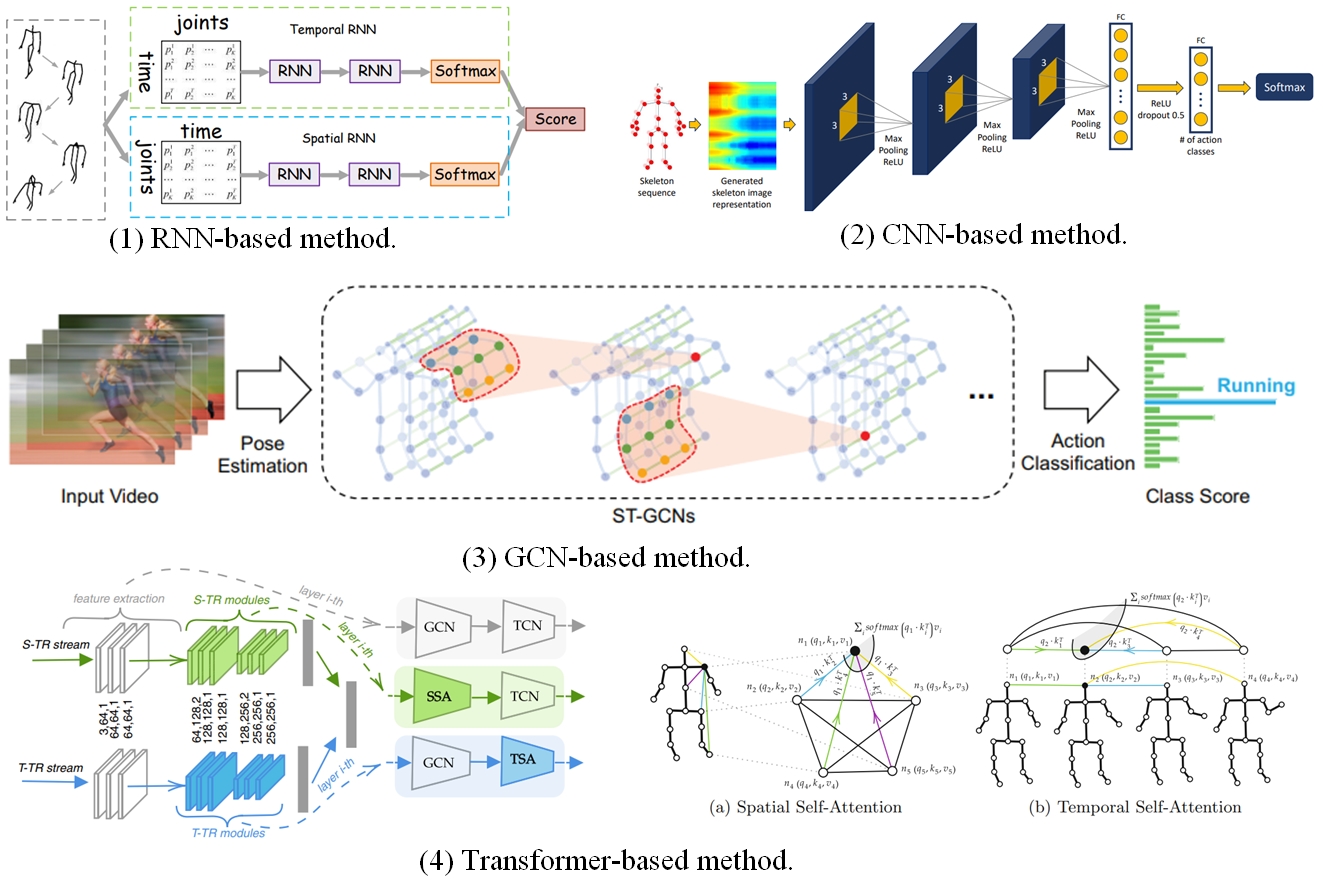}
			\caption{Four approaches for skeleton-based action recognition. (1) RNN example~\citep{wang2017modeling}. (2) CNN  example~\citep{caetano2019skeleton}.
            (3) GCN example~\citep{yan2018spatial}.
            (4) Transformer example~\citep{plizzari2021spatial}.
			\label{fig:skeleton-3D}
		          }
        }
	\end{center}
\end{figure*}

Action recognition aims to identify the class labels of human actions in the input images or videos. For the connection with pose estimation and tracking, this paper only reviews the action recognition methods based on poses. Pose-based action recognition can be categorized into two approaches: estimated pose-based and skeleton-based. Estimated pose-based action recognition approaches apply RGB videos as the input and classify actions using poses estimated from RGB videos. On the other hand, skeleton-based action recognition methods utilize skeletons as their input which can be obtained through various sensors, including motion capture devices, time-of-flight cameras, and structured light cameras. Fig.~\ref{fig:img-AR} illustrate the prevailing frameworks of these two categories approaches of pose-based action recognition.

\vspace{-0.3cm}
\subsection{Estimated pose-based action recognition}
\label{subsec-estimated}

Pose features have been shown in performing much better than low/mid features and acting as discriminative cues for action recognition~\citep{jhuang2013towards}. With the success of pose estimation, some methods follow a two-stage strategy which first applies existing pose estimation methods to generate poses from videos and then conduct action recognition using pose features. Cheron et al.~\citep{cheron2015p} proposed P-CNN to extract appearance and flow features conditioned on estimated human poses for action recognition. Mohammadreza et al.~\citep{zolfaghari2017chained} designed a body part segmentation network to generate poses and then applied it to a multi-stream 3D-CNN to integrate poses, optical flow and RGB visual information for action recognition. After generating joint heatmaps by pose estimator, Choutas et al. ~\citep{choutas2018potion} proposed a Pose moTion (PoTion) representation by temporally aggregating the heatmaps for action recognition. To avoid relying on the inaccurate poses from pose estimation maps, Liu et al.~\citep{liu2018recognizing} aggregated pose estimation maps to form poses and heatmaps, and then evolved them for action recognition. Moon et al.~\citep{moon2021integralaction} proposed an algorithm for a pose-driven approach to integrate appearance and pre-estimated pose information for action recognition. Shah et al.~\citep{shah2022pose} designed a Joint-Motion Reasoning Network (JMRN) for better capturing inter-joint dependencies of poses generated followed by running a pose detector on each video frame. This line of methods considers pose estimation and action recognition as two separate tasks so that action recognition performance may be affected by inaccurate pose estimation. Duan et al.~\citep{duan2022revisiting} proposed PoseConv3D to form 3D heatmap volume by estimating 2D poses by existing pose estimator and stacking 2D heatmaps along the temporal dimension, and to classify actions by 3D CNN on top of the volume. Sato et al.~\citep{Sato_2023_CVPR} presented a user prompt-guided zero-shot learning method based on target domain-independent joint features and the joints are pre-extracted by the existing multi-person pose estimation technique. Rajasegaran et al.~\citep{rajasegaran2023benefits} proposed a Lagrangian Action Recognition with Tracking (LART) method to apply the tracking results for predicting actions. Pose and appearance features are firstly obtained by the PHALP tracking algorithm~\citep{rajasegaran2022tracking}, and then fused as the input of a transformer network to predict actions. Hachiuma et al.~\citep{Hachiuma_2023_CVPR} introduced a unified framework based on structured keypoint pooling for enhancing the adaptability and scalability of skeleton-based action recognition. Human keypoints and object contour points are initially obtained through multi-person pose estimation and object detection. A structured keypoint pooling is then applied to aggregate keypoint features to overcome skeleton detection and tracking errors. Addtionally, non-human object keypoints are severed as additional input for eliminating the variety restrictions of targeted actions. Finally, A pooling-switch trick is proposed for weakly supervised spatio-temporal action localization to achieve action recognition for each person in each frame.

Another line of methods jointly solves pose estimation and action recognition tasks. Luvizon et al.~\citep{luvizon20182d} proposed a multi-task CNN for joint pose estimation from still images and action recognition from video sequences based on appearance and pose features. Due to the different output formats of the pose estimation and the action recognition tasks, Foo et al.~\citep{foo2023unified} designed a Unified Pose Sequence (UPS) multi-task model, which constructs text-based action labels and coordinate-based poses into a heterogeneous output format, for simultaneously processing the two tasks. 

\subsection{Skeleton-based Action Recognition}

Skeleton data is one form of 3D data commonly used for action recognition. It consists of a sequence of skeletons, representing a schematic model of the locations of trunk, head, and limbs of the human body. Compared with another two commonly used data including RGB and depth, skeleton data is robust to illumination change and invariant to camera location and subject appearance. With the development of deep learning techniques, skeleton-based action recognition has transitioned from hand-crafted features to deep learning-based features. This survey mainly reviews the recent methods based on different deep learning networks which can be categorized into CNN-based, RNN-based, GCN-based, and Transformer-based methods, as shown in Fig.~\ref{fig:skeleton-3D}.

\vspace{-0.2cm}
\subsubsection{CNN-based approach}
Convolutional Neural Networks (CNN), widely employed in the realm of computer vision, possess a natural advantage in image feature extraction due to their exceptional local perception and weight-sharing capabilities. Due to the success of CNN in image processing, CNN can better capture spatial information in skeleton sequences. CNN-based methods for skeleton-based action recognition can be categorized into 2D and 3D CNN-based approaches, depending on the type of neural network utilized.

Most of the 2D CNN-based methods~\citep{du2015skeleton,wang2016action,hou2016skeleton,li2017joint,liu2017enhanced,ke2017skeletonnet,caetano2019skeleton,li2019learning} first convert the skeleton sequence into a pseudo-image, in which the spatial-temporal information of the skeleton sequence is embedded in the colors and textures. Du et al.~\citep{du2015skeleton} mapped the Cartesian coordinates of the joints to RGB coordinates and then quantized the skeleton sequences into an image for feature extraction and action recognition. To reduce the inter-articular occlusion due to perspective transformations, some works ~\citep{wang2016action,hou2016skeleton} proposed to encode the spatial-temporal information of skeleton sequences into three orthogonal color texture images. The pair-wise distances between joints on single or multiple skeleton sequences are represented by Joint Distance Map (JDM)~\citep{li2017joint} which is encoded as a color change in the texture image. To explore better spatial feature representations, Ding et al.~\citep{ding2017investigation} encoded the distance, direction and angle of the joints as spatial features into the texture color images. Ke et al.~\citep{ke2017new} proposed to represent segments of skeleton sequences by images and classified actions using a multi-task learning network based on CNN. Similarly, Liang et al.~\citep{liang2019three} applied a multi-tasking learning based on three-stream CNN to encode skeletal fragment features, position and motion information.

When compressing skeleton sequences into images by 2D CNN, it is unavoidable to lose some temporal information. By contrast, 3D CNN-based methods~\citep{liu2017two,hernandez20173d} are more excellent at learning spatio-temporal features. Hernandez et al.~\citep{hernandez20173d} encoded skeleton sequences as stacked Euclidean Distance Matrices (EDM) computed over joints and then performed convolution along time dimension for learning spatiao-temporal dynamics of the data. 

\vspace{-0.2cm}
\subsubsection{RNN-based approach}
RNN-related networks are often used for processing time-series data to effectively capture the temporal information within skeleton sequences. Except for temporal information, spatial information is another important cue for action recognition which may be ignored by RNN-related networks. Some methods focus on solving this problem by spatial division of the human body. For exmaple, Du et al.~\citep{du2015hierarchical,du2016representation} proposed a hierarchical RNN for processing skeleton sequences of five body parts for action recognition. Shahroudy et al.~\citep{shahroudy2016ntu} proposed a Partially-aware LSTM (P-LSTM) for separately modeling skeleton sequences of body parts and classified actions based on the concatenation of memory cells. 

To better focus on the key spatial information in the skeleton data, some methods tend to incorporate attention mechanism. Song et al.~\citep{song2017end} proposed a spatiotemporal attention model using LSTM which includes a spatial attention module to adaptively select key joints in each frame, and a temporal attention module to select keyframes in skeleton sequences. Similarly, Liu et al.~\citep{liu2017global} proposed a cyclic attention mechanism to iteratively enhance the performance of attention for focusing on key joints. The subsequent improvement work by Song et al.~\citep{song2018spatio} used spatio-temporal regularization to encourage the exploration of relationships among all nodes rather than overemphasizing certain nodes and avoided an unbounded increase in temporal attention. Zhang et al.~\citep{zhang2019eleatt} proposed a simple, effective, and generalized Element Attention Gate (EleAttG) to enhance the attentional ability of RNN neurons. Si et al.~\citep{si2019attention} proposed an Attention enhanced Graph Convolutional LSTM (AGC-LSTM) to enhance the feature representations of key nodes.

To simultaneously exploit the temporal and spatial features of skeleton sequences, some methods aim to design spatial and/or temporal networks. Wang et al.~\citep{wang2017modeling} proposed a two-stream RNN for simultaneously learning spatial and temporal relationships of skeleton sequences and enhancing the generalization ability of the model through a skeleton data enhancement technique with 3D transformations. Liu et al.~\citep{liu2016spatio} proposed a spatial-temporal LSTM network, extending the traditional LSTM-based learning into the temporal and spatial domains. Considering the importance of the relationships between non-neighboring joints in the skeleton data, Zhang et al.~\citep{zhang2017geometric} designed eight geometric relational features to model the spatial information and evaluated them in a three-layer LSTM network. Si et al.~\citep{si2018skeleton} proposed a spatial-based Reasoning and Temporal Stack Learning (SR-TSL) novel model to capture high-level spatial structural information within each frame, and model the detailed dynamic information by combining multiple jump-segment LSTMs. 

\vspace{-0.1cm}
\subsubsection{GCN-based approach}
GCN is a recent popular network for skeleton-based action recognition due to the human skeleton is a natural graph structure. Compared with CNN and RNN-based methods, GCN-based methods could better capture the relationship between joints in the skeleton sequence. According to whether the topology (namely vertex connection relationship) is dynamically adjusted during inference, GCN-based methods can be classified into static methods~\citep{yan2018spatial,huang2020spatio,liu2020disentangling,zhang2020context} and dynamic methods~\citep{li2019actional,shi2019two,cheng2020skeleton,korban2020ddgcn,chen2021channel,chi2022infogcn,duan2022dg,wang2022skeleton,9763364,lin2023actionlet,li2022smam,DAI2023109540,9997556,shu2022multi,WU2023109231}.

For static methods, the topologies of GCNs remian fixed during inference. For instance, an early application of graph convolutions, spatial-temporal GCN (ST-GCN)~\citep{yan2018spatial}, is proposed which applies a predefined and fixed topology based on the human body structure. Liu et al.~\citep{liu2020disentangling} proposed a multi-scale graph topology to GCNs for modeling multi-range joint relationships. 

For dynamic methods, the topologies of GCNs are dynamically inferred during inference. Action structure graph convolution network (AS-GCN)~\citep{li2019actional} applies an A-link inference module to capture action-specific correlations. Two-stream adaptive GCN (2s-AGCN)~\citep{shi2019two} and semantics-guided network (SGN)~\citep{zhang2020semantics} enhanced topology learning with self-attention mechanism for modeling correlations between two joints. Although topology dynamic modeling is beneficial for inferring intrinsic relations of joints, it may be difficult to encode the context of an action since the captured topologies are independent of a pose. Therefore, some methods focus on context-dependent intrinsic topology modeling. In Dynamic GCN~\citep{ye2020dynamic}, contextual features of all joints are incorporated to learn the relations of joints. Channel topology refinement GCN (CTR-GCN)~\citep{chen2021channel} focuses on embedding joint topology in different channels, while InfoGCN~\citep{chi2022infogcn} introduces attention-based graph convolution to capture the context-dependent topology based on the latent representation learned by information bottleneck. Multi-Level Spatial-Temporal excited Graph Network (ML-STGNet)~\citep{9997556} introduces a spatial data-driven excitation module based on Transformer to learn joint relations of different samples in a data-dependent way. Multi-View Interactional Graph Network (MV-IGNet)~\citep{9234715} designs a global context adaptation module for adaptive learning of topology structures on multi-level spatial skeleton contexts. Spatial Graph Diffusion Convolutional (S-GDC) network~\citep{10023982} aims to learn new graphs by graph diffusion for capturing the connections of distant joints on the same body and two interacting bodies. In the above dynamic methods, the topology modeling is based only on joint information. By contrast, a language model knowledge-assisted GCN (LA-GCN)~\citep{xu2023language} applies large-scale language model to incorporate action-related prior information to learn topology for action recognition. 

No matter the static or dynamic methods, they aim to construct different GCNs for modeling spatial and temporal features of actions. In contrast, some papers work on strategies to assist the ability of different GCNs. For instance, Wang et al.~\citep{Wang_2023_CVPR} proposed neural Koopman pooling to replace the temporal average/max pooling for aggregating spatial-temporal features. The Koopman pooling learns class-wise dynamics for better classification. Zhou et al.~\citep{Zhou_2023_CVPR} presented a Feature Refinement head (FR Head) based on contrastive learning to improve the discriminative power of ambiguous actions. With the FR Head, the performance of some existing methods (eg. 2s-AGCN~\citep{shi2019two}, CTR-GCN~\citep{chen2021channel}) can be improved by about 1\%.

In summary, GCN-based methods can effectively utilize and handle the joint relations by topological networks but are generally limited to local spatial-temporal neighborhoods. Compared with static methods, dynamic methods have stronger generalization capabilities due to the dynamic topologies. 

\vspace{-0.3cm}
\subsubsection{Transformer-based approach}

Transformer was originally designed for machine translation tasks in natural language processing. Vision Transformer (ViT)~\citep{dosovitskiy2020image}  is the first work to use a Transformer encoder to extract image features in computer vision. When introducing Transformer to skeleton-based action recognition, the core is how to design a better encoder for modeling spatial and temporal information of skeleton sequences. Compared with GCN-methods, Transformer-based methods can quickly obtain global topology information and enhance the correlation of non-physical joints. There are mainly three categories of methods: pure Transformer, hybid Transformer and unsupervised Transformer.

The first category of methods applies the standard Transformer for learning spatial and temporal features. A spatial Transformer and a temporal Transformer are often applied alternately or together based on one stream~\citep{shi2020decoupled,wang2021iip,ijaz2022multimodal} or two-stream~\citep{zhang2021stst,shi2021star,gedamu2023relation} network. Shi et al.~\citep{shi2020decoupled} proposed to decouple the data into spatial and temporal dimensions, where the spatial and temporal streams respectively include motion-irrelevant and motion-relevant features. A Decoupled Spatial-Temporal Attention Network (DSTA-Net) was proposed to encode the two streams sequentially based on the attention module. It allows modeling spatial-temporal dependencies between joints without the information about their positions or mutual connections. Ijaz et al.~\citep{ijaz2022multimodal} proposed a multi-modal Transformer-based network for nursing activity recognition which fuses the encoding results of the spatial-temporal skeleton model and acceleration model. The spatial-temporal skeleton model comprises of spatial and temporal Transformer encoder in a sequential processing, which computes spatial and temporal features from joints. The acceleration model has one Transformer block, which computes correlation across acceleration data points for a given action sample. Zhang et al.~\citep{zhang2021stst} proposed a Spatial-Temporal Special Transformer (STST) to capture skeleton sequences in the temporal and spatial dimensions separately. STST is a two-stream structure including a spatial transformer block and a directional temporal transformer block. Relation-mining Self-Attention Network (RSA-Net)~\citep{gedamu2023relation} applies seven RSA bolcks in spatial and temporal domains for learning intra-frame and inter-frame action features. 
Such a two-stream structure leads to the extension of the feature dimension and makes the network capture richer information, but at the same time increases the computational cost. To reduce the computational cost, Shi et al.~\citep{shi2021star} proposed a Sparse Transformer-based Action Recognition (ST-AR) model. ST-AR consists of a sparse self-attention module performed on sparse matrix multiplications for capturing spatial correlations, and a segmented linear self-attention module processed on variable lengths of sequences for capturing temporal correlations to further reduce the computation and memory cost. 

Since Transformer is weak in extracting discriminative information from local features and short-term temporal information, the second category of methods~\citep{plizzari2021spatial,zhou2022hypergraph,qiu2022spatio,kong2022mtt,zhang2022zoom,gao2022focal,liu2022graph,pang2022igformer,Wang_2023_CVPR,duan2023skeletr} integrate Transformer with GCN and CNN for better feature extraction, which is beneficial to utilize the advantages of different networks. Plizzari et al.~\citep{plizzari2021spatial} proposed a two-stream Spatial-Temporal TRansformer network (ST-TR) by integrating spatial and temporal Transformers with Temporal Convolution Network and GCN. Qiu et al.~\citep{qiu2022spatio} proposed a Spatio-Temporal Tuples Transformer (STTFormer) which includes a spatio-temporal tuples self-attention module for capturing joint relationship in consecutive frames, and an Inter-Frame Feature Aggregation (IFFA) module for enhancing the ability to distinguish similar actions. Similar to ST-TR, the IFFA module applies TCN to aggregate features of sub-actions. Yang et al.~\citep{zhang2022zoom} presented Zoom-Former for extending single-person action recognition to multi-person group activities. The Zoom-Former improves the traditional GCN by designing a Relation-aware Attention mechanism, which comprehensively leverages the prior knowledge of body structure and the global characteristic of human motion to exploit the multi-level features. With this improvement, Zoom-Former could hierarchically extract the low-level motion information of a single person and the high-level interaction information of multiple people. To effectively capture the relationship between key local joints and global contextual information in the spatial and temporal dimension, Gao et al.~\citep{gao2022focal} proposed an end-to-end Focal and Global Spatial-Temporal transFormer (FG-STForm) by integrating temporal convolutions into a global self-attention mechanism. Liu et al.~\citep{liu2022graph} proposed a Kernel Attention Adaptive Graph Transformer Network to use a graph transformer operator for modeling higher-order spatial dependencies between joints. Wang et al.~\citep{Wang_2023_CVPR} proposed a Multi-order Multi-mode Transformer (3Mformer) by applying a higher-order Transformer to process hypergraphs of skeleton data for better capturing higher-order motion patterns between body joints. SkeleTR~\citep{duan2023skeletr} initially employs a GCN to capture intra-person dynamic information and then applies a stacked Transformer encoder to model the person interaction. It can handle different tasks including video-level action recognition, instance-level action detection and group activity recognition.

To improve the generalization ability of features, the third category of methods~\citep{kim2022global,dong2023hierarchical,shah2023halp,cheng2021motion,10222534,ijcai2023p95} focus on unsupervised or self-supervised action recognition based on Transformer which has demonstrated excellent performance in capturing global context and local joint dynamics. These methods normally apply contrastive learning or Encoder-Decoder architecture for learning a better representation of actions. Kim et al.~\citep{kim2022global} proposed GL-Transformer, which designs a global and local attention mechanism to learn the local joint motion changes and global contextual information of skeleton sequences. With the motion sequence representation, actions are classified based on their average pooling on the temporal axis. Anshul et al.~\citep{shah2023halp} designed the HaLP module by generating hallucinating latent positive samples for self-supervised learning based on contrastive learning. This module can explore the potential space of human postures in the appropriate directions to generate new positive samples, and optimize the solution efficiency by a new approximation function. 

In summary, the research on skeleton-based action recognition has made great progress in recent years. CNN-based methods mainly convert skeleton sequences into images, excelling at capturing spatial information of actions but potentially losing temporal information. With the help of RNN for representing temporal information, RNN-based methods focus on representing spatial information based on the spatial division of the human body combining attention mechanism. 
Compared with CNN and RNN-based methods, GCN and Transformer-based methods have greater advantages and become the mainstream methods. GCN-based methods are beneficial for representing joint relations by topological networks in which dynamic topology-based methods have stronger generalization ability than static ones. However, they are mostly confined to local spatial-temporal neighborhoods. Transformer-based methods can quickly obtain global topology information and enhance the correlation of non-physical joints. Combining Transformers with CNN and GCN represents a promising approach for extracting both local and global features, enhancing action recognition performance.

\section{Benchmark datasets}
	\label{datasets}
This section reviews the commonly used datasets for the three tasks and also compares the performance of different methods on some popular datasets.

\begin{table*}[h!]
	\centering
	\caption{Datasets for 2D HPE. PCP: Percentage of Correct Localized Parts, PCPm: Mean Percentage of Correctly Localized Parts, PCK: Percentage of Correct Keypoints, PCKh: Percentage of Correct Keypoints with a specified head size, AP: Average Precision, mAP: mean Average Precision. IB: Image-based, VB: Video-based. SP: single person, MP: multi-person. Train, Val and Test represent frame numbers except for Penn Action and PoseTrack, and they represent video numbers.}
	\label{tab-2dhpe}
	\resizebox{\textwidth}{!}
	{\begin{tabular}{clllllllllll}    
			\hline \textbf{} & \textbf{Dataset} & \textbf{Year}& \textbf{Citation}& \textbf{\#Poses} & \textbf{\#Joints} & \textbf{Train}&\textbf{Val} &\textbf{Test} & \textbf{SP/MP} &Actions& \textbf{Metrics} \\ \hline
			
			\multirow{9}{*}{IB}&LSP~\citep{johnson2010clustered} & 2010&971 &2,000& {14} & {1k}&-&1k & {SP}&$\times$ & {PCP/PCK} \\ 
			&\tabincell{c}{LSPET~\citep{johnson2011learning}} & 2011&509 &10,000& {14} & {10k}&-& -& {SP} &$\times$& {PCP} \\ 
			&\tabincell{c}{FLIC~\citep{sapp2013modec}} & 2013&537&5,003& {10} & {4k}&-&1k & {SP}&$\times$ & {PCK/PCP} \\ 
			&\tabincell{c}{MPII~\citep{andriluka20142d}} & 2014&2583 &26,429& {16} & {29k}&-&12k & {SP}&\checkmark & {PCPm/PCKh} \\ 
			&\tabincell{c}{MPII multi-person~\citep{andriluka20142d}} & 2014&2583 & 14,993&{16} & {3.8k}&-&1.7k & {MP}&\checkmark & {mAP} \\
			&\tabincell{c}{MSCOCO16~\citep{lin2014microsoft}} & 2014&37862 &105,698& {17} &45k&22k&80k & {MP}&$\times$ & {AP} \\ 
			&\tabincell{c}{MSCOCO17~\citep{lin2014microsoft}} & 2014&37862 &-& {17} & 64k&2.7k&40k &{MP}&$\times$ & {AP} \\ 
            &\tabincell{c}{LIP~\citep{gong2017look}} & 2017&482 &50462 & {16} & {30k}&10k&10k & {SP}&$\times$ & {PCK} \\ 
			&\tabincell{c}{CrowdPose~\citep{li2019crowdpose}} & 2019&423&80000& {14} & {10k}&2k&8k & {MP}&$\times$ & {mAP} \\ 
			\cmidrule(lr){1-12}
			\multirow{3}{*}{VB}&\tabincell{c}{J-HMDB~\citep{jhuang2013towards}} & 2013&849 &31,838& {15} & {2.4k}&-&0.8k & {SP}&\checkmark & {PCK} \\ 
			&\tabincell{c}{Penn Action~\citep{zhang2013actemes}} & 2013&367&159,633& {13} & {1k}&-&1k & {SP}&\checkmark & {PCK} \\ 
			&\tabincell{c}{PoseTrack17~\citep{andriluka2018posetrack}} & 2017&420&153,615& {15} & 292&50&208 & {MP}&\checkmark & {mAP} \\  
   			&\tabincell{c}{PoseTrack18~\citep{andriluka2018posetrack}} & 2018&420&-& {15} & 593&170&375 & {MP}&\checkmark & {mAP} \\ 
         	&\tabincell{c}{PoseTrack21~\citep{doering2022posetrack21}} & 2022&15&-& {15} & 593&170&- & {MP}&\checkmark & {mAP} \\
      \hline
			\end{tabular}}
\end{table*}

\begin{table*}[h!]
	\centering
	\caption{Datasets for 3D HPE. MPJPE:Mean Per Joint Position Error, PA-MPJPE: Procrustes Analysis Mean Per Joint Position Error, MPJAE: Mean Per Joint Angular Erro, 3DPCK: 3D Percentage of Correct Keypoints, MPJAE: Mean Per Joint Angular Error, AP: Average Precision. }    
	\label{tab-3dhpe}
	\resizebox{\textwidth}{!}
	{
		\begin{tabular}{cllllllll}    
			\hline 
			&\textbf{Dataset} & \textbf{Year} & \textbf{Citation} & \textbf{\#Joints} & \textbf{\#Frames} & \textbf{SP/MP} &Actions& \textbf{Metrics} \\ \hline   
			
			\multirow{6}{*}{VB}&\tabincell{c}{HumanEva-I~\citep{sigal2010humaneva}} & 2010&1678 & {15} & {37.6k} & {SP}   &\checkmark &{MPJPE/PA-MPJPE} \\ 
            &\tabincell{c}{Human3.6M~\citep{ionescu2013human3}} & 2014&2677 & {17} & {3.6M} & {SP}  &\checkmark& {MPJPE} \\ 
			
			&\tabincell{c}{MPI-INF-3DHP~\citep{mehta2017monocular}} & 2017&851 & {15} & {1.3M} & {SP}  &\checkmark& {3DPCK} \\
			&\tabincell{c}{CMU Panoptic~\citep{joo2017panoptic}} & 2017&680& {15} & {1.5M } & {MP} &\checkmark& {3DPCK/MPJPE} \\ 
            &\tabincell{c}{3DPW~\citep{von2018recovering}} & 2018&674 & {18} & {51k } & {MP} &$\times$& {MPJPE/MPJAE/PA-MPJPE} \\
			&\tabincell{c}{MuPoTs-3D~\citep{mehta2018single}} & 2018&346& 15 & {8k} & {MP}  &$\times$ & {3DPCK} \\
            &\tabincell{c}{MuCo-3DHP~\citep{mehta2018single}} & 2018&346& - & - & {MP}  &$\times$ & 3DPCK \\\hline
	\end{tabular}}
\end{table*}

\vspace{-0.3cm}
\subsection{Pose estimation}
The datasets are reviewed based on 2D and 3D pose estimation tasks and the details are summarized in Table~\ref{tab-2dhpe} and \ref{tab-3dhpe}. Due to the page limit, we mainly review some popular and large-scale pose datasets in the following sections.

\vspace{-0.3cm}
\subsubsection{Datasets for 2D pose estimation}
\label{dataset-2DHPE}
For the image-based 2D pose estimation, Microsoft Common Objects in Context (COCO)~\citep{lin2014microsoft} and Max Planck Institute for Informatics (MPII)~\citep{andriluka20142d} are popular datasets. Joint-annotated HMDB (J-HMDB) dataset~\citep{jhuang2013towards} and Penn Action~\citep{zhang2013actemes} datasets are often used for the 2D video-based single-person pose estimation (SPPE), while PoseTrack~\citep{andriluka2018posetrack} is often used for video-based multiple-person pose estimation (MPPE). 

\textbf{The COCO dataset}~\citep{lin2014microsoft} is the most widely used large-scale dataset for pose estimation. It was created by extracting everyday scene images with common objects and labeling the objects using per-instance segmentation. This dataset consists of more than 330,000 images and 200,000 labeled persons, and each person is labeled with 17 keypoints. It has two versions for pose estimation including COCO2016 and COCO2017. The two versions are different with the number of images for training, testing and validation as shown in Table~\ref{tab-2dhpe}. Except of pose estimation, this dataset can be also suitable for object detection, image segmentation and captioning.

\textbf{The MPII dataset}~\citep{andriluka20142d} was collected from 3,913 YouTube videos by the Max Planck Institute for Informatics. It consists of 24,920 images including over 40,000 individuals with 16 annotated body joints. These images were collected by a two-level hierarchical method to capture everyday human activities. This dataset involves 491 activity samples in 21 classes and all the images are labeled. Except for joints, rich annotations including body occlusion, 3D torso and head orientations are also labeled on Amazon Mechanical Turk.  The MPII dataset serves as a valuable resource for both 2D single-person and multi-person pose estimation.

\textbf{The J-HMDB dataset} ~\citep{jhuang2013towards} was created by annotating human joints of the HMDB51 action dataset. From HMDB51, 928 videos including 21 actions of a single person were extracted and the human joints of each were annotated using a 2D articulated human puppet model. Each video consists of 15-40 frames. In total, there are 31,838 annotated frames. This dataset can serve as a benchmark for human detection, pose estimation, pose tracking and action recognition. It also presents a new challenge for video-based pose estimation or tracking since it includes more variations in camera motions, motion blur and partial or full-body visibility. \textbf{Sub-J-HMDB dataset}~\citep{jhuang2013towards} is a subset of the J-HMDB dataset and contains 316 videos with a total of 11,200 frames.

\textbf{The Penn Action dataset}~\citep{zhang2013actemes} is also an annotated sports action dataset collected by the University of Pennsylvania. It consists of 2,326 videos with 15 actions and each frame was annotated with 13 keypoints for each person. The dataset can be used for the tasks of pose estimation, action detection and recognition.

\textbf{The PoseTrack Dataset}~\citep{andriluka2018posetrack} was collected from raw videos of the MPII Pose Dataset. For each frame in MPII, 41-298 neighboring frames with crowded scenes and multiple individuals were selected for PoseTrack dataset. The selected videos were annotated with person locations, identities, body pose and ignore regions. According to different number of videos, this dataset currently exists in three versions: PoseTrack2017, PoseTrack2018, and PoseTrack2021. In total, PoseTrack2017 contains 292 videos for training, and 50 videos for validation and 208 videos for testing. Among them, 23,000 frames are labeled with a very lager number (i.e. 153,615) of annotated poses. PoseTrack2018 increases the number of the video and contains 593 videos for training, 170 videos for validation, and 315 videos for testing, and consists of 46,933 labeled frames. PoseTrack2021 is an extension of PoseTrack2018 with more annotations (eg. bounding box of small persons, joint occlusions). With the person identities, this dataset has been widely used as a benchmark to evaluate multi-person pose estimation and tracking algorithms.

\begin{table*}[h!]
	\centering
	\caption{Performance comparison for 2D image-based pose estimation on COCO dataset.
	}
	\label{tab-2D-image-single&multi}
	\resizebox{\textwidth}{!}
	{
		\begin{tabular}{c c c c c c c c c c c}\hline 
			
			&\multirow{2}{*}{Category}&\multirow{2}{*}{Year}&\multirow{2}{*}{Method}&\multicolumn{7}{@{}c@{}}{\textbf{COCO}}
			\\ 
            \cmidrule(lr){5-11}
			\textbf{}& \textbf{} & \textbf{} & \textbf{}& 
			\textbf{Backbone}& \textbf{Inputsize}& \textbf{AP} & \textbf{AP.5} & \textbf{AP.75}  & \textbf{APM} & \textbf{APL}
			
			\\ \hline 
			
			\multirow{7}{*}{SP}&\multirow{3}{*}{Regression-based}&2021&TFPose~\citep{mao2021tfpose}&ResNet-50&384×288&72.2&90.9&80.1&69.1&78.8
			\\&&2021&PRTR~\citep{li2021pose} &HRNet-W32&512×384&72.1&90.4&79.6&68.1&79.4
			\\&&2022&Panteleris et al.~\citep{panteleris2022pe} &-&384×288&72.6&-&-&-&-
			\\
            \cmidrule(lr){2-11}&\multirow{3}{*}{Heatmap-based}&2021&Li et al.~\citep{li2021human} &HRNet-W48&-&75.7&92.3&82.9&72.3&81.3
            \\&&2022&Li et al.~\citep{li2022simcc} &HRNet-W48&384×288&76.0&92.4&83.5&72.5&81.9
			\\&&2023&DistilPose~\citep{ye2023distilpose} &HRNet-W48-stage3&256×192&73.7&91.6&81.1&70.2&79.6
			\\\cmidrule(lr){1-11}\multirow{33}{*}{MP}&\multirow{19}{*}{Top-down}&2017&Papandreou et al.~\citep{papandreou2017towards} &ResNet-101&353×257&68.5&87.1&75.5&65.8&73.3
			\\&&2017&RMPE~\citep{fang2017rmpe} &Hourglass&-&61.8&83.7&69.8&58.6&67.6
			\\&&2018&Xiao et al.~\citep{xiao2018simple} &ResNet-152&384×288&73.7&91.9&81.1&70.3&80.0
			\\&&2018&CPN~\citep{chen2018cascaded} &ResNet&384×288&73.0&91.7&80.9&69.5&78.1
			\\&&2019&Posefix~\citep{moon2019posefix} &ResNet-152&384×288&73.6&90.8&81.0&70.3&79.8
			\\&&2019&Sun et al.~\citep{sun2019deep} &HRNet-W48&384×288&77&\underline{92.7}&84.5&73.4&83.1
			\\&&2019&Su et al.~\citep{Su_2019_CVPR} &ResNet-152&384×288&74.6&91.8&82.1&70.9&80.6
           \\
            &&2020&Cai et al.~\citep{cai2020learning} &4×RSN-50&384×288&\underline{78.6}&94.3&\underline{86.6}&\underline{75.5}&83.3
			\\&&2020&Huang et al.~\citep{huang2020devil} &HRNet&384×288&77.5&\underline{92.7}&84.0&73.0&82.4
			\\&&2020&Zhang et al.~\citep{zhang2020distribution} &HRNet-W48&384×288&77.4&92.6&84.6&73.6&\underline{83.7}
			\\&&2020&Graphpcnn~\citep{wang2020graph} &HR48&384×288&76.8&92.6&84.3&73.3&82.7
			\\&&2020&Qiu et al.~\citep{qiu2020peeking} &-&384×288&74.1&91.9&82.2&-&-
			\\&&2021&TransPose~\citep{yang2021transpose} &HRNet-W48&256×192&75.0&92.2&82.3&71.3&81.1
			\\&&2021&TokenPose~\citep{li2021tokenpose} &-&384×288&75.9&92.3&83.4&72.2&82.1
			\\&&2021&HRFormer~\citep{YuanFHLZCW21} &-&384×288&76.2&\underline{92.7}&83.8&72.5&82.3
			\\&&2022&ViTPose~\citep{xu2022vitpose} &ViTAE-G&576×432&\textbf{81.1}&\textbf{95.0}&\textbf{88.2}&\textbf{77.8}&\textbf{86.0}
			\\&&2022&Xu et al.~\citep{xu2022adaptive} &HR48&384×288&76.6&92.4&84.3&73.2&82.5
			\\
           
            &&2023&PGA-Net~\citep{JIANG2023109429} &HRNet-W48&384x288&76.0&92.5&83.5&72.4&82.1
			\\
            &&2023&BCIR~\citep{10093110} &HRNet-W48&384x288&76.1&-&-&-&-
			\\
            \cmidrule(lr){2-11}&\multirow{10}{*}{Bottom-up}&2017&Associative embedding~\citep{newell2017associative} &Hourglass&512×512&65.5&86.8&72.3&60.6&72.6
			\\&&2018&Multiposenet~\citep{kocabas2018multiposenet} &ResNet50&480×480&69.6&86.3&76.6&65.0&76.3
			\\&&2018&OpenPose~\citep{cao2018realtime} &-&-&61.8&84.9&67.5&57.1&68.2
			\\&&2019&Pifpaf~\citep{Kreiss_2019_CVPR} &ResNet50&-&55.0&76.0&57.9&39.4&76.4
			\\&&2020&Jin et al.~\citep{jin2020differentiable}&Hourglass&512×512&67.6&85.1&73.7&62.7&74.6
			\\&&2020&Higherhrnet~\citep{cheng2020higherhrnet} &HrHRNet-W48&640×640&72.3&91.5&79.8&67.9&78.2
			\\
            &&2021&DEKR~\citep{geng2021bottom} &HRNet-W48&640x640&71.0&89.2&78.0&67.1&76.9
			\\
            &&2023&HOP~\citep{qu2023characteristic} &HRNet-W48&640×640&70.5&89.3&77.2&66.6&75.8
			\\
            &&2023&Cheng et al.~\citep{CHENG2023109403} &HRNet-W48&640×640&71.5&89.1&78.5&67.2&78.1
			\\&&2023&PolarPose~\citep{10034548} &HRNet-W48&640x640&70.2&89.5&77.5&66.1&76.4
			\\
            
            \cmidrule(lr){2-11}&\multirow{7}{*}{One-stage}&2019&Directpose~\citep{tian2019directpose} &ResNet-101&800×800&64.8&87.8&71.1&60.4&71.5
			\\&&2021&FCPose~\citep{mao2021fcpose} &DLA-60&736 × 512&65.9&89.1&72.6&60.9&74.1
			\\&&2021&InsPose~\citep{shi2021inspose} &HRNet-w32&-&71.0&91.3&78.0&67.5&76.5
			\\&&2022&PETR~\citep{shi2022end} &Swin-L&-&71.2&91.4&79.6&66.9&78.0
			\\&&2023&ED-pose~\citep{yang2023explicit} &Swin-L&-&72.7&92.3&80.9&67.6&80.0
			\\&&2023&GroupPose~\citep{liu2023group} &Swin-L&-&72.8&92.5&81.0&67.7&80.3
			\\&&2023&SMPR~\citep{MIAO2023109743} &HRNet-w32&800x800&70.2&89.7&77.5&65.9&77.2
			\\
			\hline
			
		\end{tabular}
	}
	
\end{table*}

\begin{table}[h!]
	\centering
	\caption{Performance comparison for 2D video-based SPPE on Penn Action dataset and JHMDB dataset. FF: frame-by-frame; SF: sample frame-based.}
	\label{tab-2D-video-single}
 \footnotesize{
	\setlength{\tabcolsep}{0.2mm}{
		\begin{tabular*}{\linewidth}{@{\extracolsep{\fill}}lcccc}\hline 
			
			\multirow{2}{*}{Category}&\multirow{2}{*}{Year}&\multirow{2}{*}{Method}&\textbf{Penn}&\textbf{JHMDB}
			\\ \cmidrule(lr){4-5}
			&  & & \textbf{PCK}& \textbf{PCK}\\	
			\hline 
	\multirow{8}{*}{FF}&2016&Gkioxari et al.~\citep{gkioxari2016chained}&91.8&-\\
            &2017&Song et al.~\citep{song2017thin}&96.4&92.1\\
			&2018&LSTM~\citep{luo2018lstm}&97.7&93.6\\
			&2019&DKD~\citep{nie2019dynamic}&97.8&94\\
			&2019&Li et al.~\citep{li2019temporal}&-&94.8\\
			&2022&RPSTN~\citep{dang2022relation}&\textbf{98.7}&97.7\\
               &2023&HANet~\citep{jin2023kinematic}&-&\textbf{99.6}\\ 
             \cmidrule(lr){2-5}
			\multirow{4}{*}{SF}&2020&K-FPN~\citep{zhang2020key}&98&94.7\\
			&2022&REMOTE~\citep{ma2022remote}&\underline{98.6}&95.9\\
			&2022&DeciWatch~\citep{zeng2022deciwatch}&-&98.9\\
            &2023&MixSynthFormer~\citep{sunmixsynthformer}&-&\underline{99.3}\\

			\hline
		\end{tabular*}}
  }
  \vspace{-0.3cm}
\end{table}

 
\begin{table}[h!]
	\centering
	\caption{Performance comparison for 2D video-based MPPE on PoseTrack2017 dataset.}
	\label{tab-2D-video-multi}
 \footnotesize{
	\setlength{\tabcolsep}{0.2mm}
	{
 
		\begin{tabular*}{\linewidth}{@{\extracolsep{\fill}}l c c c c}\hline 
			
			\multirow{2}{*}{Category}&\multirow{2}{*}{Year}&\multirow{2}{*}{Method}&\textbf{Val}&\textbf{Test}
			\\ \cmidrule(lr){4-5}
			\textbf{} & \textbf{} & \textbf{}& \textbf{mAP}&\textbf{mAP}
			\\ \hline 
			
			\multirow{7}{*}{Top-down}&2018&Xiao et al.~\citep{xiao2018simple}&76.7&\textbf{73.9}\\
			&2018&Pose Flow~\citep{xiu2018pose}&66.5&63.0\\
			&2018&Detect-Track~\citep{girdhar2018detect}&-&64.1\\
			&2020&Wang et al.~\citep{wang2020combining}&81.5&\underline{73.5}\\
			&2022&AlphaPose~\citep{fang2022alphapose}&74.7&-\\
            &2023&SLT-Pose~\citep{gai2023spatiotemporal}&81.5&-\\
            &2023&DiffPose~\citep{feng2023diffpose}&83.0&-\\
            &2023&TDMI~\citep{feng2023mutual}&\textbf{83.6}&-\\
            \cmidrule(lr){2-5}
			\multirow{1}{*}{Bottom-up}&2019&PGG~\citep{jin2019multi}&\underline{77.0}&-\\
			\hline
			
		\end{tabular*}
  }
}
\vspace{-0.3cm}
\end{table}

\vspace{-0.3cm}
\subsubsection{Datasets for 3D pose estimation}

Compared with the 2D datasets, acquiring high-quality annotation for 3D poses is more challenging and requires motion caption systems (eg., Mocap, wearable IMUs). Therefore, 3D pose datasets are normally built in constrained environments. Currently, Human3.6M and MPI-INF-3DHP are widely used for the task of SPPE, and MuPoTs-3D is often used for MPPE task.

\textbf{The Human3.6M dataset}~\citep{ionescu2013human3} is the largest and most representation indoor dataset for 3D single-person pose estimation. It was collected by recording videos of 11 human subjects performing 17 activities from 4 camera views, and capturing poses by marker-based Mocap systems. In total, this dataset consists of 3.6 million poses with one pose in one frame. This dataset is suitable for the HPE task from images or videos. With video-based HPE, a sequence of frames in a suitable receptive field is considered as the input. Protocol 1 is the most common protocol which applies frames of 5 subjects (S1, S5, S6, S7, S8) for training and the frames of 2 subjects (S9, S11) for test.

\textbf{The MPI-INF-3DHP dataset}~\citep{mehta2017monocular} is a large 3D single-person pose dataset in both indoor and outdoor environments. It was captured by a maker-less MoCap system in a multi-camera studio. There are 8 subjects performing 8 activities from 14 camera views. This dataset provides 1.3 million frames, but more diverse motions than Human3.6M. Same as Human3.6M, this dataset is also suitable for the HPE task from images or videos. The test set includes the frames of 6 subjects with different scenes.

\textbf{The MuPoTs-3D dataset}~\citep{mehta2018single} is a multi-person 3D pose dataset in both indoor and outdoor environments. Same as MPI-INF-3DHP, it was also captured by a multi-view marker-less MoCap system. Over 8,000 frames were collected in 20 videos by 8 subjects. There are some challenging frames with occlusions, drastic illumination changes and lens flares in some outdoor scenes.

\begin{table*}[h!]
	\centering
	\caption{Performance comparison for 3D SPPE on Human3.6M and MPI-INF-3DHP dataset. IB: Image-based, VB: Video-based.}
	\label{tab-3D-image&video-single}
	\resizebox{0.98\linewidth}{!}
	{
		\begin{tabular}{c c c c c c c c}
  \hline 
			
			&\textbf{Category} & \textbf{Year} & \textbf{Method}& \multicolumn{2}{c}{\textbf{\tabincell{c}{Human3.6M\\ \hline MPJPE$\downarrow$\ \ \ \textbf{PMPJPE}$\downarrow$}}}& \multicolumn{2}{c}{\textbf{\tabincell{c}{MPI-INF-3DHP\\ \hline PCK\ \ \ \ \ \ AUC}}}
			\\ \hline 
			
			\multirow{29}{*}{IB}&\multirow{4}{*}{One-stage}&2015&Li et al.~\citep{li2015maximum}&122.0&-&-&-
			\\&&2016&Zhou et al.~\citep{zhou2016deep}&107.3&-&-&-
			\\&&2017&Mehta et al.~\citep{mehta2017monocular}&74.1&-&57.3&28.0
			\\&&2017&WTL~\citep{zhou2017towards}&64.9&-&69.2&32.5
			\\\cmidrule(lr){2-8}&\multirow{25}{*}{Two-stage}&2017&Martinez et al.~\citep{martinez2017simple}&62.9&47.7&-&-
			\\&&2017&Tekin et al.~\citep{tekin2017learning}&69.7&-&-&-
			\\&&2017&Jahangiri et al.~\citep{jahangiri2017generating}&-&68.0&-&-
			\\&&2018&Drpose3d~\citep{wang2018drpose3d}&57.8&42.9&-&-
			\\&&2018&Yang et al.~\citep{yang20183d}&58.6&37.7&80.1&45.8
			\\&&2019&Habibie et al.~\citep{habibie2019wild}&49.2&-&82.9&45.4
			\\&&2019&Chen et al.~\citep{chen2019unsupervised}&-&68.0&71.1&36.3
			\\&&2019&RepNet~\citep{wandt2019repnet}&80.9&65.1&82.5&58.5
			\\&&2019&Hemlets pose~\citep{zhou2019hemlets}&-&-&75.3&38.0
			\\&&2019&Sharma et al.~\citep{sharma2019monocular}&58.0&40.9&-&-
			\\&&2019&Li and Lee~\citep{li2019generating}&52.7&42.6&67.9&-
			\\&&2019&LCN~\citep{ci2019optimizing}&52.7&42.2&74.0&36.7
			\\&&2019&semantic-GCN~\citep{zhao2019semantic}&-&57.6&-&-
			\\&&2020&Iqbal et al.~\citep{iqbal2020weakly}&67.4&54.5&79.5&-
			\\&&2020&Pose2mesh~\citep{choi2020pose2mesh}&64.9&48.0&-&-
			\\&&2020&Srnet~\citep{zeng2020srnet}&44.8&-&77.6&43.8
			\\&&2020&Liu et al.~\citep{liu2020comprehensive}&52.4&41.2&-&-
			\\&&2021&Zou et al.~\citep{zou2021modulated}&49.4&39.1&86.1&53.7
			\\&&2021&GraphSH~\citep{xu2021graph}&51.9&-&80.1&45.8
			\\&&2021&Lin et al.~\citep{lin2021end}&54.0&36.7&-&-
			\\&&2021&Yu et al.~\citep{yu2021towards}&92.4&52.3&86.2&51.7
			\\&&2022&Graformer~\citep{zhao2022graformer}&51.8&-&-&-
			\\&&2022&PoseTriplet~\citep{gong2022posetriplet}&78&51.8&89.1&53.1
			\\&&2023&HopFIR~\citep{zhai2023hopfir}&48.5&-&87.2&57.0
			\\&&2023&SSP-Net~\citep{CARBONERALUVIZON2023109714}&51.6&-&83.2&44.3
			\\&&2023&PHGANet~\citep{2023Learning}&49.1&-&86.9&55.0
			\\&&2023&RS-Net~\citep{10179252}&47.0&38.6&85.6&53.2
			\\
            \cmidrule(lr){1-8}\multirow{29}{*}{VB}&\multirow{5}{*}{One-stage}&2016&Tekin et al.~\citep{tekin2016direct}&125.0&-&-&-
			\\&&2017&Vnect~\citep{mehta2017vnect}&80.5&-&79.4&41.6
			\\&&2018&Dabral et al.~\citep{dabral2018learning}&52.1&36.3&76.7&39.1
			\\&&2022&IVT~\citep{qiu2022ivt}&40.2&\textbf{28.5}&-&-
			\\&&2023&CSS~\citep{9921314}&60.1&46.0&-&-
			\\
            \cmidrule(lr){2-8}&\multirow{22}{*}{Two-stage}&2017&RPSM~\citep{lin2017recurrent}&73.1&-&-&-
			
			\\&&2018&Rayat et al.~\citep{rayat2018exploiting}&51.9&42.0&-&-
			\\&&2018&p-LSTMs~\citep{lee2018propagating}&55.8&46.2&-&-
			\\&&2018&Katircioglu et al.~\citep{katircioglu2018learning}&67.3&-&-&-
            \\&&2019&Cheng et al.~\citep{cheng2019occlusion}&42.9&32.8&-&-
   	    \\&&2019&Cai et al.~\citep{cai2019exploiting}&48.8&39.0&-&-
			\\&&2019&TCN~\citep{pavllo20193d}&46.8&36.5&-&-
			\\&&2019&Chirality Nets~\citep{yeh2019chirality}&46.7&-&-&-
            \\&&2020&UGCN~\citep{wang2020motion}&42.6&32.7&86.9&62.1
            \\&&2020&GAST-Net~\citep{liu2020gast}&44.9&35.2&-&-
			\\&&2021&Chen et al.~\citep{chen2021anatomy}&44.1&35.0&87.9&54.0
			\\&&2021&PoseFormer~\citep{zheng20213d}&44.3&34.6&88.6&56.4
			\\&&2022&Strided~\citep{li2022exploiting}&43.7&35.2&-&-
			\\&&2022&Mhformer~\citep{li2022mhformer}&43.0&-&93.8&63.3
            \\&&2022&MixSTE~\citep{zhang2022mixste}&39.8&\underline{30.6}&94.4&66.5
			\\&&2022&UPS~\citep{foo2023unified}&40.8&32.5&-&-\\
            &&2023&DSTFormer~\citep{zhu2022motionbert}&37.5&-&-&-
			\\
            &&2023&GLA-GCN~\citep{yu2023gla}&44.4&34.8&\underline{98.5}&\underline{79.1}
			\\
            &&2023&D3DP~\citep{shan2023diffusionbased}&\underline{35.4}&-&98.0&\underline{79.1}
			\\
            &&2023&DiffPose~\citep{holmquist2022diffpose}&43.3&32.0&84.9&-
			\\
            &&2023&STCFormer~\citep{tang20233d}&40.5&31.8&\textbf{98.7}&\textbf{83.9}
			\\
            &&2023&PoseFormerV2~\citep{zhao2023poseformerv2}&45.2&35.6&97.9&78.8
			\\
            &&2023&MTF-Transformer~\citep{9815549}&\textbf{26.2}&-&-&-
			\\
            
			\hline
		\end{tabular}
  }
  \vspace{-0.3cm}
\end{table*}

\begin{table*}[h!]
	\centering
	\caption{Performance comparison for 3D Image-based MPPE on MuPoTS-3D dataset.}
	\label{tab-3D-image-multi}
		\resizebox{\textwidth}{!}
	{
		\begin{tabular}{c c c c c c c c c}\hline 
			
			&&&\multicolumn{6}{@{}c@{}}{\textbf{MuPoTS-3D}}
			\\ \cmidrule(lr){4-9}
			\textbf{Category}&\textbf{Year}&\textbf{Method}&\multicolumn{2}{@{}c@{}}{\textbf{All people}}  & \multicolumn{4}{@{}c@{}}{\textbf{Matched people}}
			\\ \cmidrule(lr){4-9} 
			\textbf{}&\textbf{}&\textbf{}&\textbf{PCKrel}&\textbf{PCKabs}&\textbf{PCKrel}&\textbf{PCKabs}&\textbf{PCKroot}&\textbf{AUCrel}
			\\ \hline 
			
			\multirow{5}{*}{Top-down}&2019&LCR-Net~\citep{rogez2019lcr}&70.6&-&74.0&-&-&-\\
			&2019&Moon et al.~\citep{moon2019camera}&81.8&31.5&82.5&31.8&31.0&40.9\\
			&2020&HDNet~\citep{lin2020hdnet}&-&-&83.7&35.2&-&-\\
			&2020&HMOR~\citep{wang2020hmor}&-&-&82.0&\textbf{43.8}&-&-\\
            &2022&Cha et al.~\citep{cha2022multi}&\textbf{89.9}&-&\textbf{91.7}&-&-&-\\ 
           \cmidrule(lr){2-9}
			\multirow{6}{*}{Bottom-up}&2018&Mehta et al.~\citep{mehta2018single}&65.0&-&69.8&-&-&-\\
            &2020&Kundu et al.~\citep{kundu2020unsupervised}&74.0&28.1&75.8&-&-&-\\
			&2020&XNect~\citep{mehta2020xnect}&70.4&-&75.8&-&-&-\\
			&2020&Smap~\citep{zhen2020smap}&73.5&35.4&80.5&38.7&45.5&42.7\\
			&2022&Liu et al.~\citep{liu2022explicit}&79.4&36.5&\underline{86.5}&39.3&-&-\\
            &2023&AKE~\citep{chen2023multi}&74.7&37.2&81.1&40.1&-&-\\ 
            \cmidrule(lr){2-9}
			\multirow{3}{*}{One-stage}&2022&Wang et al.~\citep{wang2022distribution}&\underline{82.7}&\underline{39.2}&-&-&-&-\\
	 &2022&DRM~\citep{jin2022single}&80.9&\textbf{39.3}&85.1&\underline{41.0}&45.6&45.4\\
            &2023&WSP~\citep{qiu2023weakly}&82.4&-&83.2&-&-&-\\
			
			\hline
		\end{tabular}
	}
\end{table*}

			

   
\vspace{-0.3cm}
\subsubsection{Performance comparison}

In Table \ref{tab-2D-image-single&multi}, we present a comparison of different methods for 2D image-based SPPE and MPPE on 
the COCO dataset. For the SPPE task, the performance of heatmap-based methods generally outperforms the regression-based methods.  This superiority can be attributed to the richer spatial information provided by heatmaps, where the probabilistic prediction of each pixel enhances the accuracy of keypoint localization. However, heatmap-based methods~\citep{ye2023distilpose} suffer seriously from the quantization error problem and high-computational cost using high resolution heatmaps. For the MPPE task, the top-down methods overall outperform the bottom-up methods by the success of existing SPPE techniques after detecting individuals. However, they suffer from early commitment and have greater computational costs than bottom-up methods. One-stage methods speed up the process by eliminating the intermediate operations (eg., grouping, ROI, NMS) introduced by top-down and bottom-up methods, while their performance~\citep{liu2023group} is still lower (about 9\% of AP score in the best case) than top-down methods~\citep{xu2022vitpose}. Moreover, It is also observed that the backbone and input image size are two factors for the results. The commonly used backbone includes ResNet, HRNet and Hourglass. The recent Transformer-based network (eg., ViTAE-G, Swin-L) can be also used as the backbone and the method~\citep{xu2022vitpose} based on ViTAE-G network achieves the best performance. When using the same backbone~\citep{zhang2020distribution,yang2021transpose} for the same category of methods, the larger the image size, the better the performance.

Table \ref{tab-2D-video-single} and Table \ref{tab-2D-video-multi} compare the different methods for 2D video-based SPPE and MPPE. Overall, two categories of methods for video-based SPPE achieve comparable results on two datasets. Yet sample frames-based methods\citep{zeng2022deciwatch} are generally faster than frame-by-frame ones by ignoring looking at all frames. Similar to image-based MPPE, the top-down methods achieve better performance than the bottom-up methods for video-based MPPE.

For 3D pose estimation, taken Human3.6M, MPI-INF-3DHP and MuPoTS-3D datasets as examples, Table~\ref{tab-3D-image&video-single} and Table \ref{tab-3D-image-multi} respectively shows the comparisons for SPPE and MPPE from images or videos. The comparison for video-based MPPE was not conducted due to only fewer existing methods. For the SPPE task, two-stage methods normally lift 3D poses from the estimated 2D poses, they generally outperform one-stage methods due to the success of the 2D pose estimation technique. It is also noted that the recent one-stage method based on Transformer network~\citep{qiu2022ivt} also achieves pretty good results. Compared to the same category of methods between images and videos, the performance based on videos is better than the ones based on images. It demonstrates that the temporal information of videos is beneficial for estimating more accurate poses. From Table~\ref{tab-3D-image-multi}, good progress has been made in recent years for the MPPE task. Specifically, one-stage methods generally perform better than most top-down and bottom-up methods, which further implies that the end-to-end training could reduce intermediate errors such as human detection and joint grouping.

\begin{table*}[h!]
	\centering
	\caption{Datasets for Pose tracking. MOTA: Multiple Object Tracking Accuracy, PCP: Percentage of Correct Parts, KLE: Keypoint Localization Error.}    
	\label{tab-posetracking}
	\resizebox{\textwidth}{!}{
		\begin{tabular}{lllllll}    
			\hline \textbf{Dataset} & \textbf{Year}& \textbf{Citation} & \textbf{\#Joints} & \textbf{Size} & \textbf{2D/3D} & \textbf{Metrics} \\ \hline   
			\tabincell{c}{VideoPose2.0~\citep{sapp2011parsing}} & 2011&198 & {-} & 44 videos & {2D} & \tabincell{c}{AP} \\ 
   			\tabincell{c}{Multi-Person PoseTrack~\citep{iqbal2017posetrack}} & 2017&238 & {14} & 16 subjects, 60 videos & {2D} & {MOTA} \\ 
			\tabincell{c}{PoseTrack17~\citep{andriluka2018posetrack}} & 2018&420& {15} & 40 subjects, 550 videos & {2D} & {MOTA} \\ 
   			\tabincell{c}{PoseTrack18~\citep{andriluka2018posetrack}} & 2018&420& {15} & 1138 videos & {2D} & {MOTA} \\ 
			\tabincell{c}{ICDPose~\citep{girdhar2018detect}} & 2018&250 & {14} & 60 videos & {2D} & {MOTA} \\
   			\tabincell{c}{Campus dataset~\citep{berclaz2011multiple}} & 2011&1253 & {-} & 3 subjects, 3 views, 6k frames& {3D} & {PCP} \\ 
			\tabincell{c}{Outdoor Pose~\citep{ramakrishna2013tracking}} & 2013&61 & {14} & 4 subjects, 828 frames & {3D} & {PCP/KLE} \\ 
			\tabincell{c}{CMU Panoptic~\citep{joo2017panoptic}} & 2017&680 & {15} & 8 subjects, 480 views, 65 videos & {3D} & {MOTA} \\
			
			\hline
	\end{tabular}}
\end{table*}

\begin{table}[h!]
	\centering
	\caption{Performance comparison for 2D single person pose tracking on Videopose2.0.}
        \setlength{\tabcolsep}{0.4mm}{
		\begin{tabular*}{\linewidth}{@{\extracolsep{\fill}}lccc}
         \hline 
			
			\textbf{Method} & \textbf{Category} & \textbf{Year} & \textbf{AP}
			\\ \hline 
			Zhao et al. \\~\citep{zhao2015tracking} &Post-processing&2015&85.0\\
			Samanta et al.\\ ~\citep{samanta2016data}&Post-processing&2016&\underline{89.9}\\
			Zhao et al. \\~\citep{zhao2015learning}&Integrated&2015&80.0\\
			Ma et al. \\~\citep{ma2016local}&Integrated&2016&\textbf{95.0}\\    
			\hline		
		\end{tabular*}
 }
  	\label{tab1-2SPPT-acc}
\end{table}

\begin{table*}[h!]
	\centering
	\caption{Performance comparison for 2D multi-person pose tracking on PoseTrack2017 and PoseTrack2018.}
	\label{tab1-2MPPT-MOTA}
	\resizebox{\textwidth}{!}
	{
		\begin{tabular}{l c c c c c c}\hline 		
			\textbf{Method} & \textbf{Category} & \textbf{Year} 		
			& \ {\textbf{\tabincell{c}{2017 Testing\\ \hline MOTA}}} 		
			& \ {\textbf{\tabincell{c}{2017 Validation\\ \hline MOTA}}}
                & \ {\textbf{\tabincell{c}{2018 Testing\\ \hline MOTA}}}
                & \ {\textbf{\tabincell{c}{2018 Validation\\ \hline MOTA}}}	
			\\ \hline 
			Detect-and-Track\\~\citep{girdhar2018detect} &Top-down&2018&51.8&55.2&-&-\\
			Pose Flow\\~\citep{xiu2018pose}&Top-down&2018&51.0&58.3&-&-\\
			Flow Track\\~\citep{xiao2018simple}&Top-down&2018&57.8&65.4&-&-\\
			Fastpose\\~\citep{zhang2019fastpose}&Top-down&2019&57.4&63.2&-&-\\
			LightTrack\\~\citep{ning2020lighttrack}&Top-down&2020&58.0&-&-&64.6\\
			Umer et al.\\~\citep{umer2020self}&Top-down&2020&\underline{60.0}&68.3&60.7&\underline{69.1}\\
        	Clip Tracking\\~\citep{wang2020combining}&Top-down&2020&\textbf{64.1}&71.6&64.3&68.7\\
                Yang et al.\\~\citep{yang2021learning}&Top-down&2021&-&\textbf{73.4}&-&\textbf{69.2}\\
                AlphaPose\\~\citep{fang2022alphapose}&Top-down&2022&-&65.7&-&64.7\\
                GatedTrack\\~\citep{doering2023gated}&Top-down&2023&-&-&-&64.5\\
			Posetrack\\~\citep{iqbal2017posetrack}&Bottom-up&2017&48.4&-&-&-\\
			Raaj et al.\\~\citep{raaj2019efficient}&Bottom-up&2019&53.8&62.7&-&60.9\\
			Jin et al.\\ ~\citep{jin2019multi}&Bottom-up&2019&-&\underline{71.8}&-&-\\						
			\hline
			
		\end{tabular}
	}
\end{table*}

%

\begin{table}[h!]
	\centering
	\caption{Performance comparison for 3D multi-person pose tracking on CMU Panoptic and Campus dataset.}
	\label{tab1-3MPPT-MOTA}
	 \setlength{\tabcolsep}{0.3mm}
	{
		\begin{tabular}{l c c c c}\hline 	
			
			\textbf{Method}  & \textbf{Category} & \textbf{Year} & \ {\textbf{\tabincell{c}{ CMU \\ \hline MOTA}}}& \ {\textbf{\tabincell{c}{ Campus\\ \hline PCP}}} 	
			
			\\ \hline 
			Bridgeman et al. \\\citep{bridgeman2019multi}& Multi-stage&2019&-&92.6\\
			Tessetrack \\\citep{reddy2021tessetrack}&One-stage&2021&94.1&\textbf{97.4}\\
			Voxeltrack \\\citep{zhang2022voxeltrack}&One-stage&2022&\textbf{98.5}&\underline{96.7}\\
			Snipper\\\citep{zou2023snipper}&One-stage&2023&93.4&-\\
            TEMPO\\\citep{choudhury2023tempo}&One-stage&2023&\underline{98.4}&-\\
			\hline
			
		\end{tabular}
	}
\end{table}

\vspace{-0.3cm}
\subsection{Pose tracking}
This section reviews the datasets for pose tracking and also compares different methods on some datasets.
\vspace{-0.3cm}
\subsubsection{Datasets}

Table ~\ref{tab-posetracking} summarizes the datasets, with a focus on the Campus, CMP Panoptic, and PoseTrack datasets, which are highly cited and frequently used for evaluating multi-person pose tracking. These datasets are preferred because multi-person poses are more representative of real-world scenarios. In the earlier stage, VideoPose2.0 was often applied for single-person pose tracking. The PoseTrack dataset has been discussed in Section~\ref{dataset-2DHPE}. In the following, we only review other three datasets.

\textbf{The VideoPose2.0 dataset}~\citep{sapp2011parsing} is a video dataset for tracking the poses of upper and lower arms. The videos were collected from TV shows "Friends" and "Lost" and are normally with a single actor and a variety of movements. This dataset includes 44 videos, each lasting 2-3 seconds, totaling 1,286 frames. Each frame is hand-annotated with joint locations. This dataset is an extension of the VideoPose dataset~\citep{weiss2010sidestepping}, but more challenging since about 30\% of lower arms are significantly foreshortened. 

\textbf{The CMU Panoptic Dataset}~\citep{joo2017panoptic} was created by capturing subjects engaged in social interactions using the camera system with 480 views. Subjects were engaged in different games: Ultimatum (with 3 subjects), Prisoner’s dilemma (with 8 subjects), Mafia (with 8 subjects), Haggling (with 3 subjects), and 007-bang game (with 5 subjects). The number of subjects in each game varies from three to eight. In total, this dataset consists of 65 videos and 1.5 million 3D poses estimated using Kinects. It is often used for evaluating multi-person 3D pose estimation and pose tracking methods.

\textbf{The Campus Dataset}~\citep{belagiannis20143d} was collected by capturing interactions among three individuals in an outdoor environment using 3 cameras. It contains 6,000 frames including 3 views, and each view provides 2,000 frames. It is widely used for 3D multi-person pose estimation and tracking. Due to a small number of cameras and wide baseline views, it is challenging for pose tracking.
\vspace{-0.3cm}
\subsubsection{Performance comparison}

Table \ref{tab1-2SPPT-acc} and Table \ref{tab1-2MPPT-MOTA} respectively show the comparison of 2D pose tracking methods. For 2D single-person pose tracking, integrated methods jointly optimize pose estimation and pose tracking within a unified framework, leveraging the benefits of each to achieve better results. From Table \ref{tab1-2SPPT-acc}, it can be observed that one of the integrated methods~\citep{ma2016local} exhibits state-of-the-art performance. For 2D multi-person pose tracking, most methods follow the top-down strategy by well-estimated poses of single-person estimation technique. Undoubtedly, these methods outperform bottom-up ones about 2-15\% of MOTA scores on the Posetrack2017 and 2018 datasets. 
Regarding 3D multi-person pose tracking, there are currently fewer existing works. Among them, one-stage methods perform better than multi-stage methods shown in Table~\ref{tab1-3MPPT-MOTA}, and Voxeltrack ~\citep{zhang2022voxeltrack} achieves the best results. This is because one-stage methods jointly estimate and link 3D poses, which can propagate the errors of sub-tasks in the multi-stage methods back to the input image pixels of videos.

\vspace{-0.4cm}
\subsection{Action recognition}
This section reviews the datasets that are more commonly used for pose-based action recognition and also compares different categories of the methods. 

\begin{table*}[h!]
	\centering
	\caption{A review of human action recognition datasets. C: Colour, D: Depth, S: Skeleton, I: Infrared frame; LOSubO: Leave One Subject Out, CS: Cross Subject, CV: Cross Validation; tr: training, va: validation, te: test}
	\label{tab-AR}
	\resizebox{\textwidth}{!}
	{
		\begin{tabular}{lllllllll}    
			\hline \textbf{Dataset} & \textbf{Year}  & \textbf{Citation}& \textbf{Modality}  &\textbf{Sensors} & \textbf{\#Actions} & \textbf{\#Subjects} & \textbf{\#Samples}  & \textbf{Protocol} \\ \hline
			\tabincell{l}{HDM05\\~\citep{muller2007mocap}} & {2007} &503& {C,D,S} & {RRM} & 130 & 5 & 2317 & {10-fold CV}  \\ 
			\tabincell{l}{MSR-Action3D\\~\citep{li2010action}} & {2010} &1736& {D,S} &Kinect & 20 &10&557 &CS(1/3 tr; 2/3 tr; half tr, half te)\\ 
			\tabincell{l}{MSRC-12\\~\citep{fothergill2012instructing}} & {2012} &494& {S} & {Kinect}& 12&30&6244 & {LOSubO} \\ 
			\tabincell{l}{G3D\\~\citep{bloom2012g3d}} & {2012} &262& {C,D,S} & {Kinect}& 20&10&659 & {CS(4 tr, 1 va, 5 te)} \\ 
			\tabincell{l}{SBU Kinect\\~\citep{yun2012two}} & {2012} &575& {C,D,S} & {Kinect}  & 8&7&300& {5-fold CV} \\ 
			\tabincell{l}{UTKinect-Action3D\\~\citep{xia2012view}} & {2012} &1716& {C,D,S} & {Kinect} & 10&10&200 & {LOSubO} \\ 
			\tabincell{l}{Northwestern-UCLA\\~\citep{wang2014cross}} & {2014}&497 & {C,D,S} & {Kinect}& 10&10&1494  & LOSubO; cross view(2 tr, 1 te) \\ 
			\tabincell{l}{UTD-MHAD\\~\citep{chen2015utd}} & {2015} &706& {C,D,S,I}  & {Kinect}& 27&8&861 & {CS(odd tr, even te)} \\ 			
			\tabincell{l}{SYSU\\~\citep{hu2015jointly}} & {2015} &594& {C,D,S}& {Kinect} & 12&40&480 &CS(half tr, half te) \\ 
			\tabincell{l}{NTU-RGB+D\\~\citep{shahroudy2016ntu}} & {2016} &2452& {C,D,S,I}  & {Kinect}& 60&40&56880 & CS(half tr, half te); cross view(half tr, half te) \\
			\tabincell{l}{PKU-MMD\\~\citep{liu2017pku}} & {2017} &195& {C,D,S,I} & {Kinect}& 51&66&1076  & CS(57 tr, 9 te); cross view(2 tr, 1 te) \\ 
			
			\tabincell{l}{Kinetics\\~\citep{kay2017kinetics}} & {2017} &3402& {C,S} & {YouTube}& 400 &-&306245  & {CV(250-1000 tr, 50 va, 100 te per action)} \\ 
   			\tabincell{l}{NTU RGB+D 120\\~\citep{liu2019ntu}} & {2019} &907& {C,D,S,I}& {Kinect} & 120&106&114480  & CS(half tr, half te); cross view(half tr, half te)\\ 
			\hline
		\end{tabular}
	}
\end{table*}

\vspace{-0.4cm}
\subsubsection{Datasets}

In Section~\ref{sec:ar}, we have reviewed the pose-based action recognition methods which can be divided into estimated pose-based and skeleton-based action recognition. The former one applies RGB data and the latter one directly uses skeleton data as the input. Table~\ref{tab-AR} summaries the large-scale datasets that are prevalent in deep learning-based action recognition. 

\textbf{NTU RGB+D dataset}~\citep{shahroudy2016ntu} was constructed by Nanyang Technological University, Singapore. Four modalities were collected using Mincrosoft Kinect v2 sensor including RGB, depth maps, skeletons and infrared frames. The dataset consists of 60 actions performed by 40 subjects. The actions can be divided into three groups including: 40 daily actions, 9 health-related actions and 11 person-person interaction actions. The age range of the subjects is from 10 to 35 years and each subject performs an action for several times. In total, there are 56880 samples which are captured in 80 distinct camera views. The large amount of variation in subjects and views makes it possible to have more cross-subject and cross-view evaluations for action recognition methods.

\textbf{NTU RGB+D 120 dataset}~\citep{liu2019ntu} is an extension of the NTU RGB+D dataset~\citep{shahroudy2016ntu}. An additional 60 action categories performed by another 66 subjects including 57,600 samples were added to the NTU RGB+D dataset. This dataset also provides four modalities including RGB, depth maps, skeletons and infrared frames. More number of actions, subjects and samples enable it more challenging than NTU RGB+D dataset in action recognition.

\begin{table}[h!]
\centering
\caption{Performance of estimated pose-based action recognition methods on three datasets for showing the benefits of pose estimation or tracking for recognition. GT: ground-truth.}
\label{tab-estimated-pose}
\footnotesize{
\setlength{\tabcolsep}{0.2mm}{
\begin{tabular*}{\linewidth}{@{\extracolsep{\fill}}l c c c}
  \hline 	
  \textbf{Dataset}&\textbf{Method}&\textbf{Highlights}&\textbf{Accuracy}\\
   \hline
\multirow{3}{*}{JHMDB}&\multirow{2}{*}{PoTion}&estimated poses&58.5$\pm$1.5\\
&&GT poses&62.1$\pm$1.1\\
&\citep{choutas2018potion}&GT poses + crop&67.9$\pm$2.4\\
\cmidrule(lr){2-4}
\multirow{3}{*}{AVA}&\multirow{2}{*}{LART}&-poses-tracking&40.2\\
& &-poses&41.4\\
&\citep{rajasegaran2023benefits}&full model&42.3\\
\cmidrule(lr){2-4}
\multirow{2}{*}{NTU60}&UPS&separate training &89.6\\
&\citep{foo2023unified}&joint training &92.6\\
\hline	
\end{tabular*}
}
}
\end{table}

\textbf{PKU-MMD dataset}~\citep{liu2017pku} is a large-scale multi-modality dataset for action detection and recognition tasks. Four modalities including RGB, depth maps, skeletons and infrared frames were captured by Mincrosoft Kinect v2 sensor. This dataset consists of 1,076 videos composed of 51 actions which are performed by 66 subjects in 3 views. The action classes cover 41 daily actions and 10 person-person interaction actions. Each video contains more than twenty action samples. In total, this dataset includes 3,000 minutes and 5,400,000 frames. The large amount of actions in one untrimmed video makes the robustness of action detection methods.

\textbf{Kinetics-Skeleton dataset}~\citep{kay2017kinetics} is an extra large-scale action dataset captured by searching RGB videos from YouTube and generating skeletons by OpenPose. It has 400 actions, with 400-1150 clips for each action, each from a unique YouTube video. Each clip lasts around 10 seconds. The total number of video samples is 306,245. The action classes include: person actions, person-person actions and person-object actions. Due to the source of YouTube, the videos are not as professional as the ones recorded in experimental background. Therefore, the dataset has considerable camera motion, illumination variations, shadows, background clutter and a large variety of subjects.

\begin{table*}[h!]
	\centering
	\caption{Performance comparison of action recognition methods on NTU RGB+D and NTU RGB+D 120 datasets.}
	\label{tab1-3hpe-acc}
	\resizebox{\textwidth}{!}
	{
		\begin{tabular}{l c c c c c c c}\hline

			\textbf{Method} & \textbf{Category} & \textbf{Sub-category} & \textbf{Year} 
			
			& \multicolumn{2}{c}{\textbf{\tabincell{l}{NTU RGB + D 60 \\ \hline C-Sub\ \ \ \ \ \ C-Set}}} 
			
			&\multicolumn{2}{c}{\textbf{\tabincell{l}{NTU RGB + D 120\\ \hline C-Sub\ \ \ \ \ \ C-Set}}}
			
			\\ \hline 
			Zolfaghari et al.~\citep{zolfaghari2017chained}&Estimated Pose-based&two-stage strategy&2017&80.8&-&-&-\\  

			Liu et al.~\citep{liu2018recognizing}&Estimated Pose-based&two-stage strategy&2018&91.7&95.3&-&-\\       IntegralAction~\citep{moon2021integralaction}&Estimated Pose-based&two-stage strategy&2021&91.7&-&-&-\\
			PoseConv3D~\citep{duan2022revisiting}&Estimated Pose-based&two-stage strategy&2021&\underline{94.1}&97.1&86.9&90.3\\
			Luvizonet al.~\citep{luvizon20182d}&Estimated Pose-based&one-stage strategy&2018&85.5&-&-&-\\  
   UPS~\citep{foo2023unified}&Estimated Pose-based&one-stage strategy&2023&92.6&97.0&89.3&91.1\\  
			2 Layere P-LSTM~\citep{shahroudy2016ntu}&RNN-based& spatial division of human body&2016&62.9&70.3&-&-\\
			Trust Gate ST-LSTM~\citep{liu2016spatio}&RNN-based& spatial and/or temporal networks&2016&69.2&77.7&-&-\\
			Two-stream RNN~\citep{wang2017modeling}&RNN-based& spatial and/or temporal networks&2017&71.3&79.5&-&-\\ 
			
			Zhang et al.~\citep{zhang2017geometric}&RNN-based& spatial and/or temporal networks&2017&70.3 &82.4&-&-\\
			SR-TSL~\citep{si2018skeleton}&RNN-based& spatial and/or temporal networks&2018&84.8&92.4&-&-\\ 
			GCA-LSTM~\citep{liu2017global}&RNN-based&attention mechanism&2017&74.4&82.8&58.3&59.2\\

			STA-LSTM~\citep{song2018spatio}&RNN-based&attention mechanism&2018&73.4&81.2&-&-\\   
			EleAtt-GRU~\citep{zhang2019eleatt}&RNN-based&attention mechanism&2019&80.7&88.4&-&-\\   
			2s AGC-LSTM~\citep{si2019attention}&RNN-based&attention mechanism&2019&89.2&95.0&-&-\\
			JTM~\citep{wang2016action}&CNN-based&2D CNN&2017&73.4&75.2&-&-\\
			JDM~\citep{li2017joint}&CNN-based&2D CNN&2017&76.2&82.3&-&-\\
			Liu et al.~\citep{liu2017enhanced}&CNN-based&2D CNN&2017&80.0&87.2&60.3&63.2\\
			SkeletonNet~\citep{ke2017skeletonnet}&CNN-based&2D CNN&2017&75.9&81.2&-&-\\
			Ke et al.~\citep{ke2017new}&CNN-based&2D CNN&2017&79.6&86.8&-&-\\
			Li et al.~\citep{li2017skeleton}&CNN-based&2D CNN&2017&85.0&92.3&-&-\\
			Ding et al.~\citep{ding2017investigation}&CNN-based&2D CNN&2017&-&82.3&-&-\\
			Li et al.~\citep{li2019learning}&CNN-based&2D CNN&2017&82.8&90.1&-&-\\
			TSRJI~\citep{caetano2019skeleton}&CNN-based&2D CNN&2019&73.3&80.3&65.5&59.7\\
			SkeletonMotion~\citep{caetano2019skeleton}&CNN-based&2D CNN&2019&76.5&84.7&67.7&66.9\\
			3SCNN~\citep{liang2019three}&CNN-based&2D CNN&2019&88.6&93.7&-&-\\           
			DM-3DCNN~\citep{hernandez20173d}&CNN-based&3D CNN&2017&82.0&89.5&-&-\\
			ST-GCN~\citep{yan2018spatial}&GCN-based&static method&2018&81.5&88.3&-&-\\
			STIGCN~\citep{huang2020spatio}&GCN-based&static method&2020&90.1&96.1&-&-\\
			MS-G3D~\citep{liu2020disentangling}&GCN-based&static method&2020&91.5&96.2&86.9&88.4\\
			CA-GCN~\citep{zhang2020context}&GCN-based&static method&2020&83.5&91.4&-&-\\   
	    	AS-GCN~\citep{li2019actional}&GCN-based&dynamic method&2018&86.8&94.2&-&-\\
			2s-AGCN~\citep{shi2019two}&GCN-based&dynamic method&2020&88.5&95.1&-&-\\
			SGN~\citep{zhang2020semantics}&GCN-based&dynamic method&2020&89.0&94.5&79.2&81.5\\
			4s Shift-GCN~\citep{cheng2020skeleton}&GCN-based&dynamic method&2020&90.7&96.5&85.9&87.6\\
			DC-GCN+ADC~\citep{cheng2020decoupling}&GCN-based&dynamic method&2020&90.8&96.6&86.5&88.1\\
			DDGCN~\citep{korban2020ddgcn}&GCN-based&dynamic method&2020&91.1&97.1&-&-\\
			
			Dynamic GCN~\citep{ye2020dynamic}&GCN-based&dynamic method&2020&91.5&96.0&87.3&88.6\\
			
			CTR-GCN~\citep{chen2021channel}&GCN-based&dynamic method&2021&92.4&96.8&88.9&90.6\\
			InfoGCN~\citep{chi2022infogcn}&GCN-based&dynamic method&2021&93.0&97.1&89.8&91.2\\ 
			DG-STGCN~\citep{duan2022dg}&GCN-based&dynamic method&2022&93.2&97.5&89.6&91.3\\
			TCA-GCN~\citep{wang2022skeleton}&GCN-based&dynamic method &2022&92.8&97.0&89.4&90.8\\
       
            ML-STGNet~\citep{9997556}&GCN-based&dynamic method&2023&91.9&96.2&88.6&90.0\\ 
            MV-IGNet~\citep{9234715}&GCN-based&dynamic method&2023&89.2&96.3&83.9&85.6\\ 
            S-GDC~\citep{10023982}&GCN-based&dynamic method&2023&88.6&94.9&85.2&86.1\\   
            Motif-GCN+TBs~\citep{9763364}&GCN-based&dynamic method&2023&90.5&96.1&87.1&87.7\\ 
            3s-ActCLR~\citep{lin2023actionlet}&GCN-based&dynamic method&2023&84.3&88.8&74.3&75.7\\ 
            GSTLN~\citep{DAI2023109540}&GCN-based&dynamic method&2023&91.9&96.6&88.1&89.3\\ 
             4s STF-Net~\citep{WU2023109231}&GCN-based&dynamic method&2023&91.1&96.5&86.5&88.2\\
            LA-GCN~\citep{xu2023language}&GCN-based&dynamic method&2023&93.5&97.2&\underline{90.7}&\underline{91.8}\\
            
	    DSTA-Net~\citep{shi2020decoupled}&Transformer-based&pure Transformer&2020&91.5&96.4&86.6&89.0\\  
			STAR~\citep{shi2021star}&Transformer-based&pure Transformer&2021&83.4&89.0&78.3&80.2\\   
			STST~\citep{zhang2021stst}&Transformer-based&pure Transformer&2021&91.9&96.8&-&-\\
			IIP-Former~\citep{wang2021iip}&Transformer-based&pure Transformer&2022&92.3&96.4&88.4&89.7\\
            RSA-Net~\citep{gedamu2023relation}&Transformer-based&pure Transformer&2023&91.8&96.8&88.4&89.7\\ 

			ST-TR~\citep{plizzari2021spatial}&Transformer-based&hybrid Transformer&2021&89.9&96.1&81.9&84.1\\  
            Zoom Transformer~\citep{zhang2022zoom}&Transformer-based&hybrid Transformer&2022&90.1&95.3&84.8&86.5\\             
            KA-AGTN~\citep{liu2022graph}&Transformer-based&hybrid Transformer&2022&90.4&96.1&86.1&88.0\\  
			STTFormer~\citep{qiu2022spatio}&Transformer-based&hybrid Transformer&2022&92.3&96.5&88.3&89.2\\   
			FG-STFormer~\citep{gao2022focal}&Transformer-based&hybrid Transformer&2022&92.6&96.7&89.0&90.6\\
			GSTN~\citep{jiang2022graph}&Transformer-based&hybrid Transformer&2022&91.3&96.6&86.4&88.7\\ 
            
            IGFormer~\citep{pang2022igformer}&Transformer-based&hybrid Transformer&2022&93.6&96.5&85.4&86.5\\        
            
            3Mformer~\citep{Wang_2023_CVPR}&Transformer-based&hybrid Transformer&2023&\textbf{94.8}&\textbf{98.7}&\textbf{92.0}&\textbf{93.8}\\ 
            
            SkeleTR~\citep{duan2023skeletr}&Transformer-based&hybrid Transformer&2023&\textbf{94.8}&\underline{97.7}&87.8&88.3\\ 
            
            GL-Transformer~\citep{kim2022global}&Transformer-based&unsupervised Transformer&2022&76.3&83.8&66.0&68.7\\ 
            HiCo-LSTM~\citep{dong2023hierarchical}&Transformer-based&unsupervised Transformer&2023&81.4&88.8&73.7&74.5\\ 
            
            HaLP+CMD~\citep{shah2023halp}&Transformer-based&self-supervised Transformer&2023&82.1&88.6&72.6&73.1\\               
            SkeAttnCLR~\citep{ijcai2023p95}&Transformer-based&self-supervised Transformer&2023&82.0&86.5&77.1&80.0\\ 
            
            SkeletonMAE~\citep{10222534}&Transformer-based&self-supervised Transformer&2023&86.6&92.9&76.8&79.1\\             
			\hline
			
		\end{tabular}
	}
\end{table*}

\vspace{-0.3cm}
\subsubsection{Performance comparison}
In Table \ref{tab1-3hpe-acc}, we compare the results of different action recognition methods on two prominent datasets. Estimated poses-based methods apply RGB data as the input, and the best performance~\citep{duan2022revisiting,foo2023unified} is lower than the ones~\citep{Wang_2023_CVPR} used skeletons as the input on two datasets (especially the larger one). This is reasonable because some facts (eg. illumination, background) could affect the performance when using RGB. In particular, methods based on one-stage strategy jointly address pose estimation and action recognition, thus reducing the errors of intermediate steps and generally achieving better results than the methods based on a two-stage strategy. Moreover, Table \ref{tab-estimated-pose} illustrates the effects of pose estimation (PE) and tracking on action recognition (AR). It can be easily seen that pose estimation and tracking results can improve the performance of action recognition, which further emphasizes the relationship of these three tasks.

For the skeleton-based methods, the recent methods mainly apply GCN and Transformer, consistently outperforming CNN
and RNN-based methods. This improvement demonstrate the benefit of local and global feature learning based on GCN and Transformer for action recognition. Specifically, dynamic GCN-based methods generally perform better than static GCN-based ones due to stronger generalization capabilities. Hybrid Transformer-based methods outperform pure Transformer-based ones on large datasets since integrating the Transformer with GCN or CNN can better learn both local and global features. Specifically, the method~\citep{Wang_2023_CVPR} of applying transformer encoder on hypergraph achieved the best performance on two datasets, which provides a hint of representing actions using hypergraph for classification. It is also worth noting that the method~\citep{xu2023language} based on the guidance of natural language respectively achieves pretty good performance on two datasets, which implies the advantage of incorporating linguistic context for action recognition.

\section{Challenges and Future Directions}\label{challenges}

This paper has reviewed recent deep learning-based approaches for pose estimation, tracking and action recognition. It also includes a discussion of commonly used datasets and a comparative analysis of various methods. Despite the the remarkable
successes in these domains, there are still some challenges and corresponding research directions to promote advances for the three tasks.

\vspace{-0.2cm}
\subsection{Pose estimation}
There are five main challenges for the pose estimation task as follows.

(1) Occlusion

Although the current methods have achieved outstanding performance on public datasets, they still suffer from the occlusion problem. Occlusion results in unreliable human detection and declined performance for pose estimation. Person detectors in top-down approaches may fail in identifying the boundaries of overlapped human bodies and body part association for occluded scenes may fail in bottom-up approaches. Mutual occlusion in crowd scenarios caused largely declined performance for current 3D HPE methods.

To overcome this problem, some methods~\citep{dong2019fast,tu2020voxelpose,zhang2021direct} have been proposed based on multi-view learning. This is because the occluded part in one view may become visible in other views. However, these methods often need large memory and expensive computation costs, especially for 3D MPPE under multi-view. Moreover, some methods based on multi-modal learning have also been demonstrated for robustness to occlusion, which could extract enrich features from different sensing modalities such as depth~\citep{shah2019robustness} and wearable inertial measurement units~\citep{zhang2020fusing}. When applying pose estimation from different modalities, it may face another problem of few available datasets with different modalities. With the development of vision-language models, texts could provide semantics for pose estimation and also be easily generated by GPT, thus a better direction for another modality. Based on pose semantics, the occluded part can be inferred. With regard the semantics, human-scene relationships can also provide some semantic cues such as a person cannot be simultaneously present in the locations of other objects in the scene.

(2) Low resolution

In the real-word application, low-resolution images or videos are often captured due to wide-view cameras, long-distance shooting capturing devices and so on. Obscured persons also exist due to environmental shadows. The current methods are usually trained on high-resolution input, which may cause low accuracy when applying them to low-resolution input. One solution for estimating poses from low-resolution input is to recover image resolution by applying super-resolution methods as image pre-processing. However, the optimization of super-resolution does not contribution to high-level human pose analysis. Wang et al.~\citep{wang2022low} observed that low-resolution would exaggerate the degree of quantization error, thus offset modeling may be helpful for pose estimation with low-resolution input.

(3) Computation complexity

As reviewed in Section~\ref{poseestimation}, many methods have been proposed for solving computation complexity. For example, one-stage methods for image-based MPPE are proposed to save the increased time consumption caused by intermediate steps. Sample frames-based methods for video-based pose estimation are proposed to reduce the complexity of processing each frame. However, such one-stage methods may sacrifice accuracy when improving efficiency (eg. the recent ED-pose network~\citep{yang2023explicit} takes the shortest time and would sacrifice about \%4 AP on CoCO val2017 dataset). Therefore, more effort into one-stage methods for MPPE is required to achieve computationally efficient pose estimation while maintaining high accuracy. Sample frames-based methods~\citep{zeng2022deciwatch} estimate poses based on three steps, which still results in more time consumption. Hence, an end-to-end network is preferred to incorporate with sample frames-based methods for video-based pose estimation. 

Transformer-based architectures for video-based 3D pose estimation inevitably incur high computational costs. This is because that they typically regard each video frame as a pose token and apply extremely long video frames to achieve advanced performance. For instance, Strided~\citep{li2022exploiting} and Mhformer~\citep{li2022mhformer} require 351 frames, and MixSTE~\citep{li2022mhformer} and DSTformer~\citep{zhu2022motionbert} require 243 frames. Self-attention complexity increases quadratically with the number of tokens. Although directly reducing the frame number can reduce the cost, it may result in lower performance due to a small temporal receptive field. Therefore, it is preferable to design an efficient architecture while maintaining a large temporal receptive field for accurate estimation. Considering that similar tokens may exist in deep transformer blocks~\citep{wang2022vtc}, one potential solution is to prune pose tokens to improve the efficiency.

(4) Limited data for uncommon poses

The current public datasets have limited training data for uncommon poses (eg. falling), which results in model bias and further low accuracy on such poses. Data augmentation~\citep{jiang2022posetrans,10050391} for uncommon poses is a common method for generating new samples with more diversity. Optimization-based methods~\citep{jiang2023back} can mitigate the impact of domain gaps, by estimating poses case-by-case rather than learning. Therefore, deep-learning-based method combining optimization techniques might be helpful for uncommon pose estimation. Moreover, open vocabulary learning can be also applied to estimating uncommon poses by the semantic relationship between these poses with other common poses.

(5) High uncertainty of 3D poses

Predicting 3D poses from 2D poses is required to handle uncertainty and indeterminacy due to depth ambiguity and potential occlusion. However, most of the existing methods~\citep{shan2023diffusionbased} belong to deterministic methods which aim to construct single and definite 3D poses from images. Therefore, how to handle uncertainty and indeterminacy of poses remains an open question. Inspired by the strong capability of diffusion models to generate samples with high uncertainty, applying diffusion models is a promising direction for pose estimation. Few methods~\citep{gong2022diffpose,holmquist2022diffpose,feng2023diffpose} have been recently proposed by formulating 3D pose estimation as a reverse diffusion process.

\vspace{-0.2cm}
\subsection{Pose tracking}
Most pose tracking methods follow pose estimation and linking strategy, pose tracking performance highly depends on the results of pose estimation. Therefore, some challenges of pose estimation also exist in pose tracking, such as occlusion. Multi-view features fusion~\citep{zhang2022voxeltrack} is one method of eliminating unreliable appearances by occlusion for improving the results of pose linking. Linking every detection box rather than only high score detection boxes~\citep{zhang2022bytetrack} is another method to make up non-negligible true poses by occlusion. In the following, we will present some more challenges for pose tracking.

(1) Multi-person pose tracking under multiple cameras

The main challenge is how to fuse the scenes of different views. Although Voxteltrack~\citep{zhang2022voxeltrack}  tends to fuse multi-view features fusion, it would be researched more. If scenes from non-overlapping cameras are fused and projected in a virtual world, poses can be tracked in a long area continuously. 

(2) Similar appearance and diverse motion

To link poses across frames, the general solution is to measure the similarity between every pair of poses in neighboring frames based on appearance and motion. Persons sometimes have uniform appearance and diverse motions at the same time, such as group dancers, and sports players. They are highly similar and almost undistinguished in appearance by uniform clothes, and in complicated motion and interaction patterns. In this case, measuring the similarity is challenging. However, such poses with similar appearance can be easily distinguished by textual semantics. One possible solution is to incorporate some multi-modality pre-training models, such as Contrastive Language-Image Pre-training (CLIP)~\citep{radford2021learning}, for measuring similarity based on their semantic representation.

(3) Fast camera motion

Existing methods mainly address pose tracking by assuming slow camera motion. However, fast camera motion with ego-camera capturing is very often in real-world application. How to address egocentric pose tracking with fast camera motion is a challenging problem. Khirodkar et al.~\citep{khirodkar2023egohumans} proposed a new benchmark (EgoHumans) for egocentric pose estimation and tracking, and designed a multi-stream transformer to track multiple persons. Experiments have shown that there is still a gap between the performance of static and dynamic capture systems due to camera synchronization and calibration. More effort can be made to bridge the gap.
\vspace{-0.3cm}
\subsection{Action recognition}
With the rapid advancement of deep learning techniques, promise results have been achieved on large-scale action datasets. There are still some open questions as follows.

(1) Computation complexity

According to the performance comparison (Table~\ref{tab1-3hpe-acc}) of different methods, the method of integrating transformer with GCNs achieves the best accuracy. However, as mentioned before the computation required for a transformer and the amount of memory required increases on a quadratic scale with the number of tokens~\citep{ulhaq2022vision}. Therefore, how to select significant tokens from video frames or skeletons is an open question for efficient transformer-based action recognition. Similar to transformer-based pose estimation, pruning tokens or discarding input matches~\citep{qing2023mar} tend to reduce the cost. Moreover, integrating lightweight GCNs~\citep{kang2023efficient} can be further beneficial for efficiency.

(2) Zero-shot learning on skeletons

Annotating and labeling large-amount data is expensive, and zero-shot learning is desirable in real-world applications. Existing zero-shot action recognition methods mainly apply RGB data as the input. However, skeleton data has become a promising alternative to RGB data due to its robustness to variations in appearance and background. Therefore, zero-shot skeleton-based action recognition is more desirable. Few methods~\citep{gupta2021syntactically,zhou2023zero} were proposed to learn a mapping between skeletons and word embedding of class labels. Class labels may possess less semantics than textual descriptions which are natural languages for describing how an action is performed. In the future, new methods can be pursued based on textual descriptions for zero-shot skeleton-based action recognition. 

(3) Multi-modality fusion

Estimated pose-based methods take RGB data as the input and recognize actions based on RGB and estimated skeletons. Moreover, text data can guide improving the performance of visually similar actions and zero-shot learning, which is another modality for action recognition. Due to the heterogeneity of different modalities, how to fully utilize them deserves to be further explored by researchers. Although some methods~\citep{duan2022revisiting} tend to propose a particular model for fusing different modalities, such model lacks of generalization. In the future, a universal fusing method regardless of models is a better option.

\vspace{-0.3cm}
\subsection{Unified models} 
As reviewed in Section~\ref{subsec-estimated}, some methods tend to conduct action recognition based on results of pose estimation or tracking. Table~\ref{tab-estimated-pose} further demonstrates pose estimation and tracking can improve action recognition performance. These observations emphasize these three tasks are closely related together, which provides a direction for designing unified models for solving three tasks. Recently, a unified model (UPS~\citep{foo2023unified}) has been proposed for 3D video-based pose estimation and estimated poses-based action recognition, however, their performance is largely lower than the ones of separate models. Hence, more unified models are preferable for jointly solving these three tasks.

\section{Conclusion}\label{conclusion}
This survey has presented a systematic overview of recent works about human pose-based estimation, tracking and action recognition with deep learning. We have reviewed pose estimation approaches from 2D to 3D, from single-person to multi-person, and from images to videos. After estimating poses, we summarized the methods of linking poses across frames for tracking poses. Pose-based action recognition approaches have been also reviewed which are taken as the application of pose estimation and tracking. For each task, we have reviewed different categories of methods and discussed their advantages and disadvantages. Meanwhile, end-to-end methods were highlighted for jointly conducting pose estimation, tracking and action recognition in the category of estimated pose-based action recognition. Commonly used datasets have been reviewed and performance comparisons of different methods have been covered to further demonstrate the benefits of some methods.

Based on the strengths and weaknesses of the existing works, we point out a few promising future directions. For pose estimation, more effort can be made on pose estimation with occlusion, low resolution, limited data with uncommon poses and balancing the performance with computation complexity. Multi-person pose tracking can be further resolved under multiple cameras, similar appearance, diverse motions and fast camera motion. Zero-shot learning on skeletons and multi-modality fusion can be also further explored for action recognition.

\bmhead{Acknowledgements}
This work is supported by the National Natural Science Foundation of China (Grant No. 62006211, 61502491) and China Postdoctoral Science Foundation (Grant No. 2019TQ0286, 2020M682349).

\bibliographystyle{sn-mathphys}      
\bibliography{IJCV}   

\end{document}